\documentclass{article}
\usepackage{iclr2027_conference,times}

\usepackage[utf8]{inputenc}
\usepackage[T1]{fontenc}

\usepackage{graphicx}
\usepackage{multirow}
\usepackage{amsmath,amsfonts,amsthm,amssymb,dsfont}
\DeclareMathOperator*{\argmin}{arg\,min}
\usepackage{mathrsfs}
\usepackage{xcolor}
\usepackage{hyperref}
\definecolor{linkblue}{rgb}{0,0.08,0.45} 
\hypersetup{colorlinks=true, linkcolor=linkblue, citecolor=linkblue, urlcolor=linkblue}
\usepackage{url}
\usepackage{cleveref}
\usepackage{caption,subcaption}
\usepackage{wrapfig}
\usepackage{booktabs,tabularx}
\usepackage{csquotes}
\usepackage{enumitem}
\usepackage{tikz}
\usepackage{rotating}
\usepackage{placeins}
\usepackage{etoc} %
\usetikzlibrary{calc,decorations.pathreplacing}
\usetikzlibrary{positioning, arrows.meta}
\usetikzlibrary{fit, backgrounds}

\graphicspath{ {./figs/} }

\newcommand{\method}{MUNDO}
\newcommand{\longmethod}{\textbf{M}ultiresolution \textbf{UN}ified \textbf{DO}wnscaling}
\providecommand{\sd}[1]{{\scriptsize$\pm$#1}}

\title{\method: a multiresolution generative\\ framework for temporal downscaling}

\newcommand{\yes}{\textcolor{teal}{\checkmark}}
\newcommand{\yesp}{\textcolor{teal}{\checkmark}$^{\pi}$}   %
\newcommand{\soft}{\textcolor{orange!75!black}{$\sim$}} %
\newcommand{\no}{\textcolor{red!65!black}{$\times$}}
\newcommand{\na}{--}

\newcommand\blfootnote[1]{%
  \begingroup\renewcommand\thefootnote{}\NoHyper\footnotetext{#1}\endNoHyper\endgroup}

\author{Aymeric Delefosse$^{1*}$, Manon Tavernier$^{1*}$, Julie Keisler$^{1}$, \\
\bf Anastase Charantonis$^{1}$ \& Dominique Béréziat$^{2}$ \\
$^{1}$Inria, ARCHES, Paris, France \\
$^{2}$Sorbonne Universit\'e, CNRS, LIP6, Paris, France
}

\iclrfinalcopy

\begin{document}

\maketitle
\lhead{Preprint} %
\blfootnote{$^{*}$Equal contribution.}

\begin{abstract}
  Dynamical systems are often observed at coarse temporal resolution through sparse states or temporal aggregates, leaving the underlying fine-scale trajectories unresolved. Recovering these trajectories requires probabilistic models that capture the unresolved variability while remaining consistent with the observations. 
  We present \method{} (\longmethod{}), a generative framework for temporal downscaling under general linear observation operators. \method{} introduces a null-space parameterization that enforces exact observation consistency in a single network evaluation, unifying temporal interpolation and disaggregation. To capture complex multiscale dynamics, we leverage reverse Markov learning to factorize the observation operator into a hierarchy of successive coarsenings. Each level models only the degrees of freedom lost between adjacent resolutions.
  We evaluate \method{} on two-scale Lorenz-96, four two-dimensional physical systems, and multivariate ERA5 reanalysis over Europe, for both temporal interpolation and disaggregation. Across datasets and tasks, \method{} generally outperforms flow-based baselines. Single-step \method{} achieves strong predictive skill, while recursive \method{} generally improves temporal realism and ensemble calibration, with better spread-skill agreement and predictive coverage.
\end{abstract}

\etocdepthtag.toc{main}
\graphicspath{ {../figs/} }

\section{Introduction}
\label{sec:introduction}

\begin{figure}[h]
	\centering
	\input{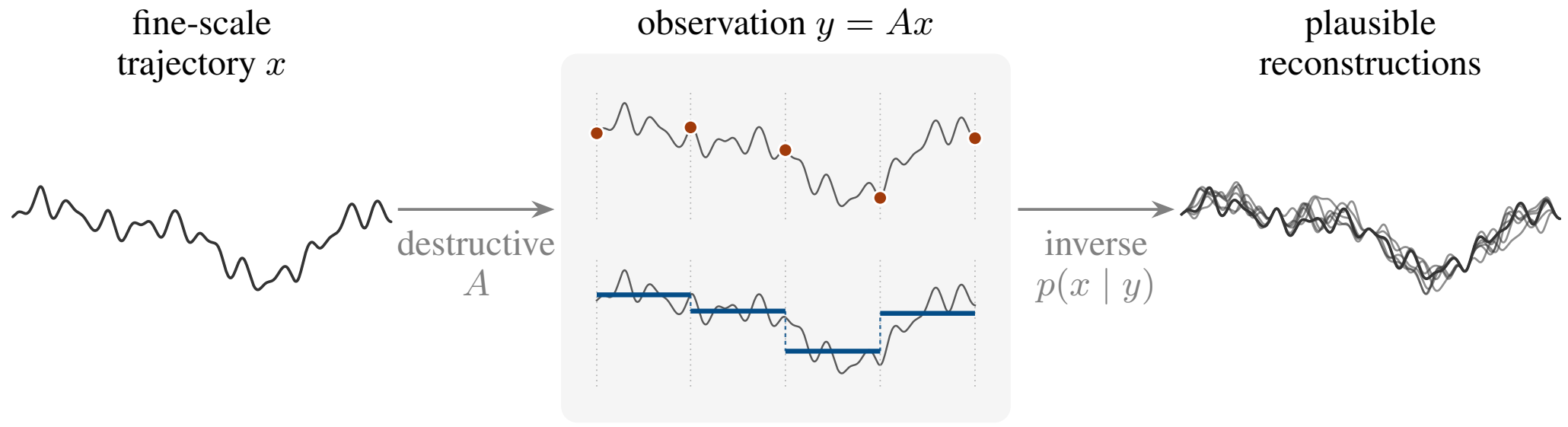}%
	\caption{\textbf{Coarse observations of a dynamical system define a one-to-many inverse problem.} Subsampling (orange dots) yields interpolation, while averaging (blue bars) yields disaggregation. In both cases, the reconstruction target is the conditional law $p(x \mid y)$.}
	\label{fig:two-faces}
\end{figure}%

A dynamical system is rarely observed at the temporal resolution at which it evolves. Numerical models integrate at fine time steps but typically archive only a sparse subset or an aggregate of states, as storing the full trajectory is prohibitively expensive. This is especially common in weather and climate: climate simulations often retain daily or monthly means of sub-hourly dynamics \citep{eyringOverviewCoupledModel2016}, while data-driven weather forecasters typically predict states at six-hour intervals \citep{lamLearningSkillfulMediumrange2023,kurth2023fourcastnet}. The same loss of fine-scale temporal information occurs in other simulated and observed physical systems, including fluid and wave dynamics. Temporal downscaling reconstructs the missing information at a fraction of the cost of rerunning or training models \citep{brenowitz2025climate}. It covers, often without distinction, \emph{interpolation}, where sparse states are observed, and \emph{disaggregation}, where aggregates are given (Figure~\ref{fig:two-faces}). Both are instances of the same inverse problem: recovering a fine trajectory from an incomplete observation of it.

For chaotic, multiscale dynamics, this inverse problem is one-to-many: endpoints do not determine the trajectory between them, and an aggregate does not determine the fine-scale variability that produced it. Therefore, the target is a distribution over trajectories whose samples are consistent with the observation. Learned temporal downscalers are mostly deterministic and optimized for pointwise error \citep{harilalSTintSelfsupervisedTemporal2023a,wangTemDeepSelfsupervisedFramework2025,leinonenModulatedAdaptiveFourier2024}, so they return one reconstruction, the conditional mean, and none of the range compatible with the observation. Probabilistic methods represent that range but enforce the observation only indirectly, through conditioning or penalty terms \citep{ingstadHourGlassProbabilisticDatadriven2026,bassetti2024}, so a draw need not satisfy it. 
Posterior samplers accept arbitrary linear operators and enforce the observation through guidance or repeated data-consistency updates \citep{kimFlowdpsFlowdrivenPosterior2025,aichWINDWeatherInverse2026a}.
Among these methods, none returns draws that satisfy the observation exactly under both operators without an iterative solve.

This paper presents \method{} (\longmethod{}), a generative framework for temporal downscaling under general linear observation operators. \method{} builds on engression \citep{shenEngressionExtrapolationLens2025}, which learns a conditional sampler with the energy score \citep{gneiting2007strictly}, a strictly proper scoring rule.
\method{} generates each draw in a single network evaluation, as a classical reconstruction plus a learned residual in the null space of the observation operator. Every draw therefore satisfies the observation exactly, for both interpolation and disaggregation.
Such a sampler must recover all unresolved scales at once. Temporal downscaling, however, decomposes naturally across resolutions: a large temporal coarsening can be expressed as a composition of smaller coarsenings. We exploit this structure through reverse Markov learning \citep{shenReverseMarkovLearning2025}, which turns the decomposition into a hierarchy of conditional samplers. Each level generates only the degrees of freedom lost between adjacent resolutions. The null-space parameterization ensures exact consistency of each level with its parent, and hence with the observation.

Our contributions are:
\begin{itemize}[nosep]
	\item a null-space parameterization for general linear observation operators, which gives exact observation consistency in one network evaluation per draw, for interpolation and disaggregation alike;
	\item a multiresolution factorization of the observation operator in which each level generates only the degrees of freedom lost between adjacent resolutions and is exactly consistent with its parent;
	\item end-to-end fine-tuning of the hierarchy on its own samples, which removes the mismatch between the ground-truth parents seen in training and the generated parents received at sampling;
	\item a spatial-coherence term complementary to the energy score.
\end{itemize}

We conduct a broad evaluation of \method{} on chaotic systems, simulated physical dynamics, and multivariate ERA5 reanalysis, measuring probabilistic skill, calibration, observation consistency, and temporal structure. 
Throughout, we compare recursive \method{} with its single-step counterpart to isolate what the hierarchy adds.

The paper is organized as follows. Section~\ref{sec:related} reviews related work on temporal downscaling. Section~\ref{sec:formulation} states the inverse problem and the target conditional law, then describes the method. Section~\ref{sec:experiments} describes the systems, baselines, and evaluation protocol, and reports the results.

\section{Related work}
\label{sec:related}

Appendix~\ref{app:related} gives a broader review and a comparative table of the methods below.

\paragraph{Statistical temporal downscaling.}
Denton--Cholette \citep{denton1971,cholette1984} reconstructs the smoothest temporal record consistent with each aggregate. CDF-t \citep{deque2007,michelangeliProbabilisticDownscalingApproaches2009} adjusts marginal distributions across timescales. Both provide classical baselines for temporal disaggregation, but neither models the conditional distribution of fine-scale trajectories.

\paragraph{Learned temporal downscaling.}
STint \citep{harilalSTintSelfsupervisedTemporal2023a}, TemDeep \citep{wangTemDeepSelfsupervisedFramework2025}, and ModAFNO \citep{leinonenModulatedAdaptiveFourier2024} adapt video frame interpolation \citep{jiangSuperSloMoHigh2018} to geophysical fields. Their pointwise losses target the conditional mean, hence the variance deficit of Section~\ref{variance-deficit}.
Generative models avoid this deficit by sampling the conditional law, and differ in how they enforce the observation.
Diffusion models condition on coarse records, for instance to reconstruct daily fields from monthly means \citep{bassetti2024}. However, their samples are not guaranteed to reproduce the conditioning aggregates, and generation requires an iterative denoising chain. HourGlass \citep{ingstadHourGlassProbabilisticDatadriven2026} generates samples in a single forward pass, but encourages endpoint and window-statistics consistency through the loss, and only addresses six-hourly to hourly interpolation.
RainDisaggGAN \citep{scher2021temporal} is exact in a single pass, by scaling a temporal softmax with the daily total, but only for the disaggregation of one non-negative variable.

\paragraph{Inverse-problem solvers.}
A complementary line of work imposes observations on an unconditional generative prior during sampling. DDNM \citep{wang2022zero} preserves the measurement-determined component while refining null-space content during reverse diffusion, which is exact for a linear operator. Guidance-based solvers instead add likelihood-gradient updates to the sampling dynamics, which is only approximate. FlowDPS \citep{kimFlowdpsFlowdrivenPosterior2025} does so for flow matching, \citet{schmidtGenerativeFrameworkProbabilistic2025} to generate hourly ensembles from six-hourly climate fields, and WIND \citep{aichWINDWeatherInverse2026a} with one diffusion prior pretrained on ERA5.

\section{Problem and method}
\label{sec:formulation}

\subsection{Setting and observation operator}

Let $X_i \in \mathbb{R}^{d}$ denote the instantaneous state of the system and $\delta$ the target temporal resolution. A window holds $N$ steps of length $\delta$, and its fine record $x = (x_1, \dots, x_N) \in \mathcal{X} \subseteq \mathbb{R}^{Nd}$ is a draw from $\mu$, the joint law of $N$ consecutive states at resolution $\delta$, whose density is unknown but from which we hold samples. We write $x$ for both the random record and one of its realizations, context resolving the ambiguity.

The fine record is never observed directly, only through an observation operator that degrades it,
\begin{equation}
	y = f(x), \qquad f : \mathbb{R}^{Nd} \to \mathbb{R}^{m}, \quad m < Nd,
	\label{eq:observation}
\end{equation}
where $y$ is the coarse record. We take $f$ linear and identify it with its matrix $A \in \mathbb{R}^{m \times Nd}$, $f(x) = Ax$, assumed to have full row rank $m$ without loss of generality. Then $f$ is surjective onto $\mathbb{R}^{m}$, with right inverse $A^{\dagger} = A^{\top}(AA^{\top})^{-1}$, but not injective: its kernel has dimension $\dim \ker A = Nd - m > 0$. Given $y$, the \emph{compatible set} is the preimage
\begin{equation}
	\mathcal{X}_y \;=\; f^{-1}(\{y\})
	\;=\; \left\{\, x \in \mathbb{R}^{Nd} \;:\; Ax = y \,\right\}
	\;=\; A^{\dagger} y \;+\; \ker A,
	\label{eq:compatible-set}
\end{equation}
an affine subspace directed by $\ker A$. In particular, the true record splits as
\begin{equation}
	x \;=\; A^{\dagger} y \;+\; r,
	\qquad r \;=\; (I - A^{\dagger}A)\,x \;\in\; \ker A,
	\label{eq:decomposition}
\end{equation}
into a component fixed by the observation and a residual orthogonal to it, which the observation leaves entirely undetermined: the kernel is exactly the information that $f$ destroys. 
Equation~\eqref{eq:decomposition} is the range-null space decomposition, $A^{\dagger}y$ carrying the range-space content and $r$ the null-space content. It underlies null-space learning for inverse problems \citep{schwab2019deep}, and is used to enforce data consistency in diffusion-based
image restoration \citep{wang2022zero}. %

In this work, $f$ coarsens the record in time: $x$ is observed only at a coarser step $\Delta = k\delta$, with $k \in \mathbb{N}^{*}$ dividing $N$, so that $n = N/k$ coarse steps remain and $m = nd$, through one of two operators
\begin{equation}
	\left(A_{\mathrm{sub}}\,x\right)_j = x_{jk},
	\qquad\qquad
	\left(A_{\mathrm{avg}}\,x\right)_j = \frac{1}{k}\sum_{i \in B_j} x_i,
	\qquad j = 1, \dots, n ,
	\label{eq:operators}
\end{equation}
where $B_j = \{(j-1)k+1, \dots, jk\}$ is the $j$-th block of $k$ fine steps. Reconstruction from $A_{\mathrm{sub}}$ is interpolation, and from $A_{\mathrm{avg}}$ disaggregation. Each coarse value draws on its own block, so that $AA^{\top} = c\,I_{nd}$, with $c = 1$ for subsampling and $c = 1/k$ for block averaging, and $A^{\dagger} = c^{-1}A^{\top}$. The kernels,
\begin{equation}
	\begin{aligned}
		\ker A_{\mathrm{sub}} & = \left\{\, u : u_{jk} = 0,\; j = 1,\dots,n \,\right\},                          \\
		\ker A_{\mathrm{avg}} & = \left\{\, u : \textstyle\sum_{i \in B_j} u_i = 0,\; j = 1,\dots,n \,\right\} ,
	\end{aligned}
\end{equation}
both of dimension $(N-n)d$, are the sub-$\Delta$ variability the operators discard: the values between retained instants in one case, the fluctuation about the block mean in the other. %

\subsection{An ill-posed inverse problem}

Recovering $x$ from $y$ alone is ill-posed in the sense of \citet{hadamard1902}. By surjectivity, a solution exists for every $y \in \mathbb{R}^{nd}$, but it is never unique, since $\mathcal{X}_y$ is an affine subspace of dimension $(N-n)d$ rather than a singleton \citep{engl1996regularization}. 
A deterministic rule $g : \mathbb{R}^{nd} \to \mathbb{R}^{Nd}$ must therefore select a single representative of $\mathcal{X}_y$. Its kernel component is a deterministic function of $y$, so the conditional law of its output given $y$ is a Dirac mass and cannot represent the variability among trajectories compatible with the same observation. %

\subsection{The variance deficit of the conditional mean}
\label{variance-deficit}

Minimizing expected squared error returns the conditional mean $\mathbb{E}\left[x \mid y\right]$. Since $f$ is linear, the conditional mean remains exactly consistent with the coarse observation, as
\begin{equation}
	f(\mathbb{E}\left[x \mid y\right]) = \mathbb{E}\left[f(x)\mid y\right] = y.
\end{equation}
However, its variance is systematically deficient. Let $\varphi : \mathbb{R}^{Nd} \rightarrow \mathbb{R}$ be a linear functional of the fine record. %
The law of total variance gives
\begin{equation}
	\operatorname{Var}\left[\varphi(\mathbb{E}\left[x \mid y\right])\right] = \operatorname{Var}\left[\varphi(x)\right] - \mathbb{E}\left[\operatorname{Var}\left[\varphi(x) \mid y\right]\right],
	\label{eq:variance-deficit}
\end{equation}
where the second term is the variability in $\varphi(x)$ left unresolved by the coarse observation. 
The conditional mean therefore underestimates the variance of $\varphi(x)$ by exactly that term. This deficit is a consequence of the quadratic criterion, not of any particular model class. It vanishes when $\varphi(x)$ is determined by $y$, and reaches $\operatorname{Var}\left[\varphi(x)\right]$ when $\mathbb{E}\left[\varphi(x) \mid y\right]$ is constant. %

\subsection{Probabilistic formulation}

The limits of deterministic reconstruction motivate a different target. Rather than selecting a single member of $\mathcal{X}_y$, we seek the conditional law of fine records given the coarse observation, defined under $\mu$ as\footnote{Since the conditioning event has probability zero, $\mu(\,\cdot \mid f(x) = y)$ is understood as the disintegration of $\mu$ along $f$, defined for $f_{\#}\mu$-almost every $y$ and satisfying $\mu = \int p(\,\cdot \mid y)\, \mathrm{d}(f_{\#}\mu)(y)$, with $p(\,\cdot \mid y)$ supported on $\mathcal{X}_y$.}
\begin{equation}
	p(x \mid y) \;=\; \mu\left(x \bigm| f(x) = y\right).
	\label{eq:loi-conditionnelle}
\end{equation}
Reconstructing then means \emph{sampling} $p(\,\cdot \mid y)$: at fixed $y$, generating multiple compatible records whose dispersion reflects the variability left unresolved by the observation. %

\subsection{Multiresolution sampling} %
\label{sec:method}

Sampling the conditional law in \eqref{eq:loi-conditionnelle} amounts to generating the residual in \eqref{eq:decomposition}.
For any right inverse $A^{\dagger}$ of $A$, every record compatible with $y$ can be written as a baseline reconstruction plus an element of $\ker A$. \method{} therefore draws
\begin{equation}
	\hat{x} = \underbrace{A^{\dagger}\, y}_{\text{classical reconstruction}}
	\;+\; \underbrace{\left(I - A^{\dagger} A\right)\, G_\theta\left(y, \varepsilon, c\right)}_{\text{residual} \,\in\, \ker A} ,
	\label{eq:level}
\end{equation}
where $G_\theta$ is a neural network, $c$ denotes additional conditioning, and, following engression \citep{shenEngressionExtrapolationLens2025}, randomness enters through $\varepsilon$, with one network evaluation per draw. Since $AA^\dagger=I$, the affine map $P_y(u)=A^\dagger y+(I-A^\dagger A)u$ is a projection onto $\mathcal{X}_y$. Thus, $\hat{x}=P_y(G_\theta(y,\varepsilon,c))$ satisfies $A\hat{x}=y$ by construction, without penalty terms. The residual head is initialized at zero, so an untrained level returns the baseline reconstruction $A^\dagger y$. \method{} does not prescribe the architecture of $G_\theta$; Section~\ref{sec:experiments} specifies the architecture used for each system.

\paragraph{Multiresolution factorization.}\label{sec:ladder}
Following reverse Markov learning \citep{shenReverseMarkovLearning2025}, we extend the single conditional sampler
across a hierarchy of temporal resolutions. Both observation operators factor into a chain of full-row-rank linear coarsenings,
\begin{equation}
	A = A_R \circ \cdots \circ A_1,
	\qquad
	x^{(r)} = A_r\, x^{(r-1)},
	\qquad
	x^{(0)} = x,
	\quad
	x^{(R)} = y .
	\label{eq:dyadic-ladder}
\end{equation}
Each $x^{(r)}$ is thus a record at an intermediate resolution, and each $A_r$ defines an inverse problem of the same form as Equation~\eqref{eq:observation}, with its own compatible set and kernel. Because every level deterministically determines all coarser levels, conditioning on $x^{(r+1)}$ also fixes $x^{(r+2)},\ldots,x^{(R)}$.
The conditional law therefore factorizes along the hierarchy,
\begin{equation}
	p\left(x \mid y\right) = \int \prod_{r=0}^{R-1} p\left(x^{(r)} \mid x^{(r+1)}\right)\, \mathrm{d}x^{(1)} \cdots \mathrm{d}x^{(R-1)}.
	\label{eq:scale-chain}
\end{equation}
This factorization is exact: a single conditional spanning coarsening ratio $k$ is replaced by a sequence of smaller conditional problems of the same form.
\method{} samples these conditionals from coarse to fine, applying Equation~\eqref{eq:level} with $A_r$ when generating $x^{(r-1)}$ from $x^{(r)}$. Every generated level therefore reproduces its parent exactly, and consistency propagates to the original coarse observation. Approximation enters only through the learned conditional factors, not through the factorization itself. We refer to the $R=1$ instance as single-step \method{} and to instances with $R>1$ as recursive \method{}.
In recursive \method{}, each level can also condition on the last state of the preceding window at its own resolution (Appendix~\ref{app:nbr-ablation}). %

\paragraph{Training objective.}
Learning a conditional sampler requires an objective whose population optimum is the target distribution. We use the energy score, the strictly proper scoring rule used in engression \citep{gneiting2007strictly,shenEngressionExtrapolationLens2025}. For an ensemble of $M$ independent draws $u^{(1)}, \dots, u^{(M)}$ and target $v$, its unbiased ensemble estimator is
\begin{equation}
	S_M\!\left(u^{(1:M)},v\right)
	=
	\frac{1}{M}
	\sum_{m=1}^{M}
	\left\lVert u^{(m)}-v\right\rVert
	-
	\frac{1}{2M(M-1)}
	\sum_{\substack{m,m'=1\\m\neq m'}}^{M}
	\left\lVert u^{(m)}-u^{(m')}\right\rVert .
	\label{eq:energy-score}
\end{equation}
Recursive \method{} evaluates the energy score on projected samples, whereas single-step \method{} trains on unprojected samples and applies projection after training (Appendix~\ref{app:esproj}).
Each level is trained against its true parent, so that it learns the conditional factor in \eqref{eq:scale-chain} rather than errors propagated from coarser learned levels.

Although strictly proper, the energy score can have weak discrimination for dependence misspecification in high dimension
\citep{scheuererVariogramBasedProperScoring2015}. In some of our spatial systems, training on the energy score alone overestimates fine-scale variability (Appendix~\ref{app:vgram-ablation}).
Where spatial coherence is not sufficiently captured, we augment the energy score with an auxiliary penalty,
\begin{equation}
	\mathcal{L}
	=
	S_M
	+
	\lambda \mathcal{R}_{\mathrm{spatial}} .
	\label{eq:training-loss}
\end{equation}

One choice is a variogram penalty \citep{scheuererVariogramBasedProperScoring2015}.
Let $\gamma_h^a(x) = \left|x_{a+h}-x_a\right|^p$ denote the order-$p$ pairwise difference at cell $a$ and spatial lag $h$, with $p=\frac 12$. An unbiased $M$-draw estimator is
\begin{equation}
	V_M\!\left(u_{1:M},v\right)
	=
	\sum_h w_h \sum_a
	\frac{1}{M(M-1)}
	\sum_{\substack{m,m'=1\\m\neq m'}}^M
	\left(
	\gamma_h^a(v)-\gamma_h^a(u_m)
	\right)
	\left(
	\gamma_h^a(v)-\gamma_h^a(u_{m'})
	\right) .
	\label{eq:variogram-score}
\end{equation}
We use lags of one, two, and four grid cells along each spatial axis.
A second option is a spectral CRPS penalty, following its use in data-driven weather forecasting \citep{bonevFourCastNet3Geometric2025}. It evaluates probabilistic skill after transforming ensemble members and targets to the spectral domain, thereby penalizing errors in the distribution of spatial scales. We treat both the variogram and spectral terms as auxiliary spatial penalties rather than as part of the definition of \method{}. %
When a spatial penalty is used, $\lambda$ is set initially so that the energy and spatial terms have comparable magnitude.

\paragraph{End-to-end fine-tuning.}
Teacher forcing trains each conditional factor on the true parent, whereas at inference it receives a generated one, so errors can accumulate along the reverse chain \citep{huangTheoreticalAnalysisEngression2026}. Therefore, we optionally fine-tune the complete hierarchy end-to-end by generating full trajectories from the coarse observation and backpropagating the final loss through all levels jointly. This exposes each level to the parent distribution encountered at inference and allows finer levels to compensate for upstream errors.
During this stage, we retain the energy-score objective and replace the variogram penalty with a spectral CRPS term (Appendix~\ref{app:vgram-ablation}). Fine-tuning is used only when validation diagnostics reveal error propagation or degraded calibration in recursive \method{}.

\section{Experiments}
\label{sec:experiments}

The experiments address three questions:
(i) do generated trajectories satisfy the coarse observation;
(ii) when does multiresolution sampling improve over generating the full window in a single step; and
(iii) does the ensemble recover the unresolved conditional variability and fine-scale temporal structure as predictability decreases?

\subsection{Experimental setup}

The protocol is shared across the three tiers of Table~\ref{tab:systems}. Starting from a fine-resolution record, we apply one of the observation operators in \eqref{eq:operators} and reconstruct the withheld trajectory from the resulting coarse record. Interpolation windows hold $N+1$ states with both endpoints $x_0$ and $x_N$ observed. For the two simulated tiers, the fully resolved trajectory is available and predictability can be varied systematically, allowing us to study performance as a function of both the window length $N$ and the intrinsic decorrelation time of the dynamics. ERA5 provides the corresponding archive-based setting. Appendix~\ref{app:experiments} gives equations, splits, and training details.

\begin{table}[ht]
	\centering
	\caption{\textbf{Experimental systems.}
		We evaluate \method{} across three tiers of increasing state dimension and realism.
		Each window of $N$ steps is coarsened to a single value; the factorization lists the coarsenings applied from coarse to fine, one level per factor ($2^6$: six factors of two).}
	\setlength{\tabcolsep}{4.5pt}
	\resizebox{\linewidth}{!}{%
		\begin{tabular}{llcccl}
			\toprule
			Tier & System & State shape & \begin{tabular}[c]{@{}c@{}}Window $N$\\and factorization\end{tabular} & \begin{tabular}[c]{@{}c@{}}Control of\\predictability\end{tabular} & Backbone \\
			\midrule
			\multicolumn{2}{l}{Lorenz-96} & $36 \times 1$ & $64$: $2^6$;\; $128$: $2^7$ & \begin{tabular}[c]{@{}c@{}}forcing\\$F \in \{10,24,32,64\}$\end{tabular} & circular CNN \\
			\midrule
			\multirow{4}{*}{AutoSim} & Advection-diffusion & $64^2 \times 1$ & \multirow{4}{*}{\shortstack[c]{$8$: $2^3$;\; $16$: $2^4$; \\ $24$: $2,2,2,3$; \\ $32$: $2^5$}} & \multirow{4}{*}{\begin{tabular}[c]{@{}c@{}}regime and $N$\\(via $N\delta/\tau$)\end{tabular}} & \multirow{4}{*}{\shortstack[l]{NATTEN\\\citep{hassaniNeighborhoodAttentionTransformer2023}}} \\
			 & Gray--Scott & $64^2 \times 2$ & & & \\
			 & Navier--Stokes & $64^2 \times 3$ & & & \\
			 & Gross--Pitaevskii & $64^2 \times 3$ & & & \\
			\midrule
			\multicolumn{2}{l}{ERA5} & $149 \times 281 \times 7$ & $24$: $2,2,2,3$ & -- & \begin{tabular}[c]{@{}l@{}}ArchesWeather\\\citep{couaironArchesWeatherGenSkillfulComputeefficient2026}\end{tabular} \\
			\bottomrule
		\end{tabular}
	}
	\label{tab:systems}
\end{table}

\paragraph{Methods.}
\label{sec:baselines}

We compare single-step and recursive \method{} with classical and learned baselines. The classical baselines are linear and natural cubic splines for interpolation, and Denton--Cholette and CDF-t for disaggregation. The learned baselines are conditional flow matching \citep{lipmanFlowMatchingGenerative2023}, FlowDPS \citep{kimFlowdpsFlowdrivenPosterior2025}, and, on the two-dimensional systems, SimVP \citep{gaoSimVPSimplerBetter2022} as a deterministic baseline. \method{}, conditional flow matching, and FlowDPS are matched in parameter count on Lorenz-96 and AutoSim, while ERA5 uses architecture-specific configurations (Appendix~\ref{app:baselines}). Separate models are trained for interpolation and disaggregation.
On all three benchmarks, the learned baselines are projected onto the constraint set after sampling using the map in Equation~\eqref{eq:level}.
Under disaggregation, \method{} and both flow-based baselines receive the aggregates of adjacent windows as cross-window context (Appendix~\ref{app:nbr-ablation}).%

\paragraph{Evaluation.}
\label{sec:metrics}

We evaluate separately whether a reconstruction satisfies the observation and whether it recovers the unresolved conditional distribution. A reconstruction is consistent if it preserves the retained states under interpolation and the imposed aggregates under disaggregation. In the tables, \emph{cons.} indicates whether a method is consistent to within $10^{-4}$ in standardized units by construction (\yes{}), by projection after sampling (\yesp{}), or not at all (\no{}). We then assess the learned conditional distribution through the fair CRPS \citep{ferroFairScoresEnsemble2014}, the spread-skill ratio \citep{raspWeatherBench2Benchmark2024}, and the mean absolute error of empirical coverage over nominal central intervals \citep{greenburyReliabilityProbabilisticEmulation2026}. Because these scores are marginal in time, we additionally report the temporal variogram ratio \(t_1\) to measure short-lag temporal structure. Appendix~\ref{app:metrics} defines all metrics. On Lorenz-96 and AutoSim, scores are averaged over three training seeds, with $\pm$ indicating the standard deviation across seeds. ERA5 uses a single seed. Appendix~\ref{app:experiments} details the evaluation protocols.

\subsection{Results}

\begin{figure}[tb]
	\centering
	\includegraphics[width=\linewidth]{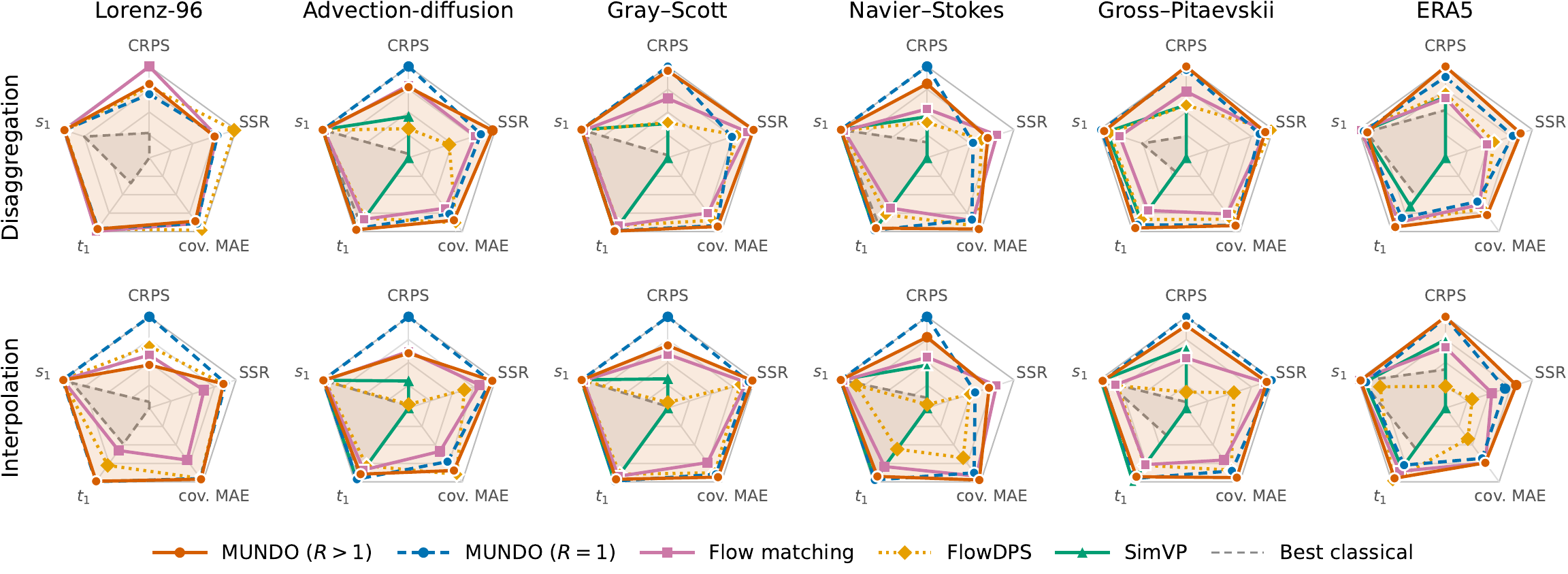}
	\caption{\textbf{Main metrics across all benchmarks}, under disaggregation (top) and interpolation (bottom). Each axis is a score in $(0,1]$, with 1 ideal: a larger polygon is better. CRPS is relative to the best method of each panel; the other axes are absolute. Dashed gray: best classical baseline of each panel, by CRPS. Scores are reported in Tables~\ref{tab:l96}, \ref{tab:era5} and~\ref{tab:autosim-L32-base}.}
	\label{fig:radar}
\end{figure}

Figure~\ref{fig:radar} summarizes the main metrics across all benchmarks. \method{} spans the largest polygon on most panels, and the best classical baseline is far behind all learned methods. Single-step \method{} has the lowest CRPS in most panels, especially under interpolation, whereas recursion improves calibration almost everywhere.

\begin{figure}[tb]
	\centering
	\includegraphics[width=\linewidth]{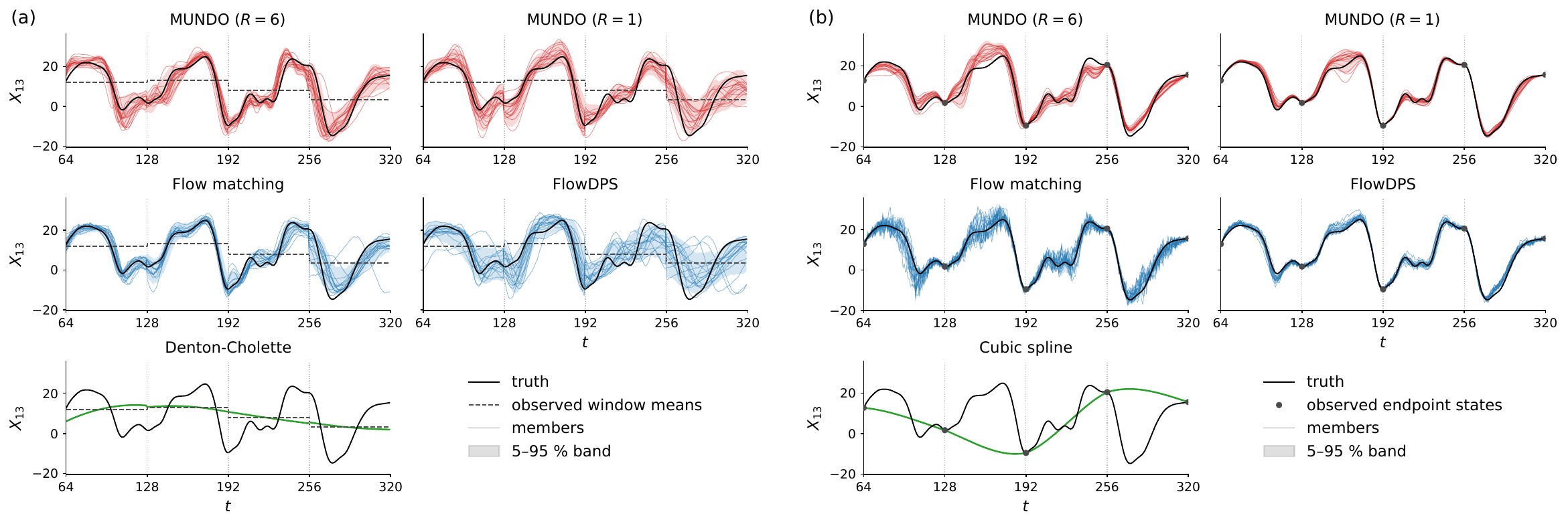}
	\caption{\textbf{Lorenz-96 at $F=32$, $N=64$, slow variable with the largest variance}, over the four windows of Figures~\ref{fig:l96-hov-disagg} and~\ref{fig:l96-hov-interp}.
		(a) Disaggregation, (b) interpolation. Each panel shows the truth, the observation, 16 members (thin lines) and the 5--95\,\% range of the full ensemble (band). The classical baselines show their single reconstruction.}
	\label{fig:l96-site}
\end{figure}

\paragraph{Lorenz-96.}
We use the two-scale Lorenz-96 system \citep{lorenzPredictabilityProblemPartly2006} as a controlled setting in which predictability can be varied directly through the forcing $F$. Table~\ref{tab:l96} reports the representative case at $N=64$, $F=32$, while Appendix~\ref{app:l96-results} reports the remaining forcing and window-length settings.	
Before the final projection, neither flow-based baseline satisfies the observation constraint on Lorenz-96, despite the consistency updates used by FlowDPS, and the projection leaves their CRPS nearly unchanged. Under disaggregation, flow matching attains a lower CRPS than recursive \method{}, with similar calibration. Under interpolation, its CRPS lies between the two \method{} variants, and its members show excessive step-to-step noise, with almost twice the true short-lag variability, whereas recursive \method{} keeps $t_1$ close to one. Across the sweep, single-step \method{} keeps the lower CRPS under interpolation in every chaotic setting, while recursion lowers it under disaggregation in all settings but one. As predictability decreases, the CRPS gap closes and the advantage of recursion shifts to calibration and short-lag structure, which degrade for single-step \method{}.
Figure~\ref{fig:l96-site} shows the ensembles at one slow variable over four consecutive windows. Under disaggregation, the members of recursive \method{} preserve continuity across window boundaries better than those of the other methods. Under interpolation, its ensemble is wider than that of single-step \method{} and covers the truth with more distinct trajectories, while flow matching members are visibly noisy from one step to the next.

\begin{table}[tb]
	\centering
	\setlength{\tabcolsep}{4pt}
	\begin{minipage}[t]{0.505\textwidth}%
		\centering
		\caption{\textbf{Lorenz-96 at $N=64$, $F=32$.} $^{\dagger}$: causal chain. Bold: best; underlined: second best per metric.}%
		\resizebox{\linewidth}{!}{%
			\begin{tabular}{lccccc}
				\toprule
				                    & cons. & CRPS $\downarrow$          & SSR $\to 1$   & cov.\ MAE $\downarrow$ & $t_1 \to 1$ \\
				\midrule
				\multicolumn{6}{l}{\emph{Disaggregation}}                                                                             \\
				\method{} ($R = 6$)$^{\dagger}$ & \yes  & \underline{0.2255}\sd{0.0030} & 0.75 & 0.083 & \underline{0.96} \\
				\method{} ($R = 1$) & \yesp & 0.2618\sd{0.0015} & \underline{0.79} & \underline{0.063} & \underline{0.96} \\
				Flow matching       & \yesp   & \textbf{0.1823}\sd{0.0046} & 0.78 & 0.065 & \textbf{1.00} \\
				FlowDPS             & \yesp   & 0.2341\sd{0.0019} & \textbf{1.02} & \textbf{0.011} & \underline{0.96} \\
				Denton--Cholette    & \yes  & 0.6715 & -- & -- & 0.36 \\
				CDF-t               & \no   & 0.7850 & -- & -- & 0.48 \\
				\midrule
				\multicolumn{6}{l}{\emph{Interpolation}}                                                                              \\
				\method{} ($R = 6$) & \yes  & 0.1112\sd{0.0068} & 0.85 & \textbf{0.021} & \underline{0.99} \\
				\method{} ($R = 1$) & \yesp & \textbf{0.0526}\sd{0.0009} & \textbf{0.88} & \underline{0.026} & \textbf{1.00} \\
				Flow matching       & \yesp   & 0.0906\sd{0.0021} & 1.59 & 0.172 & 1.74 \\
				FlowDPS             & \yesp   & \underline{0.0774}\sd{0.0065} & \underline{0.86} & 0.028 & 1.29 \\
				Linear              & \yes  & 0.7816 & -- & -- & 0.49 \\
				Cubic spline        & \yes  & 0.8177 & -- & -- & 0.50 \\
				\bottomrule
			\end{tabular}%
		}%
		\label{tab:l96}
	\end{minipage}%
	\hfill%
	\begin{minipage}[t]{0.475\textwidth}%
		\centering
		\caption{\textbf{ERA5, daily-to-hourly.} Bold: best; underlined: second best per metric.}%
		\resizebox{\linewidth}{!}{%
			\begin{tabular}{lccccc}
				\toprule
				                    & cons. & CRPS $\downarrow$ & SSR $\to 1$   & cov.\ MAE $\downarrow$ & $t_1 \to 1$ \\
				\midrule
				\multicolumn{6}{l}{\emph{Disaggregation}}                                                                    \\
				\method{} ($R = 4$) & \yes  & \textbf{0.0961}   & \textbf{0.79} & \textbf{0.147}         & \textbf{1.01}     \\
				\method{} ($R = 1$) & \yesp & \underline{0.1089}            & \underline{0.75}          & 0.250                  & 1.19              \\
				Flow matching       & \yesp & 0.1484            & 0.48          & 0.224                  & \underline{0.84}              \\
				FlowDPS             & \yesp & 0.1376            & 0.55          & \underline{0.190}                  & 0.84              \\
				SimVP               & \yesp & 0.1438            & --            & --                     & 0.64              \\
				Denton--Cholette    & \yes  & 0.1841            & --            & --                     & 0.52              \\
				CDF-t               & \no   & 0.2165            & --            & --                     & 0.78              \\
				\midrule
				\multicolumn{6}{l}{\emph{Interpolation}}                                                                     \\
				\method{} ($R = 4$) & \yes  & \textbf{0.0929}   & \textbf{0.81} & \underline{0.151}                  & \underline{1.04}              \\
				\method{} ($R = 1$) & \yesp & \underline{0.0937}            & \underline{0.68}          & 0.186                  & 1.27              \\
				Flow matching       & \yesp & 0.1397            & 0.53          & \textbf{0.150}         & 0.84              \\
				FlowDPS             & \yesp & 0.3918            & 0.31          & 0.342                  & \textbf{1.00}     \\
				SimVP               & \yesp & 0.1231            & --            & --                     & 0.80              \\
				Linear              & \yes  & 0.2204            & --            & --                     & 0.55              \\
				Cubic spline        & \yes  & 0.2211            & --            & --                     & 0.56              \\
				\bottomrule
			\end{tabular}
		}%
		\label{tab:era5}
	\end{minipage}%
\end{table}

\begin{figure}[tb]
	\centering
	\includegraphics[width=\linewidth]{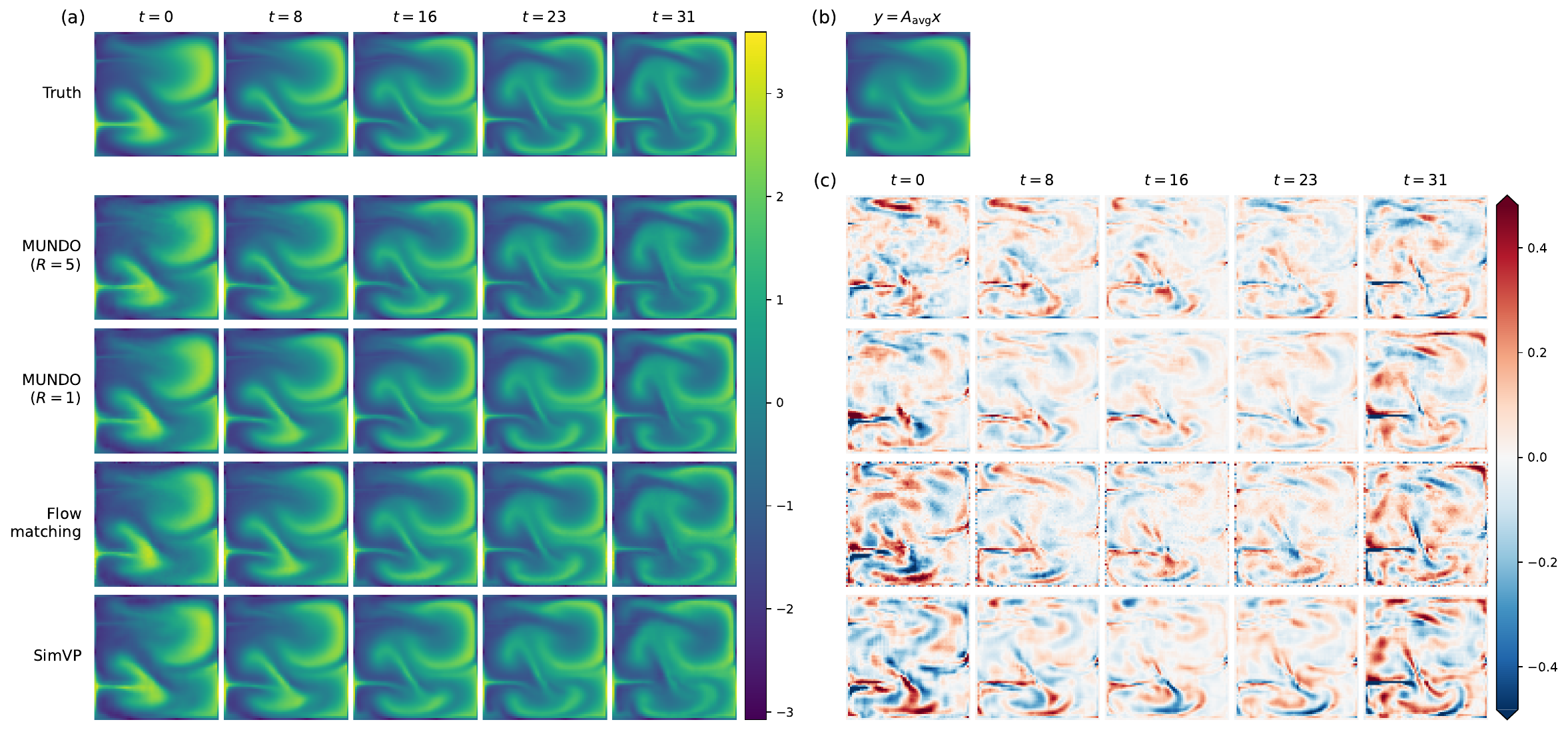}
	\caption{\textbf{Navier--Stokes disaggregation at $N=32$.}
		(a) Ground truth and reconstructions at selected times.
		(b) Coarse observation, given by the temporal mean over the window.
		(c) Reconstruction errors $\hat{x}-x$ at selected times; rows follow the method order in (a).}
	\label{fig:navier_stokes_disagg}
\end{figure}

\begin{figure}[t]
	\centering
	\includegraphics[width=\linewidth]{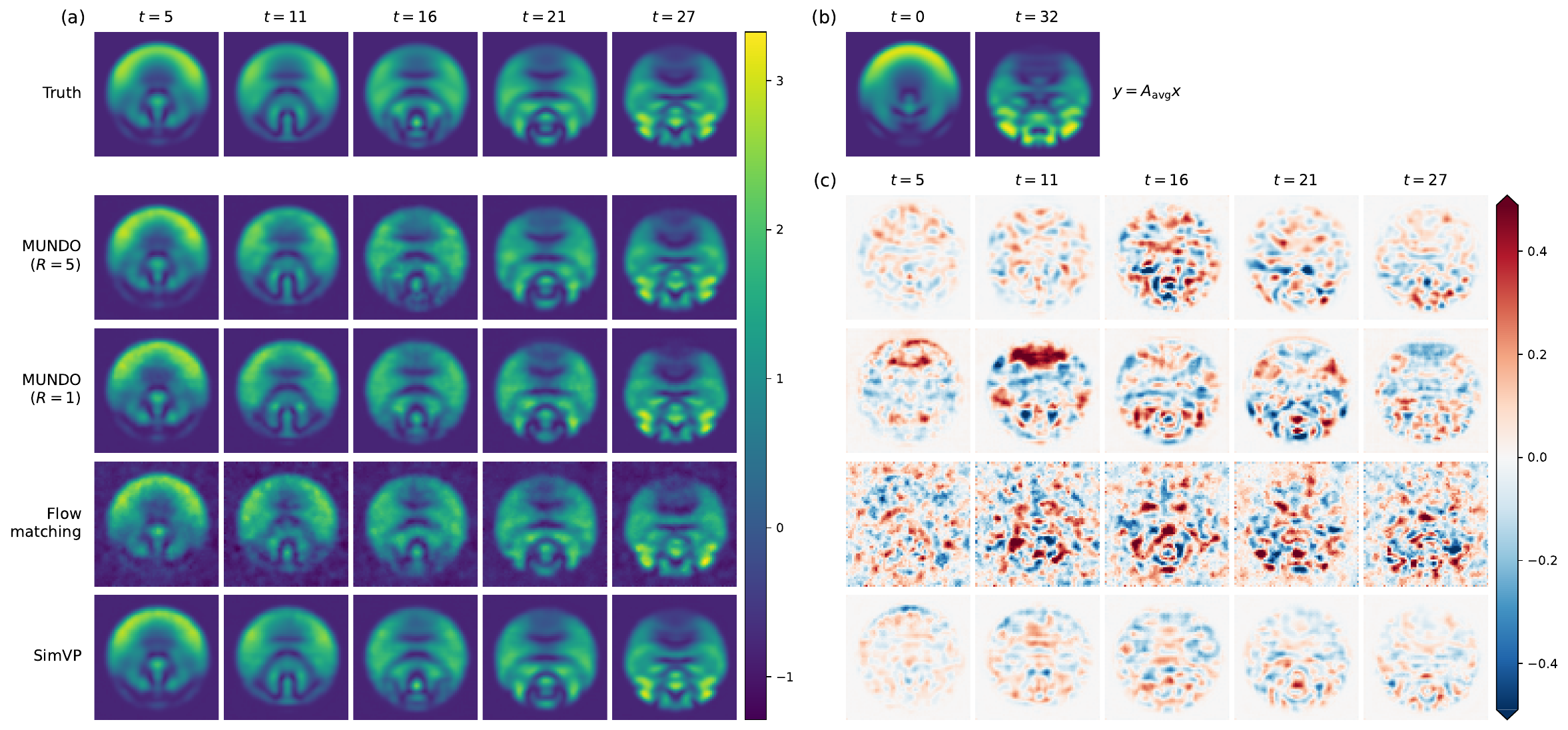}
	\caption{\textbf{Gross--Pitaevskii interpolation at $N=32$.}
		(a) Ground truth and reconstructions at selected times.
		(b) Observed endpoint states.
		(c) Reconstruction errors $\hat{x}-x$ at selected times; rows follow the method order in (a).}    \label{fig:gpe_wake_interp}
\end{figure}

\paragraph{AutoSim.}
We use four systems from AutoSim \citep{greenburyReliabilityProbabilisticEmulation2026}: advection-diffusion as a near-deterministic control, Gray--Scott for multimodal pattern formation, conditioned Navier--Stokes for turbulent dynamics, and Gross--Pitaevskii for dispersive wave dynamics. Table~\ref{tab:autosim-L32-base} reports the benchmark tier at $N=32$, while Appendix~\ref{app:autosim-results} reports the remaining window lengths and predictability regimes.
In Table~\ref{tab:autosim-L32-base}, single-step \method{} already outperforms every baseline in CRPS on all systems and tasks, and remains the strongest variant for interpolation. Both flow-based methods overestimate short-lag variability on every system and task, while $t_1$ is closer to one for \method{}. In the harder, less predictable settings, recursion generally improves ensemble calibration and, where predictability is lowest, lowers CRPS under disaggregation, while retaining realistic fine-scale temporal structure. %
Figures~\ref{fig:navier_stokes_disagg} and~\ref{fig:gpe_wake_interp} illustrate representative reconstructions for disaggregation and interpolation on Navier--Stokes and Gross--Pitaevskii, respectively. Appendix~\ref{app:autosim-results} also shows reconstructions on advection-diffusion and Gray--Scott (Figures~\ref{fig:advection_interp_app} and~\ref{fig:gray_disagg_app}).

\begin{table}[t]
	\centering
	\caption{\textbf{AutoSim at $N=32$.} $^{\ddagger}$: end-to-end fine-tuned. Bold: best; underlined: second best per metric.}
	\setlength{\tabcolsep}{4pt}
	\begin{subtable}[t]{0.49\textwidth}
		\centering
		\caption{\textbf{Advection-diffusion}}%
		\resizebox{\linewidth}{!}{%
			\begin{tabular}{lccccc}
				\toprule
				                    & cons. & CRPS $\downarrow$          & SSR $\to 1$   & cov.\ MAE $\downarrow$ & $t_1 \to 1$ \\
				\midrule
				\multicolumn{6}{l}{\emph{Disaggregation}}                                                                             \\
				\method{} ($R = 5$) & \yes & \underline{0.0026}\sd{0.0000} & \textbf{0.97} & \underline{0.090} & \textbf{1.02} \\
				\method{} ($R = 1$) & \yesp & \textbf{0.0020}\sd{0.0000} & \underline{0.83} & 0.140 & 1.04 \\
				Flow matching & \yesp & \underline{0.0026}\sd{0.0000} & 0.77 & 0.181 & 1.21 \\
				FlowDPS & \yesp & 0.0062\sd{0.0002} & 0.47 & \textbf{0.070} & 1.22 \\
				SimVP & \yesp & 0.0044 & -- & -- & \underline{1.03} \\
				Denton--Cholette & \yes & 0.0382 & -- & -- & 0.89 \\
				CDF-t & \no & 0.0432 & -- & -- & 0.88 \\
				\midrule
				\multicolumn{6}{l}{\emph{Interpolation}}                                                                              \\
				\method{} ($R = 5$) & \yes  & 0.0022\sd{0.0000}          & \textbf{1.04} & \underline{0.090}                  & 1.12              \\
				\method{} ($R = 1$) & \yesp  & \textbf{0.0013}\sd{0.0000} & \underline{0.94}          & 0.160                  & \underline{1.04}              \\
				Flow matching       & \yesp & \underline{0.0021}\sd{0.0000}          & 0.82          & 0.242                  & 1.19              \\
				FlowDPS             & \yesp & 0.0326\sd{0.0024}          & 0.64          & \textbf{0.060}         & 1.28              \\
				SimVP               & \yesp & 0.0044                     & --            & --                     & \textbf{1.03}     \\
				Linear              & \yes  & 0.0452                     & --            & --                     & 0.89 \\
				Cubic spline        & \yes  & 0.0420                     & --            & --                     & 0.92 \\
				\bottomrule
			\end{tabular}}%
		\label{tab:autosim-advection-diffusion-L32-base-sub}
	\end{subtable}\hfill
	\begin{subtable}[t]{0.49\textwidth}
		\centering
		\caption{\textbf{Gray--Scott}}%
		\resizebox{\linewidth}{!}{%
			\begin{tabular}{lccccc}
				\toprule
				                    & cons. & CRPS $\downarrow$          & SSR $\to 1$   & cov.\ MAE $\downarrow$ & $t_1 \to 1$ \\
				\midrule
				\multicolumn{6}{l}{\emph{Disaggregation}}                                                                             \\
				\method{} ($R = 5$) & \yes & \underline{0.0034}\sd{0.0000} & \textbf{1.01} & \textbf{0.040} & \textbf{1.01} \\
				\method{} ($R = 1$) & \yesp & \textbf{0.0032}\sd{0.0000} & 1.35 & \underline{0.048} & \textbf{1.01} \\
				Flow matching & \yesp & 0.0049\sd{0.0001} & \underline{0.93} & 0.144 & 1.09 \\
				FlowDPS & \yesp & 0.0083\sd{0.0002} & 0.80 & 0.089 & 1.09 \\
				SimVP & \yesp & 0.0085 & -- & -- & \textbf{1.01} \\
				Denton--Cholette & \yes & 0.1079 & -- & -- & \underline{0.93} \\
				CDF-t & \no & 0.1440 & -- & -- & 0.90 \\
				\midrule
				\multicolumn{6}{l}{\emph{Interpolation}}                                                                              \\
				\method{} ($R = 5$) & \yes  & \underline{0.0035}\sd{0.0000}          & \textbf{1.03} & \textbf{0.037}         & 1.04              \\
				\method{} ($R = 1$) & \yesp  & \textbf{0.0024}\sd{0.0000} & \underline{0.93}          & \underline{0.060}                  & \underline{1.01}              \\
				Flow matching       & \yesp & 0.0041\sd{0.0001}          & 1.10          & 0.153                  & 1.08              \\
				FlowDPS             & \yesp & 0.0399\sd{0.0008}          & 1.20          & 0.072                  & 1.12              \\
				SimVP               & \yesp & 0.0076                     & --            & --                     & \textbf{1.00}     \\
				Linear              & \yes  & 0.1563                     & --            & --                     & 0.91 \\
				Cubic spline        & \yes  & 0.1361                     & --            & --                     & 0.95 \\
				\bottomrule
			\end{tabular}}%
		\label{tab:autosim-gray-scott-L32-base-sub}
	\end{subtable}\par\medskip
	\begin{subtable}[t]{0.49\textwidth}
		\centering
		\caption{\textbf{Navier--Stokes}}%
		\resizebox{\linewidth}{!}{%
			\begin{tabular}{lccccc}
				\toprule
				                    & cons. & CRPS $\downarrow$          & SSR $\to 1$   & cov.\ MAE $\downarrow$ & $t_1 \to 1$ \\
				\midrule
				\multicolumn{6}{l}{\emph{Disaggregation}}                                                                             \\
				\method{} ($R = 5$) & \yes & \underline{0.0159}\sd{0.0002} & \underline{0.70} & \textbf{0.020} & \underline{1.05} \\
				\method{} ($R = 1$) & \yesp & \textbf{0.0129}\sd{0.0001} & 0.53 & 0.095 & \textbf{1.02} \\
				Flow matching & \yesp & 0.0236\sd{0.0003} & \textbf{1.24} & 0.088 & 1.51 \\
				FlowDPS & \yesp & 0.0334\sd{0.0007} & 0.63 & \underline{0.044} & 1.30 \\
				SimVP & \yesp & 0.0286 & -- & -- & \textbf{1.02} \\
				Denton--Cholette & \yes & 0.0892 & -- & -- & 0.89 \\
				CDF-t & \no & 0.1665 & -- & -- & 0.92 \\
				\midrule
				\multicolumn{6}{l}{\emph{Interpolation}}                                                                              \\
				\method{} ($R = 5$) & \yes  & \underline{0.0117}\sd{0.0000}          & \underline{0.71}          & \textbf{0.015}         & 1.08              \\
				\method{} ($R = 1$) & \yesp  & \textbf{0.0091}\sd{0.0000} & 0.55          & 0.072                  & \underline{1.03}              \\
				Flow matching       & \yesp & 0.0163\sd{0.0003}          & \textbf{1.25} & \underline{0.046}                  & 1.26              \\
				FlowDPS             & \yesp & 0.2365\sd{0.0050}          & 0.49          & 0.191                  & 1.80              \\
				SimVP               & \yesp & 0.0193                     & --            & --                     & \textbf{1.02}     \\
				Linear              & \yes  & 0.1041                     & --            & --                     & 0.92 \\
				Cubic spline        & \yes  & 0.0933                     & --            & --                     & 0.94 \\
				\bottomrule
			\end{tabular}}%
		\label{tab:autosim-conditioned-ns-L32-base-sub}
	\end{subtable}\hfill
	\begin{subtable}[t]{0.49\textwidth}
		\centering
		\caption{\textbf{Gross--Pitaevskii}}%
		\resizebox{\linewidth}{!}{%
			\begin{tabular}{lccccc}
				\toprule
				                                & cons. & CRPS $\downarrow$          & SSR $\to 1$   & cov.\ MAE $\downarrow$ & $t_1 \to 1$ \\
				\midrule
				\multicolumn{6}{l}{\emph{Disaggregation}}                                                                                         \\
				\method{} ($R = 5$)$^{\ddagger}$ & \yes & \textbf{0.1202}\sd{0.0018} & \underline{0.91} & \textbf{0.047} & \textbf{1.05} \\
				\method{} ($R = 1$) & \yesp & \underline{0.1249}\sd{0.0014} & 0.85 & \underline{0.049} & \underline{0.91} \\
				Flow matching & \yesp & 0.1578\sd{0.0009} & 0.90 & 0.138 & 1.39 \\
				FlowDPS & \yesp & 0.2085\sd{0.0018} & \textbf{1.02} & 0.098 & 1.20 \\
				SimVP & \yesp & 0.2071 & -- & -- & 0.85 \\
				Denton--Cholette & \yes & 0.5154 & -- & -- & 0.21 \\
				CDF-t & \no & 0.6528 & -- & -- & 0.38 \\
				\midrule
				\multicolumn{6}{l}{\emph{Interpolation}}                                                                                          \\
				\method{} ($R = 5$)$^{\ddagger}$ & \yes  & \underline{0.0450}\sd{0.0005}          & \underline{0.93}          & \textbf{0.034}         & 1.07              \\
				\method{} ($R = 1$)             & \yesp  & \textbf{0.0406}\sd{0.0004} & \textbf{0.98} & \underline{0.087}                  & \underline{1.05}              \\
				Flow matching                   & \yesp & 0.0743\sd{0.0012}          & 0.91          & 0.173                  & 1.30              \\
				FlowDPS                         & \yesp & 0.2332\sd{0.0059}          & 0.55          & 0.093                  & 1.29              \\
				SimVP                           & \yesp & 0.0612                     & --            & --                     & \textbf{1.00}     \\
				Linear                          & \yes  & 0.6231                     & --            & --                     & 0.36 \\
				Cubic spline                    & \yes  & 0.6519                     & --            & --                     & 0.37 \\
				\bottomrule
			\end{tabular}}%
		\label{tab:autosim-gpe-wake-L32-base-sub}
	\end{subtable}
	\label{tab:autosim-L32-base}
\end{table}

\begin{figure}[t]
	\centering
	\includegraphics[width=\linewidth]{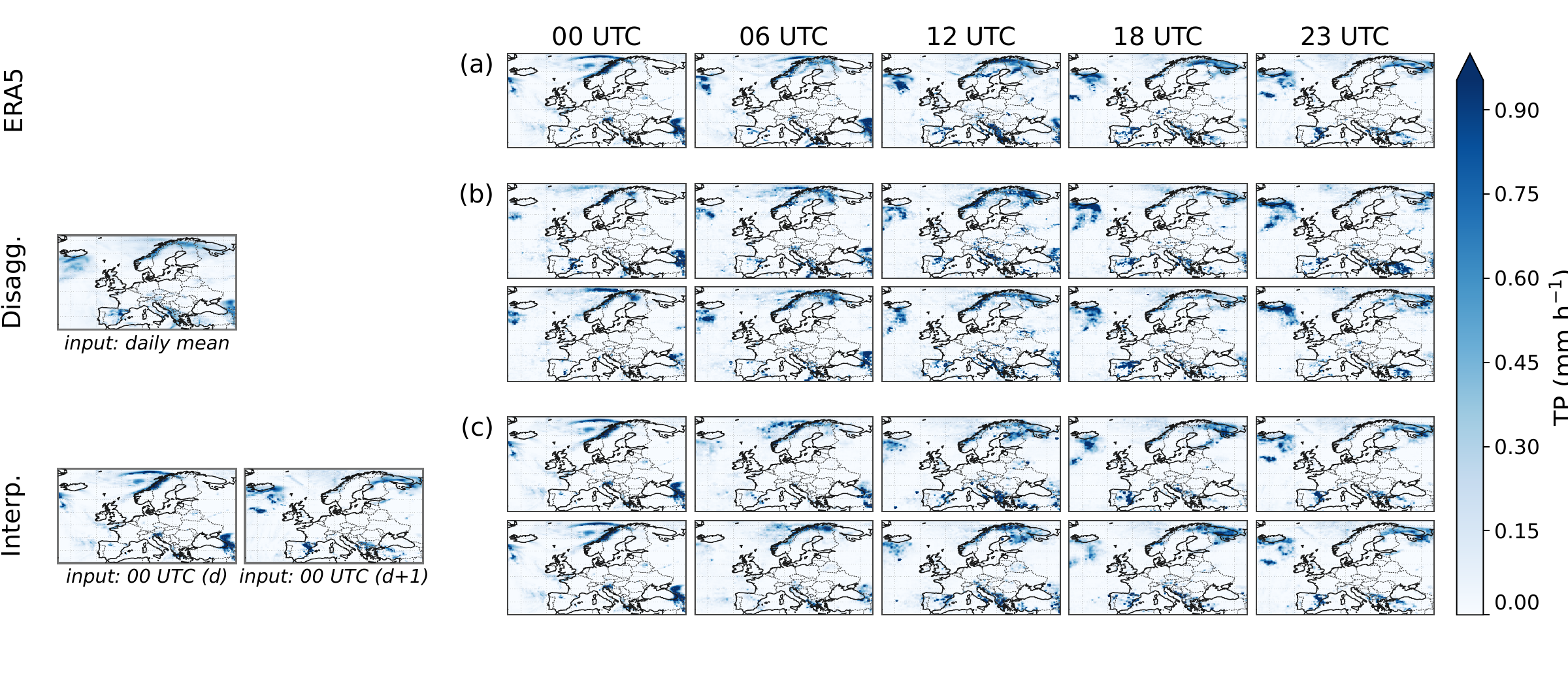}
	\caption{\textbf{ERA5 total precipitation ($\mathrm{mm\,h^{-1}}$) daily-to-hourly reconstructions.} 
		(a) Ground-truth hourly precipitation at selected times.
		(b) Disaggregation from the observed daily mean (leftmost), with two independent recursive \method{} members shown in the two rows.
		(c) Interpolation between the observed 00\,UTC states of consecutive days (leftmost), with the same layout.}%
	\label{fig:era5}
\end{figure}

\paragraph{ERA5.}
We evaluate daily-to-hourly downscaling on ERA5 reanalysis over Europe \citep{hersbachERA5GlobalReanalysis2020}, using seven surface variables on a $149\times281$ grid. Precipitation is non-negative, which the linear constraint does not enforce, so the precipitation samples of every method are projected onto $\{Ax=y,\ x\ge 0\}$ (Appendix~\ref{app:esproj}). %
Table~\ref{tab:era5} shows that both \method{} variants outperform every baseline in CRPS. At the baselines' parameter count, single-step \method{} already lowers CRPS by $21\%$ under disaggregation and $33\%$ under interpolation against the better flow-based baseline, while the flow-based ensembles are strongly under-dispersed. Recursion is most useful for disaggregation, where it lowers CRPS by a further $12\%$, improves spread-skill agreement and coverage, and brings $t_1$ close to one. Under interpolation, the two variants have nearly identical CRPS, but recursion again improves calibration and temporal structure. Appendix~\ref{app:results} reports per-variable scores and additional diagnostics. Figure~\ref{fig:era5} shows representative ERA5 total-precipitation reconstructions under both operators, with distinct members recovering plausible sub-daily structure from the same coarse observation.

\section{Conclusion}
\label{sec:conclusion}

Temporal interpolation and disaggregation are two instances of one inverse problem: recovering a distribution of fine trajectories compatible with a coarse observation. \method{} is a generative framework that addresses both. Each sampler is trained with the energy score and generates a residual in the null space of the observation operator, which enforces the observation exactly. Single-step \method{} samples the fine trajectory directly, while recursive \method{} decomposes a large coarse-to-fine conditional into a sequence of smaller ones sampled in reverse.

We used the same framework in three settings of increasing complexity, Lorenz-96, four simulated physical systems, and ERA5 reanalysis, changing only the backbone. Single-step \method{} already achieves strong predictive skill and generally keeps the lower CRPS under interpolation. Recursion improves calibration while preserving fine-scale temporal structure and, where predictability is lowest, lowers CRPS under disaggregation. On ERA5, with seven coupled fields, recursion improves every metric over single-step \method{} under both operators.

Extending the construction to joint spatiotemporal and more general observation operators, including data assimilation settings, is a natural next step. A second direction is to apply \method{} to learned forecasts and climate-model archives, including settings without paired fine-resolution truth, to generate plausible fine-scale temporal realizations of climate projections.

\section*{AI use statement}

In this work, we used generative AI tools to assist with software implementation, drafting, editing, and figure production. We have not used them to formulate hypotheses, develop the conceptual framework, design the experiments or evaluation, generate or preprocess data, or interpret the results. The remaining tasks requiring disclosure are not applicable here. All AI-assisted code was reviewed by the authors, and all reported numbers come from our experimental records. We take responsibility for the final content of this work, including text, claims or artifacts produced with the aid of generative AI.

\section*{Reproducibility statement}

Code is available at \url{https://github.com/dataymeric/mundo}.
The AutoSim systems are generated with \url{https://github.com/alan-turing-institute/autosim}. ERA5 is available from the Copernicus Climate Data Store at \url{https://doi.org/10.24381/cds.adbb2d47}. Appendix~\ref{app:experiments} details the datasets, training and evaluation.

\section*{Acknowledgments}

The authors thank Graham Clyne for insightful discussions. This work was supported by the French National Research Agency (ANR) and the Ministry of Higher Education and Research. This project was granted HPC resources by GENCI at IDRIS under the allocation AD011016443 on the Jean-Zay supercomputer. The authors declare no competing interests.

\bibliographystyle{iclr2027_conference}
\renewcommand{\bibfont}{\small}
\bibliography{biblio}

\newpage
\appendix
\raggedbottom

\etocdepthtag.toc{appendix}
\etocsettagdepth{main}{none}
\etocsettagdepth{appendix}{subsection}
\etocsettocstyle{\section*{Appendix}}{}
\tableofcontents

\section{\texorpdfstring{Extended related work}{Extended related work}}
\label{app:related}
\begin{table}[ht]
    \centering
    \setlength{\tabcolsep}{3pt}
    \caption{\textbf{Positioning of temporal downscaling methods.} Interp.\ and Disagg.: the method applies to interpolation and disaggregation, respectively. Prob.: returns draws of $p(x \mid y)$. Exact: the observation holds by construction (\yes), approximately through a penalty, a learned target or guidance (\soft), or not at all (\no). Solver-free: inference needs no iterative ODE/SDE solver. \na: not applicable. $^{\dagger}$Among our baselines. In our experiments, every learned baseline is projected onto the constraint set after sampling (Section~\ref{sec:baselines}).}
    \label{tab:positioning}
    \begin{tabularx}{\textwidth}{@{}>{\raggedright\arraybackslash\hangindent=0.8em}p{5cm} c c c c c >{\raggedright\arraybackslash}X@{}}
        \toprule
        Method                                                                       & Interp. & Disagg. & Prob. & Exact & Solver-free & Scope                              \\
        \midrule
        \multicolumn{7}{@{}l}{\textit{Classical}}                                                                                                                           \\
        Linear / cubic spline$^{\dagger}$                                            & \yes    & \no     & \no   & \yes  & \na         &                                    \\
        Denton--Cholette$^{\dagger}$ \citep{denton1971,cholette1984}                 & \yes    & \yes    & \no   & \yes  & \na         & smoothest admissible record        \\
        CDF-t$^{\dagger}$ \citep{michelangeliProbabilisticDownscalingApproaches2009} & \na     & \yes    & \no   & \no   & \na         & marginals only                     \\
        Stochastic cascade \citep{lombardo2017}                                      & \no     & \yes    & \yes  & \yes  & \na         & non-negative intermittent variable \\
        \midrule
        \multicolumn{7}{@{}l}{\textit{Learned, deterministic}}                                                                                                              \\
        Super SloMo \citep{jiangSuperSloMoHigh2018}                                  & \yes    & \no     & \no   & \yes  & \yes        & RGB video                          \\
        SimVP$^{\dagger}$ \citep{gaoSimVPSimplerBetter2022}                          & \yes    & \yes    & \no   & \no   & \yes        & video prediction                   \\
        STint \citep{harilalSTintSelfsupervisedTemporal2023a}                        & \yes    & \no     & \no   & \yes  & \yes        & pointwise loss                     \\
        ModAFNO \citep{leinonenModulatedAdaptiveFourier2024}                         & \yes    & \no     & \no   & \yes  & \yes        & pointwise loss                     \\
        TemDeep \citep{wangTemDeepSelfsupervisedFramework2025}                       & \yes    & \no     & \no   & \yes  & \yes        & pointwise loss                     \\
        \midrule
        \multicolumn{7}{@{}l}{\textit{Learned, generative}}                                                                                                                 \\
        RainDisaggGAN \citep{scher2021temporal}                                      & \no     & \yes    & \yes  & \yes  & \yes        & non-negative variable              \\
        \citet{harderHardConstrainedDeepLearning2023}                                & \na     & \na     & \yes  & \na   & \yes        & spatial block mean                 \\
        Cond.\ flow matching$^{\dagger}$ \citep{lipmanFlowMatchingGenerative2023}    & \yes    & \yes    & \yes  & \no   & \no         & conditional model                  \\
        DiffESM \citep{bassetti2024}                                                 & \no     & \yes    & \yes  & \no   & \no         & conditioning only                  \\
        HourGlass \citep{ingstadHourGlassProbabilisticDatadriven2026}                & \yes    & \no     & \yes  & \soft & \yes        & 6\,h to 1\,h only                  \\
        EnScale \citep{schillingerEnScaleTemporallyConsistentMultivariate2026}       & \na     & \na     & \yes  & \no   & \yes        & spatial, GCM to RCM                \\
        \midrule
        \multicolumn{7}{@{}l}{\textit{Inverse-problem solvers}}                                                                                                             \\
        DDNM \citep{wang2022zero}                                                    & \yes    & \yes    & \yes  & \yes  & \no         & unconditional prior                \\
        FlowDPS$^{\dagger}$ \citep{kimFlowdpsFlowdrivenPosterior2025}                & \yes    & \yes    & \yes  & \soft & \no         & unconditional prior                \\
        \citet{schmidtGenerativeFrameworkProbabilistic2025}                          & \yes    & \yes    & \yes  & \soft & \no         & unconditional prior                \\
        WIND \citep{aichWINDWeatherInverse2026a}                                     & \yes    & \yes    & \yes  & \soft & \no         & pretrained, $1.5^\circ$, 6-hourly  \\
        \midrule
        \multicolumn{7}{@{}l}{\textit{Ours}}                                                                                                                                \\
        \method{}, single-step                                                       & \yes    & \yes    & \yes  & \yes  & \yes        &                                    \\
        \method{}, recursive                                                         & \yes    & \yes    & \yes  & \yes  & \yes        &                                    \\
        \bottomrule
    \end{tabularx}
\end{table}

This appendix extends Section~\ref{sec:related}. Table~\ref{tab:positioning} compares the methods by the observation operator they address, whether they return draws of $p(x \mid y)$, whether these draws satisfy the observation, and whether inference needs an iterative solver.

\subsection{Statistical downscaling}
\label{app:classical}

Dynamical downscaling runs a high-resolution regional model driven by the coarse data. At convection-permitting resolution, its cost limits it to short periods or small domains \citep{maraunPrecipitationDownscalingClimate2010}. Statistical downscaling learns relations between coarse and fine records instead. All methods below are statistical.

\paragraph{Deterministic profiles.}
MELODIST \citep{forster2016open} distributes daily temperature over the day with a cosine whose amplitude is set by the daily minimum and maximum, and daily wind speed with a site-calibrated cosine scaled by the daily mean. \citet{han2023comparison} compare such deterministic models for wind speed. There, the fixed profile neglects sub-daily fluctuations, so the predicted standard deviation is too low \citep{forster2016open}. These methods return a single record and, for temperature, use the daily minimum and maximum instead of the mean.

\paragraph{Constrained optimization.}
Benchmarking methods from econometrics adjust a high-frequency series to low-frequency totals under the aggregation constraint \citep{denton1971,cholette1984,BenchmarkingTemporalDistribution2006}. Without an indicator series, Denton--Cholette returns the record with the smallest sum of squared first differences whose aggregates match the observation, and on a subsampled series the same criterion reduces to linear interpolation. The regression method of \citet{chow1971best} requires a related series observed at the target resolution, which none of our benchmarks provides.

\paragraph{Distributional corrections.}
Quantile mapping corrects each value through the quantile--quantile relation between the simulated and the target distributions \citep{deque2007}, and CDF-t \citep{michelangeliProbabilisticDownscalingApproaches2009} extends it by letting the target distribution evolve with the simulated one. Applied position by position, both fix the marginals and constrain neither the chronology nor the aggregates. The two usual fixes for a reconstruction that lacks variance do not recover the conditional-variance term of Equation~\eqref{eq:variance-deficit}: inflation only amplifies the variability already explained by $y$ \citep{storchUseInflationStatistical1999}, and randomization represents the variability left unresolved by $y$ with noise whose temporal structure is prescribed, not learned.

\paragraph{Weather generators and resampling.}
Stochastic weather generators simulate daily series from fitted distributions, with Markov-chain or weather-type occurrence models \citep{wilks2010,richardson1981,ailliot2015}, or resample the archive \citep{buishand2001}. The method of fragments disaggregates a day using the sub-daily fractions of an archived day with similar daily values, which preserves the daily total by construction \citep{li2018}. Teddy \citep{zabel2023} applies the same principle to seven variables. In both, the sub-daily shapes come from the archive.

\paragraph{Random cascades.}
Random cascades disaggregate rainfall by splitting each interval into sub-intervals with random weights, level by level. Microcanonical cascades conserve the volume exactly at each split \citep{olsson1998}, canonical ones only on average \citep{schertzerPhysicalModelingAnalysis1987}, and the non-multiplicative cascade of \citet{lombardo2017} restores the totals by an adjustment step. Cascades model one non-negative, intermittent variable with a prescribed parametric model, whereas the levels of \method{} learn the distribution of the children given the parent, which they reproduce exactly.

\subsection{Neural downscaling}
\label{app:neural}

\paragraph{Deterministic interpolation.}
Video frame interpolation reconstructs the frames between two RGB frames, with optical flow \citep{jiangSuperSloMoHigh2018,huangRealTimeIntermediateFlow2022} or without \citep{kalluriFLAVRFlowAgnosticVideo2023}. Geoscientific adaptations include STint \citep{harilalSTintSelfsupervisedTemporal2023a}, pretrained by cycle consistency, TemDeep \citep{wangTemDeepSelfsupervisedFramework2025}, which conditions on a continuous time offset, and ModAFNO \citep{leinonenModulatedAdaptiveFourier2024}, which interpolates six-hourly ERA5 states to hourly. Their pointwise losses pull the output toward a central value of $p(x \mid y)$ (Section~\ref{variance-deficit}), and they take bounding states as observation, so none handles aggregates. Our deterministic baseline is SimVP \citep{gaoSimVPSimplerBetter2022}, a video prediction model.

\paragraph{Generative models.}
Pointwise fidelity and distributional realism trade off against each other \citep{blau2018perception}, which motivates generative downscaling. It was first developed with adversarial networks \citep{stengelAdversarialSuperresolutionClimatological2020,leinonen2020,harrisGenerativeDeepLearning2022}, then with diffusion \citep{saharia2022image} and flow matching \citep{lipmanFlowMatchingGenerative2023}. In space, EnScale \citep{schillingerEnScaleTemporallyConsistentMultivariate2026} trains a generator with injected noise on the energy score to emulate a regional climate model, in coarse-to-fine stages that condition on block-averaged targets without imposing them. In time, DiffESM \citep{bassetti2024} downscales Earth system model output from monthly to daily, conditioning on the coarse record without imposing it, with an iterative solve per draw. HourGlass \citep{ingstadHourGlassProbabilisticDatadriven2026}, concurrent with this work, trains a single-pass generator with the almost-fair CRPS to produce the five hourly states between two six-hourly forecast states, together with the final state. For prognostic variables, the final state has a zero residual target, so the observation is learned, not imposed, and penalties on window statistics promote temporal consistency. It addresses six-hourly to hourly interpolation only. Our baseline from this family is conditional flow matching.

\paragraph{Hard-constrained networks.}
RainDisaggGAN \citep{scher2021temporal} generates hourly fractions of the daily precipitation total with a softmax over time and multiplies them by the total, so the daily sum is exact. The construction requires a non-negative variable and an aggregate observation. \citet{harderHardConstrainedDeepLearning2023} append constraint layers (additive, multiplicative or softmax) that make spatial super-resolution conserve the coarse block means. For block averaging, the additive layer is the projection of Equation~\eqref{eq:level}. Their spatio-temporal model interpolates one intermediate time step with optical flow, under a squared loss, and its constraints act on spatial means only.

\paragraph{Inverse-problem solvers.}
Inverse-problem solvers train an unconditional prior and enforce the observation during sampling. DDNM \citep{wang2022zero} replaces the range-space component of the denoised estimate at each step with the one fixed by the observation, which is exact for a noise-free linear operator. FlowDPS \citep{kimFlowdpsFlowdrivenPosterior2025} adds likelihood-gradient steps to flow-matching sampling. \citet{schmidtGenerativeFrameworkProbabilistic2025} train an unconditional score-based prior on reanalysis. To generate hourly ensembles, they condition it on coarse six-hourly climate simulations through an observation model, following score-based data assimilation. WIND \citep{aichWINDWeatherInverse2026a} trains one video diffusion prior on ERA5 at $1.5^\circ$ and six-hourly steps over 70 variables, and addresses forecasting, downscaling and reconstruction by guidance without fine-tuning. In its own daily-to-six-hourly disaggregation experiment, a task-specific model reaches a lower RMSE. Its released checkpoint does not match our ERA5 setting ($0.25^\circ$, hourly), and retraining the prior is beyond our budget. All these solvers require an iterative solve per draw. Our baseline from this family is FlowDPS, with a prior trained on our data and its samples projected onto the constraint set.

\section{Methodological details}
\label{app:derivations}

\subsection{Scoring rules and the engression objective}
\label{app:scoring}

A scoring rule $S(F,v)$ evaluates a predictive distribution $F \in \mathcal{P}$ against an observation $v \in \mathbb{R}^{D}$. Using the loss convention of Equation~\eqref{eq:energy-score}, it is \emph{proper} when
\begin{equation}
	\mathbb{E}_{v \sim Q}\bigl[S(F,v)\bigr]
	\ge
	\mathbb{E}_{v \sim Q}\bigl[S(Q,v)\bigr]
\end{equation}
for all $F,Q \in \mathcal{P}$, and \emph{strictly proper} when equality implies $F=Q$ \citep{gneiting2007strictly}. Thus the true distribution uniquely minimizes the population score.

The energy score with exponent $\beta \in (0,2)$ is
\begin{equation}
	S(P,v)
	=
	\mathbb{E}\left\lVert U-v \right\rVert_2^\beta
	-
	\frac{1}{2}\,
	\mathbb{E}\left\lVert U-U' \right\rVert_2^\beta,
	\label{eq:energy-score-general}
\end{equation}
where $U,U' \stackrel{\mathrm{iid}}{\sim} P$. It is strictly proper on distributions with finite $\beta$-th moment. Following engression, we take $\beta=1$ \citep{gneiting2007strictly,shenEngressionExtrapolationLens2025}.

Applied conditionally, strict propriety implies, for $p_Y$-almost every $y$,
\begin{equation}
	\mathbb{E}_{X \sim p^{*}(\cdot \mid y)}
	\left[
		S\left(p^{*}(\cdot \mid y),X\right)
		\right]
	\le
	\mathbb{E}_{X \sim p^{*}(\cdot \mid y)}
	\left[
		S\left(p_g(\cdot \mid y),X\right)
		\right],
\end{equation}
with equality only when $p_g(\cdot \mid y)=p^{*}(\cdot \mid y)$. Averaging over $Y$ gives the engression objective \citep{shenEngressionExtrapolationLens2025},
\begin{equation}
	g^{*}
	\in
	\argmin_{g \in \mathcal{G}}
	\mathbb{E}\left[
		\left\lVert X-g(Y,\varepsilon)\right\rVert_2
		-
		\frac{1}{2}
		\left\lVert g(Y,\varepsilon)-g(Y,\varepsilon')\right\rVert_2
		\right],
	\label{eq:engression}
\end{equation}
where $\varepsilon$ and $\varepsilon'$ are independent noise draws. Under realizability, its population optimum satisfies $g^{*}(Y,\varepsilon)\mid Y=y \sim p^{*}(\cdot \mid y)$ for $p_Y$-almost every $y$. No density evaluation is required, which allows the same objective to be used at every level of \eqref{eq:scale-chain}.

\paragraph{Limitations.}
\label{app:coherence}

The energy score is strictly proper, but its finite-sample discrimination of spatial dependence can be weak in high dimension \citep{scheuererVariogramBasedProperScoring2015}. Both terms of \eqref{eq:energy-score} depend only on Euclidean distances over the full field, so different spatial arrangements can receive similar scores even when their local or spectral structure differs. Strict propriety guarantees the correct population optimum over unrestricted distributions, but not that a restricted model will recover spatial dependence accurately under finite-sample optimization. %

\paragraph{Finite-ensemble estimator.}
The ensemble estimators used during training must also avoid penalizing spread when estimated from a small number of members. With $M=2$, the fair pairwise term in the energy score reduces to $\tfrac{1}{2}\lVert u-u'\rVert$, so Equation~\eqref{eq:energy-score} is already unbiased for the corresponding population quantity. The variogram estimator in \eqref{eq:variogram-score} uses the same cross-member construction. By contrast, the naive squared error of the ensemble-mean transformed increment is biased upward by $\tfrac{1}{2}\operatorname{Var}[\gamma_h]$, which implicitly penalizes spread. Using independent draws in the cross-member product makes the corresponding centered errors uncorrelated, so this additional variance term vanishes in expectation.

\subsection{Spatial-coherence terms}
\label{app:vgram-ablation}

We ablate the auxiliary spatial-coherence penalty term in \eqref{eq:training-loss}. During level-wise training, we compare the energy score alone and augmented with either a variogram score or spectral CRPS. During end-to-end fine-tuning, we compare the same three objectives. Single-step \method{} has no fine-tuning stage, since its only parent is the observation itself. The spectral CRPS is computed independently on the real and imaginary parts of each two-dimensional Fourier coefficient \citep{bonevFourCastNet3Geometric2025}.

Table~\ref{tab:texture-ablation} reports the comparison on Gross--Pitaevskii disaggregation at $N=24$. Because the ablation targets recovery of fine spatial scales, we additionally report the finest-band spectral power ratio, denoted Finest in the table, alongside the one-cell spatial variogram ratio $s_1$, both defined in Appendix~\ref{app:metrics}. For single-step \method{}, the variogram term is important: without it, fine-scale spectral power is strongly overestimated and $s_1$ deteriorates. For recursive \method{}, the level-wise variogram term has a smaller effect, while end-to-end fine-tuning with spectral CRPS brings the finest-band power close to the truth regardless of the level-wise objective. %

\begin{table}[t]
	\centering
	\caption{\textbf{Spatial-coherence term ablation.} Gross--Pitaevskii disaggregation at $N=24$, three seeds, with cross-window conditioning. Energy denotes the energy score, variogram the variogram score in \eqref{eq:variogram-score}, and spectral the spectral CRPS. Stage 1 is the level-wise training objective and Stage 2 the end-to-end fine-tuning objective; -- denotes no fine-tuning. $^{\dagger}$ marks the configuration used in the main experiments. SSR is the spread-skill ratio, $s_1$ is the one-cell spatial variogram ratio, Finest the finest-band spectral power ratio; all targeting one.}%
	\resizebox{\linewidth}{!}{%
		\begin{tabular}{lllcccc}
			\toprule
			Arm & Stage 1                        & Stage 2                       & CRPS $\downarrow$          & SSR $\to 1$   & $s_1 \to 1$    & Finest $\to 1$         \\
			\midrule
			\multirow{8}{*}{\shortstack{\method{}                                                                                                                       \\($R=4$)}}   & energy                         & --                            & 0.1047\sd{0.0004}          & \textbf{0.85} & 1.131          & 2.15\sd{0.06}          \\
			    & energy + variogram             & --                            & \textbf{0.1025}\sd{0.0002} & \textbf{0.85} & 1.073          & 2.78\sd{0.30}          \\
			    & energy + spectral              & --                            & \textbf{0.1025}\sd{0.0005} & 0.83          & \textbf{1.055} & \textbf{1.96}\sd{0.10} \\
			\cmidrule(lr){2-7}
			    & energy + variogram             & energy                        & 0.0802\sd{0.0008}          & 0.92          & 1.089          & 2.75\sd{0.19}          \\
			    & energy + variogram             & energy + variogram            & \textbf{0.0788}\sd{0.0007} & 0.92          & \textbf{1.057} & 2.21\sd{0.19}          \\
			    & energy + variogram             & energy + spectral$^{\dagger}$ & 0.0799\sd{0.0010}          & 0.92          & \textbf{1.057} & \textbf{1.17}\sd{0.03} \\
			\cmidrule(lr){2-7}
			    & energy                         & energy + spectral             & 0.0801\sd{0.0000}          & 0.92          & 1.083          & 1.21\sd{0.05}          \\
			    & energy + spectral              & energy + spectral             & \textbf{0.0795}\sd{0.0003} & \textbf{0.93} & \textbf{1.062} & \textbf{1.12}\sd{0.04} \\
			\midrule
			\multirow{2}{*}{\shortstack{\method{}                                                                                                                       \\($R=1$)}} & energy                         & --                            & 0.0866\sd{0.0005}          & 0.83          & 1.135          & 2.97\sd{0.35}          \\
			    & energy + variogram$^{\dagger}$ & --                            & \textbf{0.0828}\sd{0.0009} & 0.83          & \textbf{1.026} & \textbf{1.08}\sd{0.08} \\
			\bottomrule
		\end{tabular}}%
	\label{tab:texture-ablation}
\end{table}

\subsection{Projection during training and sampling}
\label{app:esproj}
Equation~\eqref{eq:level} defines the affine projection $P_y(u)=A^{\dagger}y+(I-A^{\dagger}A)u$ onto $\mathcal{X}_y$. Recursive \method{} evaluates the training loss on projected samples at every level. Single-step \method{} trains on the unprojected reconstruction $A^{\dagger}y+G_\theta(y,\varepsilon,c)$ and projects the generated samples with $P_y$ after training.

Assuming finite first moments and realizability, the expected energy score is uniquely minimized when the distribution evaluated by the score equals the target conditional law (Appendix~\ref{app:scoring}). When projection is applied inside the loss, this guarantee applies to the distribution after projection. When the loss is evaluated on unprojected samples, the optimal unprojected distribution already equals the target law and is unchanged by projection. Both formulations therefore recover the target law at the population optimum. For a finite network, however, the choice of training objective can affect the generated distribution.

Table~\ref{tab:esproj} compares the two training choices for single-step \method{} on AutoSim disaggregation in the benchmark regime at $N=32$. Applying projection inside the loss increases CRPS on all four systems, while its effects on calibration are mixed and temporal structure changes comparatively little. Single-step \method{} retains lower CRPS than the baselines on all four systems regardless of when projection is applied. %

\begin{table}[t]
	\centering
	\caption{\textbf{Effect of projection during training or sampling.} AutoSim disaggregation, benchmark tier, $N=32$.}%
	\resizebox{\linewidth}{!}{%
		\begin{tabular}{lcccccccc}
			\toprule
			                    & \multicolumn{4}{c}{sampling only} & \multicolumn{4}{c}{inside the training loss}                                                                                                                 \\
			\cmidrule(lr){2-5} \cmidrule(lr){6-9}
			System              & CRPS $\downarrow$                 & SSR $\to 1$                                  & cov.\ MAE $\downarrow$ & $t_1 \to 1$ & CRPS $\downarrow$ & SSR $\to 1$ & cov.\ MAE $\downarrow$ & $t_1 \to 1$ \\
			\midrule
			Advection-diffusion & 0.0020                            & 0.83                                         & 0.140                  & 1.04        & 0.0023            & 0.95        & 0.135                  & 1.01        \\
			Gray--Scott         & 0.0032                            & 1.35                                         & 0.048                  & 1.01        & 0.0034            & 1.76        & 0.058                  & 1.00        \\
			Navier--Stokes      & 0.0129                            & 0.53                                         & 0.095                  & 1.02        & 0.0131            & 0.57        & 0.064                  & 1.02        \\
			Gross--Pitaevskii   & 0.1249                            & 0.85                                         & 0.049                  & 0.91        & 0.1262            & 0.97        & 0.066                  & 0.97        \\
			\bottomrule
		\end{tabular}}%
	\label{tab:esproj}
\end{table}

	\paragraph{Non-negative variables.} The projection $P_y$ enforces $Ax=y$ but not the sign of a variable. On ERA5 total precipitation, 28\% (recursive) and 24\% (single-step) of the generated pixel-hours of \method{} are negative, against 17\% for Denton--Cholette and 20\% for flow matching. For this variable we replace $P_y$ by the Euclidean projection onto $\mathcal{X}_y \cap \{x \ge 0\}$. Under disaggregation, the constraint couples the $N$ hours of a pixel through their daily mean $\bar y \ge 0$, and the projection of a block $u_{1:N}$ is
	\begin{equation}
		x_t = \max(u_t - \lambda,\, 0),
		\label{eq:posproj}
	\end{equation}
	where $\lambda$ satisfies $\frac{1}{N}\sum_{t=1}^{N} \max(u_t - \lambda, 0) = \bar y$ and is found by bisection. A block with no negative value is left unchanged ($\lambda = 0$). Under interpolation, the generated hours are not coupled by the constraint and the retained states are non-negative, so the projection reduces to clipping at zero. We apply it after sampling to every method. Alternatively, non-negative variables could be generated in $\log(x+\varepsilon)$ and rescaled to the observed aggregate, which avoids negative values at the source.

\subsection{Number of levels}
\label{app:r-ablation}

Once the coarsening ratio $k$ is fixed, the number of levels $R$ determines how the corresponding prime factors are distributed across the hierarchy. Throughout the paper, we use the finest factorization, with one level per prime factor. Coarser factorizations merge multiple factors into a single step, whose position in the hierarchy can vary. Table~\ref{tab:r-ablation} investigates the effect of $R$ on Navier--Stokes disaggregation at $N=32$, considering configurations ranging from single-step \method{} ($R=1$) to the finest factorization ($R=5$), and all possible factor orderings for $R=2,3,4$. The backbone width is adjusted across configurations to keep parameter counts within $10\%$ of the $R=5$ model.

\begin{figure}[t]
	\centering
	\includegraphics[width=0.55\linewidth]{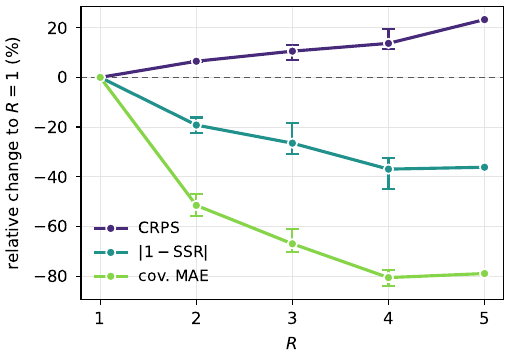}
	\caption{\textbf{Effect of the number of levels $R$.} Navier--Stokes disaggregation, $N=32$. Each curve is the change of a score relative to single-step \method{} ($R=1$), in percent. For all three scores, lower is better, so negative values are improvements: $|1-\mathrm{SSR}|$ is the distance of the spread-skill ratio from one, and cov.\ MAE the coverage error. For each $R$, the marker is the mean over the chains of Table~\ref{tab:r-ablation}, and the bar spans their minimum and maximum; $R=1$ and $R=5$ have a single chain.}
	\label{fig:r-ablation}
\end{figure}

Increasing the number of levels improves calibration at the expense of CRPS (Figure~\ref{fig:r-ablation}). As $R$ increases from 1 to 5, CRPS rises from $0.0129$ to $0.0159$ ($+23\%$), while SSR increases from $0.53$ to $0.70$ and the coverage error decreases from $0.095$ to $0.020$. Notably, $R=4$ already achieves calibration performance comparable to $R=5$: all three configurations whose coarsest level has factor $2$ match its SSR and coverage error, while yielding a $9\%$ lower CRPS. Across all depths, $t_1$ remains stable, ranging from $1.02$ to $1.05$.

Beyond the number of levels, factor ordering also affects performance, primarily through the coarsest level. For $R=3$ and $R=4$, configurations with a coarsest factor of $4$ exhibit $4$--$7\%$ higher CRPS than other orderings at the same depth, alongside higher SSR and lower coverage error. By contrast, the ordering of factors within the finer levels has little effect: at $R=4$, moving the factor $4$ among these levels changes CRPS by less than $1\%$, and the two possible orderings at $R=2$ yield similarly small differences. Thus, intermediate-depth configurations introduce an ordering choice that can affect CRPS by up to $7\%$ at $R=4$. The finest factorization, which merges no factors, avoids this ambiguity, which motivates our choice to use it throughout the paper.

\begin{table}[t]
	\centering
	\caption{\textbf{Level configurations and factor orderings.} Navier--Stokes disaggregation at $N=32$, averaged over three seeds. For CRPS, the standard deviation across seeds is also reported. Each chain lists the level factors in coarsest-to-finest order. The parameter count (Params) is the total across all levels.}
	\label{tab:r-ablation}
	\small
	\begin{tabular}{clcccccc}
		\toprule
		$R$ & Chain       & Width & Params  & CRPS $\downarrow$          & SSR $\to 1$   & cov.\ MAE $\downarrow$ & $t_1 \to 1$   \\
		\midrule
		1   & [32]        & 85    & 12.6\,M & \textbf{0.0129}\sd{0.0001} & 0.53          & 0.095                  & \textbf{1.02} \\
		\addlinespace
		2   & [4,8]       & 60    & 13.7\,M & 0.0138\sd{0.0001}          & 0.61          & 0.050                  & \textbf{1.02} \\
		    & [8,4]       & 60    & 13.7\,M & 0.0137\sd{0.0001}          & 0.63          & 0.042                  & 1.03          \\
		\addlinespace
		3   & [2,4,4]     & 48    & 12.0\,M & 0.0138\sd{0.0000}          & 0.62          & 0.037                  & 1.03          \\
		    & [4,2,4]     & 48    & 12.0\,M & 0.0144\sd{0.0001}          & 0.67          & 0.028                  & 1.03          \\
		    & [4,4,2]     & 48    & 12.0\,M & 0.0146\sd{0.0001}          & 0.67          & 0.029                  & 1.04          \\
		\addlinespace
		4   & [2,2,2,4]   & 41    & 11.4\,M & 0.0144\sd{0.0001}          & 0.68          & 0.021                  & 1.03          \\
		    & [2,2,4,2]   & 41    & 11.4\,M & 0.0144\sd{0.0002}          & 0.69          & 0.019                  & 1.04          \\
		    & [2,4,2,2]   & 41    & 11.4\,M & 0.0144\sd{0.0001}          & 0.70          & 0.018                  & 1.04          \\
		    & [4,2,2,2]   & 41    & 11.4\,M & 0.0154\sd{0.0002}          & \textbf{0.74} & \textbf{0.015}         & 1.04          \\
		\addlinespace
		5   & [2,2,2,2,2] & 37    & 12.7\,M & 0.0159\sd{0.0002}          & 0.70          & 0.020                  & 1.05          \\
		\bottomrule
	\end{tabular}
\end{table}

\subsection{Cross-window conditioning}
\label{app:nbr-ablation}

The conditioning $c$ in \eqref{eq:level} can include two kinds of information from adjacent windows. Cross-window context is especially useful for disaggregation, where consecutive reconstructed windows do not share an observed boundary state.

\paragraph{Observed context.}
The first is observed context: the aggregates of the preceding and following windows, each paired with a presence flag and randomly dropped during training, so that the same model can operate with or without them. Table~\ref{tab:nbr-ablation} evaluates the same trained models with cross-window context either provided or masked across all benchmark-regime window lengths. The benefit is largest at short windows: at $N=8$, context reduces CRPS by $25$--$87\%$ for \method{} across the four systems and brings cross-window step sizes closer to the truth, with both effects weakening as $N$ increases. Calibration changes comparatively little, except on Gray--Scott, where the added context noticeably shifts ensemble spread. Masking neighboring aggregates does not reverse the single-step comparison with flow matching at $N=32$ in the benchmark regime, even though flow matching keeps them: CRPS remains lower on all four systems ($0.0023$, $0.0036$, $0.0153$, $0.1371$, versus $0.0026$, $0.0049$, $0.0236$, $0.1578$, respectively, for advection-diffusion, Gray--Scott, Navier--Stokes, and Gross--Pitaevskii).

\begin{table}[t]
	\centering
	\caption{\textbf{Cross-window conditioning ablation.} Disaggregation on the AutoSim benchmark regimes, three seeds, evaluated with the same trained models and cross-window context either masked or fed. CRPS gain is $1-\mathrm{CRPS}_{\mathrm{fed}}/\mathrm{CRPS}_{\mathrm{masked}}$. $r_{\mathrm{seam}}$ is the cross-window step-size ratio defined in \eqref{eq:continuity}, with one indicating agreement with the truth.}
	\resizebox{\linewidth}{!}{%
		\begin{tabular}{llcccccccccc}
			\toprule
			                                     &     & \multicolumn{5}{c}{\method{}} & \multicolumn{5}{c}{\method{} ($R=1$)}                                                                                                                                                                            \\
			\cmidrule(lr){3-7}\cmidrule(lr){8-12}
			                                     &     & CRPS                          & \multicolumn{2}{c}{SSR $\to 1$}       & \multicolumn{2}{c}{$r_{\mathrm{seam}} \to 1$} & CRPS   & \multicolumn{2}{c}{SSR $\to 1$} & \multicolumn{2}{c}{$r_{\mathrm{seam}} \to 1$}                                 \\
			\cmidrule(lr){4-5}\cmidrule(lr){6-7}\cmidrule(lr){9-10}\cmidrule(lr){11-12}
			System                               & $N$ & gain                          & masked                                & fed                                           & masked & fed                             & gain                                          & masked & fed  & masked & fed  \\
			\midrule
			\multirow{4}{*}{Advection-diffusion} & 8   & 87\%                          & 0.97                                  & 0.92                                          & 2.56   & 1.03                            & 83\%                                          & 0.97   & 0.96 & 2.22   & 1.05 \\
			                                     & 16  & 49\%                          & 0.95                                  & 0.94                                          & 2.06   & 1.16                            & 48\%                                          & 0.94   & 0.92 & 1.74   & 1.12 \\
			                                     & 24  & 36\%                          & 0.91                                  & 0.93                                          & 1.88   & 1.24                            & 30\%                                          & 0.86   & 0.84 & 1.61   & 1.20 \\
			                                     & 32  & 23\%                          & 0.94                                  & 0.97                                          & 1.79   & 1.39                            & 14\%                                          & 0.85   & 0.83 & 1.49   & 1.28 \\
			\addlinespace
			\multirow{4}{*}{Gray--Scott}         & 8   & 31\%                          & 1.31                                  & 1.41                                          & 1.02   & 1.01                            & 9\%                                           & 1.20   & 1.22 & 1.01   & 1.01 \\
			                                     & 16  & 21\%                          & 1.10                                  & 1.19                                          & 1.05   & 1.01                            & 9\%                                           & 1.27   & 1.26 & 1.03   & 1.02 \\
			                                     & 24  & 19\%                          & 1.00                                  & 1.10                                          & 1.09   & 1.03                            & 9\%                                           & 1.28   & 1.35 & 1.06   & 1.03 \\
			                                     & 32  & 19\%                          & 0.86                                  & 1.01                                          & 1.17   & 1.07                            & 11\%                                          & 1.25   & 1.35 & 1.13   & 1.05 \\
			\addlinespace
			\multirow{4}{*}{Navier--Stokes}      & 8   & 59\%                          & 0.89                                  & 0.93                                          & 1.47   & 1.03                            & 48\%                                          & 0.82   & 0.84 & 1.33   & 1.03 \\
			                                     & 16  & 46\%                          & 0.85                                  & 0.85                                          & 2.20   & 1.26                            & 42\%                                          & 0.74   & 0.74 & 1.87   & 1.19 \\
			                                     & 24  & 29\%                          & 0.76                                  & 0.76                                          & 2.87   & 1.84                            & 27\%                                          & 0.63   & 0.62 & 2.48   & 1.63 \\
			                                     & 32  & 17\%                          & 0.70                                  & 0.70                                          & 4.03   & 2.97                            & 16\%                                          & 0.53   & 0.53 & 3.29   & 2.45 \\
			\addlinespace
			\multirow{4}{*}{Gross--Pitaevskii}   & 8   & 25\%                          & 0.79                                  & 0.74                                          & 1.04   & 1.01                            & 27\%                                          & 0.87   & 0.92 & 1.02   & 1.01 \\
			                                     & 16  & 15\%                          & 0.83                                  & 0.79                                          & 1.42   & 1.26                            & 19\%                                          & 0.89   & 0.91 & 1.28   & 1.18 \\
			                                     & 24  & 10\%                          & 0.88                                  & 0.85                                          & 1.86   & 1.65                            & 12\%                                          & 0.82   & 0.83 & 1.60   & 1.46 \\
			                                     & 32  & 8\%                           & 0.91                                  & 0.90                                          & 2.39   & 2.13                            & 9\%                                           & 0.84   & 0.85 & 1.93   & 1.75 \\
			\bottomrule
		\end{tabular}}%
	\label{tab:nbr-ablation}
\end{table}

\paragraph{Generated context.}
The second is generated context: recursive \method{} can additionally condition each level on the last state of the preceding window at that level's resolution. %
During training, this state comes from the true trajectory. At inference, it comes from the preceding window generated by the same ensemble member. This train--test mismatch is addressed during the end-to-end fine-tuning of Section~\ref{sec:method}.

Conditioned on the observations, the generated windows form a Markov chain. Unavailable neighboring observations are masked, with presence flags indicating their availability. The model distribution $q$ factorizes as
\begin{equation}
	q\left(x_{1:T} \mid y_{1:T}\right)
	= q\left(x_1\mid y_{1:2}\right)
	\prod_{t=2}^{T} q\left(x_t \mid x_{t-1}, y_{t-1:t+1}\right).
\end{equation}
Each window-level factor is defined by the multiresolution sampler. Writing $b_{t-1}^{(r)}$ for the last state of the preceding window at resolution $r$, for $t\geq2$ we have
\begin{equation}
	q\left(x_t \mid x_{t-1}, y_{t-1:t+1}\right)
	= \int \prod_{r=0}^{R-1}
	q_r\left(x_t^{(r)} \mid x_t^{(r+1)},\, b_{t-1}^{(r)},\, y_{t-1:t+1}\right)
	\, \mathrm{d}x_t^{(1)} \cdots \mathrm{d}x_t^{(R-1)},
	\label{eq:scale-chain-causal}
\end{equation}
where $x_t^{(0)}=x_t$ and $x_t^{(R)}=y_t$. The first window uses the same hierarchy with generated context masked. The restriction to $b_{t-1}^{(r)}$ at each level is a modeling assumption. We reserve generated context for recursive \method{}, where it is available at every intermediate resolution.

Table~\ref{tab:context-feedback} compares conditioning on true and generated preceding-window context. The two give CRPS within about $2\%$ of each other in most settings, except on advection-diffusion at $N=8$, where feeding back the generated state degrades CRPS and calibration.

\begin{table}[t]
	\centering
	\caption{\textbf{Generated cross-window context.} Disaggregation, three seeds. \emph{No input}: recursive \method{} trained without the preceding-window input. The other three columns report results for the same model trained with that input, evaluated with the input masked, filled with the true preceding state, or filled with the preceding window generated by the same ensemble member, as deployed. Lowest CRPS in each row in bold; SSR $\to 1$; $t_1$ the temporal variogram ratio at one step, $\to 1$.}%
	\setlength{\tabcolsep}{3pt}
	\resizebox{\linewidth}{!}{%
		\begin{tabular}{lllcccccccccccc}
			\toprule
			                    &           &     & \multicolumn{3}{c}{no input} & \multicolumn{3}{c}{masked} & \multicolumn{3}{c}{true} & \multicolumn{3}{c}{generated}                                                                           \\
			\cmidrule(lr){4-6} \cmidrule(lr){7-9} \cmidrule(lr){10-12} \cmidrule(lr){13-15}
			System              & Regime    & $N$ & CRPS                         & SSR                        & $t_1$                    & CRPS                          & SSR  & $t_1$ & CRPS             & SSR  & $t_1$ & CRPS    & SSR  & $t_1$ \\
			\midrule
			Advection-diffusion & benchmark & 8   & 0.00044                      & 0.89                       & 1.02                     & 0.00044                       & 0.94 & 1.02  & \textbf{0.00031} & 0.90 & 1.02  & 0.00047 & 2.26 & 1.02  \\
			Advection-diffusion & mild      & 8   & 0.00395                      & 1.23                       & 1.04                     & 0.00384                       & 1.20 & 1.04  & \textbf{0.00319} & 1.13 & 1.03  & 0.02587 & 1.93 & 1.53  \\
			Advection-diffusion & chaotic   & 8   & 0.01514                      & 1.18                       & 1.07                     & 0.01516                       & 1.20 & 1.07  & \textbf{0.01411} & 1.24 & 1.09  & 0.02774 & 1.76 & 1.41  \\
			\addlinespace
			Advection-diffusion & benchmark & 16  & 0.00173                      & 0.94                       & 1.02                     & 0.00161                       & 0.96 & 1.02  & \textbf{0.00145} & 0.94 & 1.02  & 0.00146 & 0.95 & 1.02  \\
			Advection-diffusion & mild      & 16  & 0.00657                      & 1.02                       & 1.02                     & 0.00651                       & 1.01 & 1.02  & \textbf{0.00610} & 1.02 & 1.02  & 0.00617 & 1.03 & 1.02  \\
			Advection-diffusion & chaotic   & 16  & 0.02107                      & 1.09                       & 1.03                     & 0.02131                       & 1.17 & 1.05  & \textbf{0.02094} & 1.17 & 1.05  & 0.02111 & 1.19 & 1.06  \\
			\addlinespace
			Gray--Scott         & benchmark & 8   & 0.00049                      & 0.89                       & 1.01                     & 0.00054                       & 0.91 & 1.01  & \textbf{0.00046} & 0.87 & 1.01  & 0.00047 & 0.86 & 1.01  \\
			Navier--Stokes      & benchmark & 8   & 0.00173                      & 0.94                       & 1.04                     & 0.00184                       & 0.94 & 1.04  & \textbf{0.00166} & 0.95 & 1.04  & 0.00167 & 0.95 & 1.04  \\
			\bottomrule
		\end{tabular}}%
	\label{tab:context-feedback}
\end{table}

\paragraph{Recursive variants.}
Therefore, recursive \method{} has four variants under disaggregation: with or without generated context, each with or without end-to-end fine-tuning. Under interpolation, enforcing the shared observed endpoint makes neighboring windows agree at their junction, so we do not use generated cross-window context and only vary fine-tuning. The tables report the plain or the fine-tuned variant, selected for each AutoSim system, predictability regime, and window length on validation diagnostics only, and scored on the same test windows as every other row. The causal chain conditions on its own output for the preceding window, so it is scored on chains, every window but the first of each test trajectory; Table~\ref{tab:ablation-autosim} compares the four variants and single-step \method{} on those windows for $N\in\{24,32\}$ in the benchmark and chaotic regimes. On those windows the fine-tuned chain is the best of the four variants in every setting, lowering the CRPS of plain recursive \method{} by $13$--$35\%$ on Navier--Stokes and Gross--Pitaevskii; against single-step \method{} it is ahead on Gross--Pitaevskii and chaotic advection-diffusion, level on Navier--Stokes, and behind on Gray--Scott and benchmark advection-diffusion. On Lorenz-96, we instead fix the variant in advance: the causal chain for disaggregation and the plain sampler for interpolation, both without fine-tuning.
\begin{table}[t]
	\centering
	\caption{\textbf{Recursive \method{} variants on AutoSim.} CRPS and coverage MAE on the test trajectories, three seeds. Disaggregation: the four variants and single-step \method{} on the same chain windows (every window but the first of each test trajectory) and members; plain and fine-tuned, without and with the generated context of the causal chain. Interpolation: plain against fine-tuned and single-step \method{}, on the standard evaluation windows. $^{\star}$: the variant reported in the main tables, plain or fine-tuned, selected on the validation trajectories. Bold: lowest per row and metric, including ties at the printed precision.}%
	\footnotesize
	\setlength{\tabcolsep}{3pt}
	\resizebox{\linewidth}{!}{%
		\begin{tabular}{lllcccccccccc}
			\toprule
			                    &           &     & \multicolumn{2}{c}{plain} & \multicolumn{2}{c}{fine-tuned} & \multicolumn{2}{c}{causal chain} & \multicolumn{2}{c}{causal chain, fine-tuned} & \multicolumn{2}{c}{single-step}                                                                                                                    \\
			\cmidrule(lr){4-5} \cmidrule(lr){6-7} \cmidrule(lr){8-9} \cmidrule(lr){10-11} \cmidrule(lr){12-13}
			System              & Regime    & $N$ & CRPS $\downarrow$         & cov.\ MAE $\downarrow$         & CRPS $\downarrow$                & cov.\ MAE $\downarrow$                       & CRPS $\downarrow$               & cov.\ MAE $\downarrow$ & CRPS $\downarrow$ & cov.\ MAE $\downarrow$ & CRPS $\downarrow$ & cov.\ MAE $\downarrow$ \\
			\midrule
			\multicolumn{13}{l}{\emph{Disaggregation}}                                                                                                                                                                                                                                                                                                \\
			Advection-diffusion & benchmark & 24  & 0.0022$^{\star}$          & \textbf{0.114}                          & 0.0020                           & 0.118                                        & 0.0020                          & 0.117                  & 0.0018            & 0.125                  & \textbf{0.0017}            & 0.158                  \\
			Advection-diffusion & benchmark & 32  & 0.0027$^{\star}$          & \textbf{0.093}                          & 0.0026                           & 0.100                                        & 0.0026                          & 0.100                  & 0.0025            & 0.108                  & \textbf{0.0021}            & 0.142                  \\
			Advection-diffusion & chaotic   & 24  & 0.0247                    & 0.014                          & 0.0230$^{\star}$                 & \textbf{0.010}                                        & 0.0242                          & 0.016                  & \textbf{0.0225}            & 0.014                  & 0.0344            & 0.043                  \\
			Advection-diffusion & chaotic   & 32  & 0.0281                    & 0.026                          & 0.0265$^{\star}$                 & \textbf{0.010}                                        & 0.0283                          & 0.032                  & \textbf{0.0256}            & 0.013                  & 0.0412            & 0.042                  \\
			\addlinespace
			Gray--Scott         & benchmark & 24  & 0.0014$^{\star}$          & \textbf{0.048}                          & 0.0014                           & 0.051                                        & 0.0014                          & \textbf{0.048}                  & \textbf{0.0013}            & 0.051                  & \textbf{0.0013}            & 0.052                  \\
			Gray--Scott         & benchmark & 32  & 0.0019$^{\star}$          & \textbf{0.044}                          & 0.0019                           & 0.049                                        & 0.0018                          & \textbf{0.044}                  & 0.0018            & 0.049                  & \textbf{0.0016}            & 0.053                  \\
			Gray--Scott         & chaotic   & 24  & 0.0100                    & 0.061                          & 0.0091$^{\star}$                 & 0.068                                        & 0.0088                          & 0.059                  & 0.0082            & 0.069                  & \textbf{0.0070}            & \textbf{0.043}                  \\
			Gray--Scott         & chaotic   & 32  & 0.0183                    & 0.054                          & 0.0163$^{\star}$                 & 0.058                                        & 0.0160                          & 0.055                  & 0.0146            & 0.061                  & \textbf{0.0115}            & \textbf{0.037}                  \\
			\addlinespace
			Navier--Stokes      & benchmark & 24  & 0.0084                    & \textbf{0.023}                          & 0.0081$^{\star}$                 & 0.024                                        & 0.0071                          & 0.038                  & \textbf{0.0069}            & 0.038                  & 0.0071            & 0.048                  \\
			Navier--Stokes      & benchmark & 32  & 0.0153$^{\star}$          & \textbf{0.019}                          & 0.0147                           & \textbf{0.019}                                        & 0.0127                          & \textbf{0.019}                  & \textbf{0.0123}            & 0.021                  & 0.0125            & 0.095                  \\
			Navier--Stokes      & chaotic   & 24  & 0.0574                    & 0.028                          & 0.0541$^{\star}$                 & 0.037                                        & 0.0503                          & 0.027                  & \textbf{0.0484}            & 0.035                  & 0.0519            & \textbf{0.017}                  \\
			Navier--Stokes      & chaotic   & 32  & 0.0869                    & 0.027                          & 0.0823$^{\star}$                 & 0.034                                        & 0.0777                          & \textbf{0.020}                  & 0.0754            & 0.024                  & \textbf{0.0746}            & 0.021                  \\
			\addlinespace
			Gross--Pitaevskii   & benchmark & 24  & 0.1146                    & 0.068                          & 0.0899$^{\star}$                 & 0.055                                        & 0.1002                          & 0.074                  & \textbf{0.0748}            & 0.078                  & 0.0938            & \textbf{0.048}                  \\
			Gross--Pitaevskii   & benchmark & 32  & 0.1523                    & 0.072                          & 0.1291$^{\star}$                 & \textbf{0.042}                                        & 0.1384                          & 0.077                  & \textbf{0.1130}            & 0.062                  & 0.1345            & 0.044                  \\
			\midrule
			\multicolumn{13}{l}{\emph{Interpolation}}                                                                                                                                                                                                                                                                                                 \\
			Advection-diffusion & benchmark & 24  & 0.0015$^{\star}$          & \textbf{0.099}                          & 0.0014                           & 0.118                                        & --                              & --                     & --                & --                     & \textbf{0.0010}            & 0.172                  \\
			Advection-diffusion & benchmark & 32  & 0.0022$^{\star}$          & \textbf{0.090}                          & 0.0020                           & 0.106                                        & --                              & --                     & --                & --                     & \textbf{0.0013}            & 0.160                  \\
			Advection-diffusion & chaotic   & 24  & 0.0222                    & \textbf{0.036}                          & 0.0207$^{\star}$                 & \textbf{0.036}                                        & --                              & --                     & --                & --                     & \textbf{0.0185}            & 0.048                  \\
			Advection-diffusion & chaotic   & 32  & 0.0285                    & 0.041                          & 0.0266$^{\star}$                 & \textbf{0.032}                                        & --                              & --                     & --                & --                     & \textbf{0.0213}            & 0.047                  \\
			\addlinespace
			Gray--Scott         & benchmark & 24  & 0.0030$^{\star}$          & 0.044                          & 0.0029                           & \textbf{0.039}                                        & --                              & --                     & --                & --                     & \textbf{0.0022}            & 0.059                  \\
			Gray--Scott         & benchmark & 32  & 0.0035$^{\star}$          & \textbf{0.037}                          & 0.0034                           & 0.039                                        & --                              & --                     & --                & --                     & \textbf{0.0024}            & 0.060                  \\
			Gray--Scott         & chaotic   & 24  & 0.0115                    & \textbf{0.037}                          & 0.0100$^{\star}$                 & 0.041                                        & --                              & --                     & --                & --                     & \textbf{0.0074}            & 0.041                  \\
			Gray--Scott         & chaotic   & 32  & 0.0203                    & \textbf{0.031}                          & 0.0169$^{\star}$                 & 0.040                                        & --                              & --                     & --                & --                     & \textbf{0.0112}            & 0.032                  \\
			\addlinespace
			Navier--Stokes      & benchmark & 24  & 0.0074$^{\star}$          & \textbf{0.017}                          & 0.0071                           & 0.020                                        & --                              & --                     & --                & --                     & \textbf{0.0063}            & 0.048                  \\
			Navier--Stokes      & benchmark & 32  & 0.0117$^{\star}$          & \textbf{0.015}                          & 0.0112                           & 0.022                                        & --                              & --                     & --                & --                     & \textbf{0.0091}            & 0.072                  \\
			Navier--Stokes      & chaotic   & 24  & 0.0506                    & 0.028                          & 0.0427$^{\star}$                 & 0.033                                        & --                              & --                     & --                & --                     & \textbf{0.0370}            & \textbf{0.022}                  \\
			Navier--Stokes      & chaotic   & 32  & 0.0842                    & \textbf{0.024}                          & 0.0663$^{\star}$                 & 0.036                                        & --                              & --                     & --                & --                     & \textbf{0.0650}            & 0.028                  \\
			\addlinespace
			Gross--Pitaevskii   & benchmark & 24  & 0.0320                    & 0.072                          & 0.0253$^{\star}$                 & \textbf{0.055}                                        & --                              & --                     & --                & --                     & \textbf{0.0226}            & 0.082                  \\
			Gross--Pitaevskii   & benchmark & 32  & 0.0660                    & 0.056                          & 0.0450$^{\star}$                 & \textbf{0.034}                                        & --                              & --                     & --                & --                     & \textbf{0.0406}            & 0.087                  \\
			\bottomrule
		\end{tabular}}%
	\label{tab:ablation-autosim}
\end{table}

\FloatBarrier
\section{Experimental details}
\label{app:experiments}

\subsection{Lorenz-96}
\label{app:l96}

The two-scale Lorenz-96 model~\citep{lorenzPredictabilityProblemPartly2006} couples $K$ slow variables $X_k$ to $J$ fast variables $Y_{j,k}$ per slow variable,
\begin{align}
	\frac{d}{dt}X_k     & = -X_{k-1}(X_{k-2} - X_{k+1}) - X_k + F - \frac{hc}{b} \sum_{j=1}^J Y_{j,k}, \\
	\frac{d}{dt}Y_{j,k} & = -cb Y_{j+1,k}(Y_{j+2,k} - Y_{j-1,k}) - c Y_{j,k} + \frac{hc}{b} X_k,
\end{align}
with $h=1$, $b=10$, and $c=10$. We use $K=36$ and $J=32$. We vary the forcing over $F \in \{10,24,32,64\}$ to span regimes of increasing chaoticity, with $F=10$ non-chaotic and $F=64$ strongly chaotic.
The system is integrated with fourth-order Runge--Kutta at a step of $10^{-3}$ time units for 1000 time units after a 100-unit spin-up. Only the slow variables are observed and reconstructed. The fast variables remain latent and act as unresolved forcing on the slow dynamics. %

The target resolution is $\delta=0.01$ time units, with windows of $N\in\{64,128\}$ steps. %
The trajectory is split sequentially into 70\% training (time units 100--800), 15\% validation (800--950), and 15\% test (950--1100), with no window crossing a split boundary. Training samples 50\,000 windows from the training interval. Evaluation uses 25 chains of contiguous test windows, at least 0.6 time units apart, with 256 ensemble members per window. Each chain contains 8 windows at $N=64$ or 4 at $N=128$. We exclude the first window, which has no predecessor, leaving 175 and 75 scored windows, respectively.

\subsection{AutoSim}
\label{app:autosim}

We generate four $64\times64$ physical systems with AutoSim \citep{greenburyReliabilityProbabilisticEmulation2026}, spanning transport, reaction-diffusion, turbulent flow, and nonlinear dispersive dynamics. Their governing equations are
\newcommand{\lhs}[1]{\hphantom{\nabla\cdot\mathbf{u}}\llap{$#1$}}
\begin{alignat}{2}
	 & \textbf{Advection-diffusion}        &  & \qquad
	\begin{aligned}
		\lhs{\partial_t \omega} & = \nu \nabla^2 \omega - \mu\, \mathbf{u}\cdot\nabla\omega,
	\end{aligned}
	\label{eq:advdiff}                                 \\[6pt]
	 & \textbf{Gray--Scott}                &  & \qquad
	\begin{aligned}
		\lhs{\partial_t u} & = \delta_u \nabla^2 u - uv^2 + F(1-u),   \\
		\lhs{\partial_t v} & = \delta_v \nabla^2 v + uv^2 - (F+k)\,v,
	\end{aligned}
	\label{eq:grayscott}                               \\[6pt]
	 & \textbf{Conditioned Navier--Stokes} &  & \qquad
	\begin{aligned}
		\lhs{\partial_t s}          & = -\mathbf{u}\cdot\nabla s,                                                                      \\
		\lhs{\partial_t \mathbf{u}} & = -(\mathbf{u}\cdot\nabla)\mathbf{u} - \nabla p + \nu\nabla^2\mathbf{u} + \beta s\,\mathbf{e}_y, \\
		\lhs{\nabla\cdot\mathbf{u}} & = 0,
	\end{aligned}
	\label{eq:cns}                                     \\[6pt]
	 & \textbf{Gross--Pitaevskii}          &  & \qquad
	\begin{aligned}
		\lhs{i\,\partial_t\psi} & = \bigl[-\tfrac12\nabla^2 + V(\mathbf{x},t) + g|\psi|^2\bigr]\,\psi.
	\end{aligned}
	\label{eq:gpe}
\end{alignat}

The reconstructed fields are vorticity for advection-diffusion, the two reactants for Gray--Scott, passive scalar and velocity components for conditioned Navier--Stokes, and density together with the real and imaginary parts of $\psi$ for Gross--Pitaevskii. In advection-diffusion, $\mathbf{u}$ is the velocity induced by $\omega$ through its streamfunction. In Gross--Pitaevskii, $V$ is a box trap with an oscillating obstacle.

We generate 200 training trajectories and 20 validation and test trajectories per system, with splits made by trajectory.
We evaluate windows of $N\in\{8,16,24,32\}$ steps, with $M=64$ ensemble members, on 200 test windows per system (180 for interpolation) spread over the test trajectories.
In addition to the benchmark configuration, we use mild and chaotic regimes with progressively shorter decorrelation times. For Gray--Scott, these regimes are obtained by increasing the snapshot stride rather than changing the underlying dynamics. Table~\ref{tab:autosim-gen} gives the simulation parameters.

\begin{table}[ht]
	\centering
	\caption{\textbf{Generation parameters.} Systems in the order of \eqref{eq:advdiff}--\eqref{eq:gpe}. Span denotes trajectory length and $\delta$ the saved snapshot interval, i.e., the target resolution, both in simulation units. Parameters not listed take the values of AutoSim's shipped configurations.}%
	\footnotesize
	\setlength{\tabcolsep}{3pt}
	\begin{tabular}{ll*{3}{>{\centering\arraybackslash}p{2.2cm}}}
		\toprule
		System                               & Parameter                   & Benchmark                   & Mild                     & Chaotic                    \\
		\midrule
		\multirow{4}{*}{Advection-diffusion} & advection $\mu$             & $[0.5, 2]$                  & $[1, 4]$                 & $[2, 8]$                   \\
		                                     & diffusivity $\nu$           & $[10^{-4}, 10^{-2}]$        & $[5, 500] \cdot 10^{-5}$ & $[2.5, 250] \cdot 10^{-5}$ \\
		                                     & snapshot step $\delta$      & \multicolumn{3}{c}{$0.25$}                                                          \\
		                                     & span                        & \multicolumn{3}{c}{$80$}                                                            \\
		\midrule
		\multirow{2}{*}{Gray--Scott}         & snapshot step $\delta$      & $4$                         & $8$                      & $16$                       \\
		                                     & span                        & $1280$                      & $2560$                   & $5120$                     \\
		\midrule
		\multirow{3}{*}{Navier--Stokes}      & buoyancy $\beta$            & $[0.2, 0.8]$                & $[1.6, 6.4]$             & $[3.2, 12.8]$              \\
		                                     & snapshot step $\delta$      & \multicolumn{3}{c}{$0.261$}                                                         \\
		                                     & span                        & \multicolumn{3}{c}{$85.9$}                                                          \\
		\midrule
		\multirow{4}{*}{Gross--Pitaevskii}   & obstacle frequency $\omega$ & $[0.5, 2]$                  & $[1.5, 3]$               & --                         \\
		                                     & interaction $g$             & $[80, 200]$                 & $[160, 400]$             & --                         \\
		                                     & snapshot step $\delta$      & \multicolumn{3}{c}{$0.05$}                                                          \\
		                                     & span                        & \multicolumn{3}{c}{$17.55$}                                                         \\
		\bottomrule
	\end{tabular}%
	\label{tab:autosim-gen}
\end{table}

\subsection{ERA5}
\label{app:era5}

We use hourly ERA5 reanalysis~\citep{hersbachERA5GlobalReanalysis2020} at $0.25^\circ$ on a grid covering Europe ($25^\circ$W--$45^\circ$E, $35^\circ$N--$72^\circ$N, $149 \times 281$ points), on seven surface fields: 10m u and v wind, 2m temperature and dewpoint, mean sea level pressure, total column water and total precipitation. Splits are by year: 1979--2015 for training, 2016--2018 for validation, 2019--2022 for test. \method{} and the flow-based baselines are conditioned on adjacent-window observations and calendar information (hour of day and day of year), together with static orography, land-sea mask, latitude, and longitude fields as in ArchesWeather \citep{couaironArchesWeatherGenSkillfulComputeefficient2026}. Evaluation uses 96 non-overlapping test windows evenly spaced over 2019--2022, with $M=24$ ensemble members.

\subsection{Training configuration and baselines}
\label{app:baselines}

For each tier, single-step and recursive \method{}, conditional flow matching, and FlowDPS use the same backbone and are matched in parameter count where feasible; SimVP retains its original architecture. Separate models are trained for interpolation and disaggregation. All learned arms use AdamW \citep{loshchilovDecoupledWeightDecay2019}. AutoSim and ERA5 use cosine decay with 500 warm-up steps. Lorenz-96 and AutoSim models are trained in FP32, and ERA5 models in BF16 mixed precision. The \method{} training loss is estimated from $M=2$ draws. Table~\ref{tab:hyperparams} summarizes the training configuration and sampling cost. For AutoSim, it uses Gross--Pitaevskii at $N=32$ as a representative setting.

\paragraph{Capacity on ERA5.} On ERA5, recursive \method{} uses four $57.4$\,M-parameter ArchesWeather networks, whereas single-step \method{} and the flow-based baselines each use one. The hierarchy is what makes this capacity affordable: each level is trained independently, so the four networks can be trained in parallel, each at the cost of one single-step network ($90.1$ GPU-hours in total against $22.8$, Table~\ref{tab:hyperparams}). Scaling the shared backbone to the total capacity of the four levels, whether for single-step \method{} or a flow-based baseline, was beyond our compute budget. The recursive comparison therefore does not separate the effect of the hierarchy from its added capacity, but single-step \method{} provides the comparison at matched parameter count.

\begin{table}[t]
	\centering
	\caption{\textbf{Training configuration and computational cost.} ES denotes the energy score, VS the variogram penalty, SC the spectral CRPS, and NFE the number of network evaluations per generated member. Training cost is reported in H100 GPU-hours. Budget is in epochs (ep.) on Lorenz-96 and in optimizer steps elsewhere. The \method{} training budget is reported per sampler level; end-to-end fine-tuning updates the complete hierarchy jointly. $^{a}$Sampling measured on eight CPU cores.} %
	\setlength{\tabcolsep}{2.8pt}
	\resizebox{\linewidth}{!}{%
		\begin{tabular}{llcccccccccc}
			\toprule
			     &                   & \multicolumn{3}{c}{Model} & \multicolumn{4}{c}{Training} & \multicolumn{3}{c}{Cost}                                                                                                                                \\
			\cmidrule(lr){3-5}\cmidrule(lr){6-9}\cmidrule(lr){10-12}
			Tier & Arm               & Backbone                  & Width                        & Params                   & Objective                 & Budget               & Batch & lr                                  & GPU-h     & NFE & ms/member \\
			\midrule
			\multirow{4}{*}{\shortstack[l]{Lorenz-96                                                                                                                                                                                                      \\$F=32$, $N=64$}}
			     & \method{} ($R=6$) & circ.\ CNN                & 256                          & 6.18\,M                  & ES                        & 60 ep.               & 256   & $10^{-3}$                           & 0.10      & 6   & 1.6$^{a}$ \\
			     & \method{} ($R=1$) & circ.\ CNN                & 620                          & 6.18\,M                  & ES                        & 60 ep.               & 256   & $10^{-3}$                           & 0.06      & 1   & 1.4$^{a}$ \\
			     & Flow matching     & circ.\ CNN                & 618                          & 6.18\,M                  & flow matching             & 60 ep.               & 256   & $10^{-3}$                           & 0.06      & 56  & 2.6       \\
			     & FlowDPS           & circ.\ CNN                & 618                          & 6.18\,M                  & flow matching             & 60 ep.               & 256   & $10^{-3}$                           & 0.05      & 56  & 2.7       \\
			\midrule
			\multirow{5}{*}{\shortstack[l]{Gross--Pitaevskii                                                                                                                                                                                              \\$N=32$}}
			     & \method{} ($R=5$) & NAT U-Net                 & 37                           & 12.7\,M                  & ES+VS $\rightarrow$ ES+SC & 40k $\rightarrow$ 8k & 16    & $3{\cdot}10^{-4}\rightarrow10^{-4}$ & 3.1 + 0.6 & 5   & 1.3       \\
			     & \method{} ($R=1$) & NAT U-Net                 & 85                           & 12.6\,M                  & ES+VS                     & 40k                  & 16    & $3{\cdot}10^{-4}$                   & 1.4       & 1   & 0.7       \\
			     & Flow matching     & NAT U-Net                 & 85                           & 12.4\,M                  & flow matching             & 40k                  & 16    & $3{\cdot}10^{-4}$                   & 0.72      & 56  & 39        \\
			     & FlowDPS           & NAT U-Net                 & 85                           & 12.4\,M                  & flow matching             & 40k                  & 16    & $3{\cdot}10^{-4}$                   & 0.70      & 56  & 40        \\
			     & SimVP             & SimVP-gSTA                & --                           & 9.7\,M                   & MSE                       & 40k                  & 16    & $3{\cdot}10^{-4}$                   & 0.36      & 1   & 2.4       \\
			\midrule
			\multirow{5}{*}{\shortstack[l]{ERA5                                                                                                                                                                                                           \\$N=24$}}
			     & \method{} ($R=4$) & ArchesWeather             & 96                           & $4\times57.4$\,M         & ES+VS                     & 40k                  & 4     & $3{\cdot}10^{-4}$                   & 90.1      & 4   & 507       \\
			     & \method{} ($R=1$) & ArchesWeather             & 96                           & 57.4\,M                  & ES+VS                     & 40k                  & 4     & $3{\cdot}10^{-4}$                   & 22.8      & 1   & 127       \\
			     & Flow matching     & ArchesWeather             & 96                           & 57.4\,M                  & flow matching             & 40k                  & 4     & $3{\cdot}10^{-4}$                   & 11.4      & 25  & 3150      \\
			     & FlowDPS           & ArchesWeather             & 96                           & 57.4\,M                  & flow matching             & 40k                  & 4     & $3{\cdot}10^{-4}$                   & 11.3      & 25  & 3160      \\
			     & SimVP             & SimVP-gSTA                & --                           & 7.3\,M                   & MSE                       & 40k                  & 4     & $3{\cdot}10^{-4}$                   & 2.5       & 1   & 6         \\
			\bottomrule
		\end{tabular}%
	}
	\label{tab:hyperparams}
\end{table}

On all three benchmarks, every learned baseline is projected onto the constraint set after sampling using the map in Equation~\eqref{eq:level}.
On AutoSim and ERA5, the same projection is used by FlowDPS as its consistency update at each sampling step. Its prior is trained with the same flow-matching loss and does not see the window observation, which enters only through this projection at each step. On Lorenz-96, FlowDPS follows the gradient update of \citet{kimFlowdpsFlowdrivenPosterior2025}. Conditional flow matching is the conditional counterpart and has no consistency step. Network evaluations per member are reported in Table~\ref{tab:hyperparams}. %

\subsection{Metrics}
\label{app:metrics}

Let $x = (x_1, \dots, x_N)$ be the withheld fine record of a test window, $y = Ax$ its observation, and $u^{(1)}, \dots, u^{(M)}$ the ensemble members drawn by a method from that observation. Each quantity below is computed per window and averaged over windows, variables and pixels, unless it is defined by pooling windows.

\paragraph{Accuracy.}
The CRPS is estimated in its fair form \citep{ferroFairScoresEnsemble2014},
\begin{equation}
	\mathrm{CRPS}\left(u^{1:M}, v\right) = \frac{1}{M}\sum_{m=1}^{M} \lvert u^{(m)} - v \rvert - \frac{1}{2M\left(M-1\right)} \sum_{m \neq m'} \lvert u^{(m)} - u^{(m')} \rvert ,
	\label{eq:crps}
\end{equation}
applied at each position $i$ to each variable of $x_i$. %

\paragraph{Calibration.}
For a calibrated ensemble, the rank of $x_i$ among $u^{(1)}_i, \dots, u^{(M)}_i$ is uniform on $\{1, \dots, M+1\}$, per position and per variable. We summarize its histogram by the realized coverage of central member intervals and its mean absolute error over a grid of nominal levels \citep{greenburyReliabilityProbabilisticEmulation2026},
\begin{equation}
	\widehat{c}(\alpha) = \Bigl\langle \mathbf{1}\bigl\{
	\widehat q_{(1-\alpha)/2}\bigl(u^{1:M}_i\bigr) \le x_i \le
	\widehat q_{(1+\alpha)/2}\bigl(u^{1:M}_i\bigr) \bigr\} \Bigr\rangle ,
	\qquad
	\text{cov.\ MAE} = \frac{1}{\lvert \mathcal{A} \rvert} \sum_{\alpha \in \mathcal{A}}
	\bigl\lvert \widehat{c}(\alpha) - \alpha \bigr\rvert ,
	\label{eq:covmae}
\end{equation}
with $\mathcal{A} = \{0.1, 0.2, \dots, 0.9, 0.95, 0.99\}$ and $\widehat q$ the member quantile at Weibull plotting positions. %

We also report the spread-skill ratio with a finite-ensemble correction \citep{raspWeatherBench2Benchmark2024},
\begin{equation}
	\mathrm{SSR} = \sqrt{\frac{M+1}{M}}\
	\frac{\sqrt{\bigl\langle \operatorname{var}\bigl(u^{1:M}_i\bigr) \bigr\rangle}}
	{\sqrt{\bigl\langle \bigl(\overline{u}_i - x_i\bigr)^2 \bigr\rangle}} ,
\end{equation}
where $\operatorname{var}$ is the unbiased member variance and $\langle \cdot \rangle$ averages over positions and windows. A value close to 1 indicates agreement between ensemble spread and ensemble-mean error. Values below 1 indicate insufficient spread relative to that error, and values above 1 indicate excess spread.

\paragraph{Consistency.}
A reconstruction is consistent when it reproduces the coarse observation exactly: retained states under interpolation and imposed aggregates under disaggregation. The consistency error is
\begin{equation}
	\epsilon_{\mathrm{cons}}
	=
	\begin{cases}
		\displaystyle
		\max_{j,m}
		\left\lVert
		(A_{\mathrm{avg}}u^{(m)})_j-y_j
		\right\rVert_\infty,
		 & \text{disaggregation}, \\[6pt]
		\displaystyle
		\max_{j,m}
		\left\lVert
		u^{(m)}_{i_j}-y_j
		\right\rVert_\infty,
		 & \text{interpolation},
	\end{cases}
\end{equation}
where $i_j$ is the position of the $j$-th stored state. A method passes if the corresponding error is below $10^{-4}$ in standardized units. For the methods marked \yesp{}, $\epsilon_{\mathrm{cons}}$ is measured after the projection of Appendix~\ref{app:baselines}.

\paragraph{Temporal coherence.}
We measure temporal coherence with the ratio of order-$p$ temporal differences
\begin{equation}
	t_h =
	\frac{\left\langle \left|u_{t+h}-u_t\right|^p \right\rangle}
	{\left\langle \left|x_{t+h}-x_t\right|^p \right\rangle},
	\qquad p=\tfrac12,
	\label{eq:temporal-variogram-ratio}
\end{equation}
where the averages are taken over ensemble members, valid time pairs, windows, variables, and spatial positions. We report $t_1$, with one indicating agreement with the truth. Values above one indicate larger short-lag increments than in the reference, and values below one indicate smaller increments. %

\paragraph{Spatial coherence.}
Using the order-$p$ pairwise differences $\gamma_h^a$ of \eqref{eq:variogram-score}, we measure spatial coherence with the ratio
\begin{equation}
	s_h =
	\frac{\left\langle \gamma_h^a(u) \right\rangle}
	{\left\langle \gamma_h^a(x) \right\rangle},
	\qquad p=\tfrac12,
	\label{eq:spatial-variogram-ratio}
\end{equation}
where the averages are taken over ensemble members, frames, windows, variables, and spatial positions. We report $s_1$, with one indicating agreement with the truth. Values above one indicate fields that are too rough at the grid scale, and values below one fields that are too smooth.

\paragraph{Fine-scale spectral power.}
For the two-dimensional systems, we additionally report the ratio of generated to true power in the finest resolved spatial band. On the periodic AutoSim domains, spectra are computed along one periodic axis without windowing, and the finest band is the upper half of the resolved wavenumber range, $[K/2,K]$, with $K$ the Nyquist wavenumber. The reported ratio sums power over this band and averages over ensemble members, frames, windows, and the orthogonal spatial axis. A value of one indicates that the reconstruction carries the correct amount of power at the finest resolved scales. This metric is insensitive to where that power is located and is therefore used only as a complementary diagnostic in the spatial-coherence ablation of Appendix~\ref{app:vgram-ablation}.

\paragraph{Cross-window continuity.}
Within-window metrics do not detect artifacts at the junction between consecutive reconstructions, so we evaluate stitched chains of windows. Let $\mathcal{B}$ denote the time steps crossing a window boundary, and let $\tilde u$ and $\tilde x$ be the stitched reconstruction and truth. We define the seam ratio
\begin{equation}
	r_{\mathrm{seam}}
	=
	\frac{
		\bigl\langle \lvert \tilde u_{t+1}-\tilde u_t \rvert \bigr\rangle_{t\in\mathcal{B}}
	}{
		\bigl\langle \lvert \tilde x_{t+1}-\tilde x_t \rvert \bigr\rangle_{t\in\mathcal{B}}
	},
	\label{eq:continuity}
\end{equation}
where the average is taken over boundary steps, chains, variables, and spatial positions. A value of one indicates that cross-window steps have the correct magnitude, but not necessarily the correct direction or value, so the metric can be improved by compensating errors.

\FloatBarrier
\section{Additional results}
\label{app:results}

\subsection{Lorenz-96}
\label{app:l96-results}

Tables~\ref{tab:l96-N64-sweep} and~\ref{tab:l96-N128-sweep} extend the main Lorenz-96 results to the remaining forcings and to windows of $N=128$ steps. Each panel reports two measures of how much the dynamics decorrelate within a window of duration $T=N\delta$: the ratio $N\delta/\tau$, where $\tau$ is the $e$-folding decorrelation time of the slow variables, and $\lambda T$, where $\lambda$ is the leading Lyapunov exponent, so that $\lambda T$ counts the $e$-foldings of a small perturbation over the window. Both grow with $F$ and $N$, from the non-chaotic $F=10$ ($\lambda=0$) to about thirteen $e$-foldings at $F=64$, $N=128$.

Under disaggregation, flow matching attains the lowest CRPS in the six chaotic settings, and FlowDPS is also below recursive \method{} at $N=128$ for $F\in\{32,64\}$. At $F=10$, recursive \method{} has the lowest CRPS of all methods. At $N=64$, flow matching also has the lower coverage error at every forcing. At $N=128$, recursive \method{} has the spread-skill ratio closer to one and the lower coverage error in three of the four forcings, while flow matching overestimates short-lag variability for $F\leq32$. Recursion lowers CRPS relative to single-step \method{} in seven of the eight settings and keeps the temporal variogram ratio close to one where single-step \method{} falls below it, although single-step \method{} has the lower coverage error in six of them.
Under interpolation, recursion improves CRPS only in the non-chaotic regime. For $F\geq24$, single-step \method{} has the lower CRPS at both window lengths, while recursive \method{} retains the lower coverage error in every setting. Flow matching overestimates short-lag variability at every forcing and is strongly overdispersed at low forcing, whereas recursive \method{} keeps $t_1$ close to one in every setting. The classical baselines satisfy the constraint but remain far behind in CRPS, and their $t_1$ below one reflects the missing short-lag variability of smooth reconstructions.

\begin{figure}[p]
	\centering
	\includegraphics[width=\linewidth]{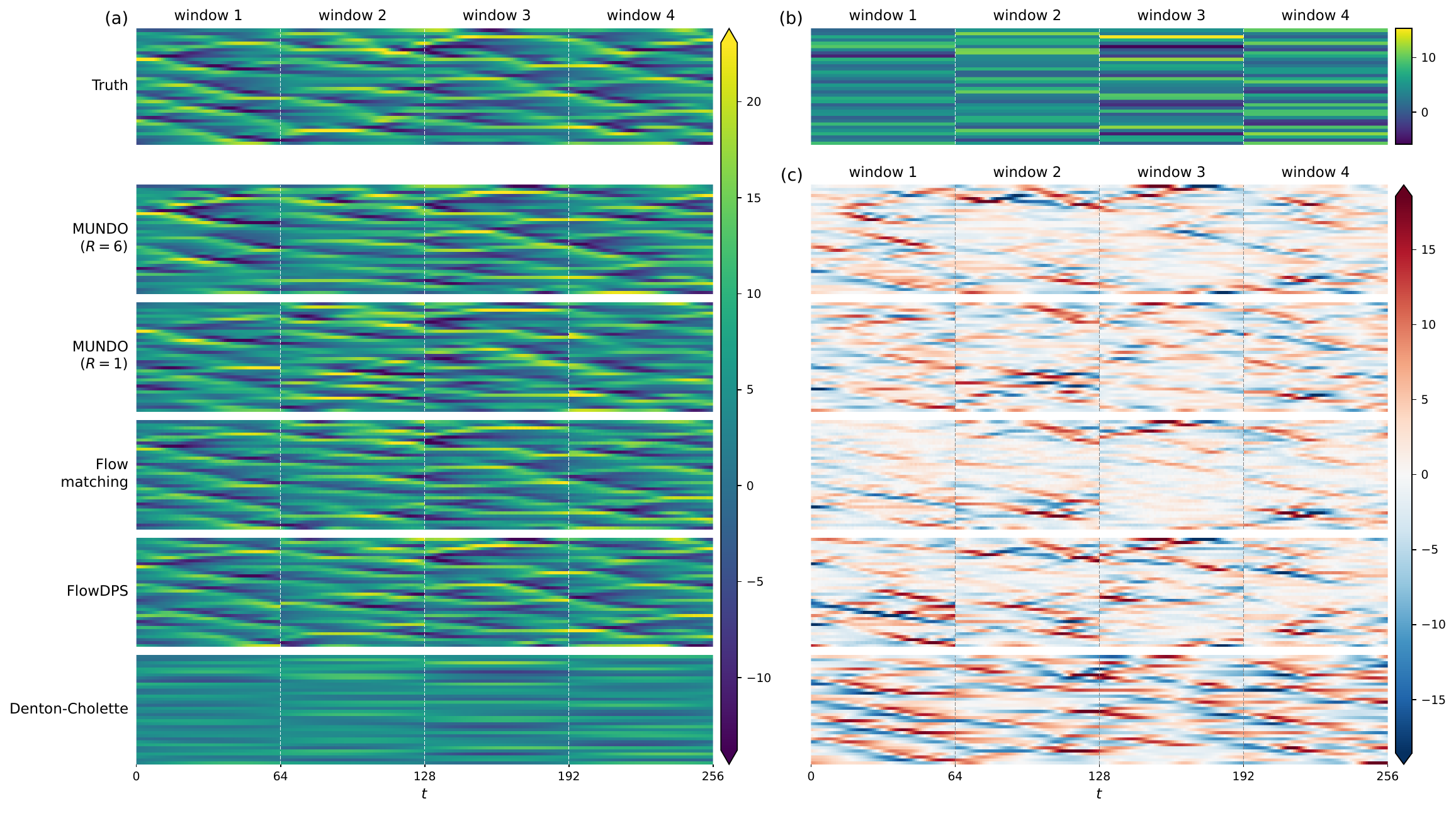}
	\caption{\textbf{Lorenz-96 disaggregation at $F=32$, $N=64$}, four consecutive windows, one member per method.
		(a) Truth and reconstructions, slow variables against time.
		(b) Observed window means.
		(c) Reconstruction errors $\hat{x}-x$; rows follow the method order in (a). Dashed lines mark the window boundaries.}
	\label{fig:l96-hov-disagg}
\end{figure}

\begin{figure}[p]
	\centering
	\includegraphics[width=\linewidth]{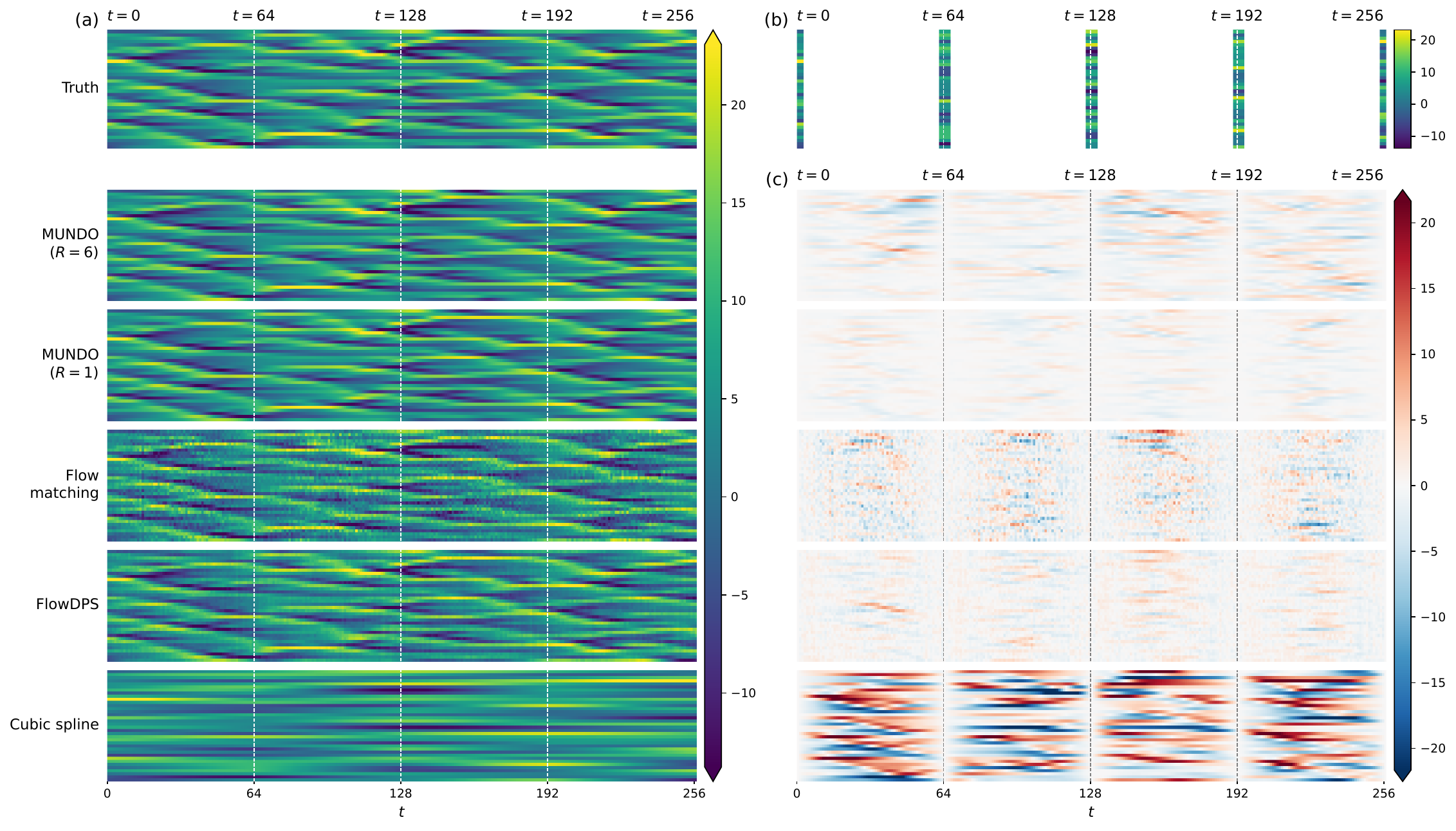}
	\caption{\textbf{Lorenz-96 interpolation at $F=32$, $N=64$}, four consecutive windows, one member per method.
		(a) Truth and reconstructions, slow variables against time.
		(b) Observed endpoint states, widened for visibility.
		(c) Reconstruction errors $\hat{x}-x$; rows follow the method order in (a). Dashed lines mark the observed endpoints.}
	\label{fig:l96-hov-interp}
\end{figure}

Figures~\ref{fig:l96-hov-disagg} and~\ref{fig:l96-hov-interp} show one member of each method at $F=32$, $N=64$ over four consecutive windows, and Figure~\ref{fig:l96-site} in the main text the ensembles at one slow variable. The recursive and single-step \method{} members are smooth in time, like the reference trajectory, whereas both flow-based methods are noisy from one step to the next, most visibly for flow matching under interpolation, which is the excess short-lag variability that $t_1$ measures. FlowDPS meets the observation only through the final projection.

\begin{table}[p]
	\centering
	\caption{\textbf{Lorenz-96 at $N=64$, $F\in\{10,24,64\}$.}
		$^{\dagger}$: causal chain. Bold: best; underlined: second best per metric.}
	\setlength{\tabcolsep}{3pt}
	\begin{subtable}[t]{0.49\textwidth}
		\centering
		\caption{$F=10$ ($N\delta/\tau=2.1$, $\lambda T=0.0$; non-chaotic).}
		\resizebox{\linewidth}{!}{%
			\begin{tabular}{lccccc}
				\toprule
				                              & cons. & CRPS $\downarrow$             & SSR $\to 1$      & cov.\ MAE $\downarrow$ & $t_1 \to 1$      \\
				\midrule
				\multicolumn{6}{l}{\emph{Disaggregation}}                                                                                            \\
				\method{} ($R=6$)$^{\dagger}$ & \yes  & \textbf{0.0085}\sd{0.0011}    & \underline{1.17} & \underline{0.062}      & \textbf{1.00}    \\
				\method{} ($R=1$)             & \yesp & \underline{0.0093}\sd{0.0008} & 1.30             & 0.088                  & \textbf{1.00}    \\
				Flow matching                 & \yesp & 0.0137\sd{0.0008}             & \textbf{1.15}    & \textbf{0.036}         & 1.12             \\
				FlowDPS                       & \yesp & 0.0139\sd{0.0011}             & 0.79             & 0.068                  & \underline{1.07} \\
				Denton--Cholette              & \yes  & 0.3733                        & --               & --                     & 0.76             \\
				CDF-t                         & \no   & 0.6146                        & --               & --                     & 0.72             \\
				\midrule
				\multicolumn{6}{l}{\emph{Interpolation}}                                                                                             \\
				\method{} ($R=6$)             & \yes  & \textbf{0.0073}\sd{0.0001}    & \textbf{1.15}    & \underline{0.053}      & \textbf{1.00}    \\
				\method{} ($R=1$)             & \yesp & \underline{0.0075}\sd{0.0002} & 1.28             & 0.062                  & \textbf{1.00}    \\
				Flow matching                 & \yesp & 0.0260\sd{0.0029}             & 1.89             & 0.162                  & 1.60             \\
				FlowDPS                       & \yesp & 0.0271\sd{0.0008}             & \underline{1.19} & \textbf{0.050}         & 1.47             \\
				Linear                        & \yes  & 0.4084                        & --               & --                     & 0.85             \\
				Cubic spline                  & \yes  & 0.3686                        & --               & --                     & \underline{0.86} \\
				\bottomrule
			\end{tabular}}

		\label{tab:l96-F10-N64}
	\end{subtable}
	\hfill
	\begin{subtable}[t]{0.49\textwidth}
		\centering
		\caption{$F=24$ ($N\delta/\tau=3.9$, $\lambda T=1.8$).}
		\resizebox{\linewidth}{!}{%
			\begin{tabular}{lccccc}
				\toprule
				                              & cons. & CRPS $\downarrow$             & SSR $\to 1$      & cov.\ MAE $\downarrow$ & $t_1 \to 1$      \\
				\midrule
				\multicolumn{6}{l}{\emph{Disaggregation}}                                                                                            \\
				\method{} ($R=6$)$^{\dagger}$ & \yes  & \underline{0.0857}\sd{0.0009} & 0.61             & 0.106                  & \textbf{0.99}    \\
				\method{} ($R=1$)             & \yesp & 0.1201\sd{0.0008}             & 0.63             & 0.098                  & \underline{0.98} \\
				Flow matching                 & \yesp & \textbf{0.0803}\sd{0.0017}    & \underline{0.73} & \underline{0.043}      & \underline{1.02} \\
				FlowDPS                       & \yesp & 0.1085\sd{0.0019}             & \textbf{0.88}    & \textbf{0.024}         & \underline{1.02} \\
				Denton--Cholette              & \yes  & 0.6434                        & --               & --                     & 0.42             \\
				CDF-t                         & \no   & 0.7404                        & --               & --                     & 0.54             \\
				\midrule
				\multicolumn{6}{l}{\emph{Interpolation}}                                                                                             \\
				\method{} ($R=6$)             & \yes  & \underline{0.0356}\sd{0.0015} & 0.89             & \textbf{0.008}         & \textbf{1.00}    \\
				\method{} ($R=1$)             & \yesp & \textbf{0.0261}\sd{0.0002}    & \underline{1.09} & 0.033                  & \textbf{1.00}    \\
				Flow matching                 & \yesp & 0.0517\sd{0.0028}             & 1.32             & 0.107                  & 1.42             \\
				FlowDPS                       & \yesp & 0.0502\sd{0.0010}             & \textbf{0.97}    & \underline{0.029}      & \underline{1.32} \\
				Linear                        & \yes  & 0.7302                        & --               & --                     & 0.56             \\
				Cubic spline                  & \yes  & 0.7709                        & --               & --                     & 0.57             \\
				\bottomrule
			\end{tabular}}
		\label{tab:l96-F24-N64}
	\end{subtable}

	\par\medskip

	\begin{subtable}[t]{0.49\textwidth}
		\centering
		\caption{$F=64$ ($N\delta/\tau=8.3$, $\lambda T=6.5$).}
		\resizebox{\linewidth}{!}{%
			\begin{tabular}{lccccc}
				\toprule
				                              & cons. & CRPS $\downarrow$             & SSR $\to 1$      & cov.\ MAE $\downarrow$ & $t_1 \to 1$      \\
				\midrule
				\multicolumn{6}{l}{\emph{Disaggregation}}                                                                                            \\
				\method{} ($R=6$)$^{\dagger}$ & \yes  & \underline{0.4848}\sd{0.0012} & \underline{0.90} & 0.049                  & \textbf{0.97}    \\
				\method{} ($R=1$)             & \yesp & 0.4886\sd{0.0015}             & \textbf{1.04}    & \textbf{0.037}         & 0.80             \\
				Flow matching                 & \yesp & \textbf{0.4623}\sd{0.0011}    & 0.88             & \underline{0.041}      & \underline{0.93} \\
				FlowDPS                       & \yesp & 0.4856\sd{0.0021}             & 0.86             & 0.052                  & 0.89             \\
				Denton--Cholette              & \yes  & 0.7205                        & --               & --                     & 0.25             \\
				CDF-t                         & \no   & 0.8554                        & --               & --                     & 0.38             \\
				\midrule
				\multicolumn{6}{l}{\emph{Interpolation}}                                                                                             \\
				\method{} ($R=6$)             & \yes  & 0.4006\sd{0.0161}             & \underline{0.91} & \textbf{0.022}         & \underline{0.92} \\
				\method{} ($R=1$)             & \yesp & \underline{0.2772}\sd{0.0019} & 0.88             & 0.054                  & \textbf{0.96}    \\
				Flow matching                 & \yesp & \textbf{0.2747}\sd{0.0054}    & \textbf{0.92}    & 0.049                  & 1.20             \\
				FlowDPS                       & \yesp & 0.2786\sd{0.0022}             & 0.82             & \underline{0.028}      & 1.09             \\
				Linear                        & \yes  & 0.8711                        & --               & --                     & 0.38             \\
				Cubic spline                  & \yes  & 0.9124                        & --               & --                     & 0.39             \\
				\bottomrule
			\end{tabular}}
		\label{tab:l96-F64-N64}
	\end{subtable}
	\label{tab:l96-N64-sweep}
\end{table}

\begin{table}[p]
	\centering
	\caption{\textbf{Lorenz-96 at $N=128$, $F\in\{10,24,32,64\}$.}
		$^{\dagger}$: causal chain. Bold: best; underlined: second best per metric.}
	\setlength{\tabcolsep}{3pt}

	\begin{subtable}[t]{0.49\textwidth}
		\centering
		\caption{$F=10$ ($N\delta/\tau=4.2$, $\lambda T=0.0$; non-chaotic).}
		\resizebox{\linewidth}{!}{%
			\begin{tabular}{lccccc}
				\toprule
				                                & cons. & CRPS $\downarrow$             & SSR $\to 1$      & cov.\ MAE $\downarrow$ & $t_1 \to 1$      \\
				\midrule
				\multicolumn{6}{l}{\emph{Disaggregation}}                                                                                              \\
				\method{} ($R = 7$)$^{\dagger}$ & \yes  & \textbf{0.0150}\sd{0.0021}    & \textbf{1.24}    & \underline{0.062}      & \textbf{1.00}    \\
				\method{} ($R = 1$)             & \yesp & \underline{0.0181}\sd{0.0033} & 1.36             & \textbf{0.055}         & \textbf{1.00}    \\
				Flow matching                   & \yesp & 0.0658\sd{0.0583}             & 6.52             & 0.246                  & 2.63             \\
				FlowDPS                         & \yesp & 0.0308\sd{0.0031}             & \underline{0.70} & 0.112                  & \underline{1.31} \\
				Denton--Cholette                & \yes  & 0.8645                        & --               & --                     & 0.32             \\
				CDF-t                           & \no   & 1.0879                        & --               & --                     & 0.54             \\
				\midrule
				\multicolumn{6}{l}{\emph{Interpolation}}                                                                                               \\
				\method{} ($R = 7$)             & \yes  & \textbf{0.0130}\sd{0.0012}    & \textbf{1.12}    & \textbf{0.043}         & \textbf{1.00}    \\
				\method{} ($R = 1$)             & \yesp & \underline{0.0157}\sd{0.0064} & \underline{1.33} & \underline{0.067}      & \textbf{1.00}    \\
				Flow matching                   & \yesp & 0.1458\sd{0.0029}             & 10.25            & 0.380                  & 4.25             \\
				FlowDPS                         & \yesp & 0.0635\sd{0.0016}             & 3.68             & 0.300                  & 2.71             \\
				Linear                          & \yes  & 0.9766                        & --               & --                     & 0.55             \\
				Cubic spline                    & \yes  & 1.0873                        & --               & --                     & \underline{0.56} \\
				\bottomrule
			\end{tabular}}
		\label{tab:l96-F10-N128}
	\end{subtable}
	\hfill
	\begin{subtable}[t]{0.49\textwidth}
		\centering
		\caption{$F=24$ ($N\delta/\tau=7.9$, $\lambda T=3.7$).}
		\resizebox{\linewidth}{!}{%
			\begin{tabular}{lccccc}
				\toprule
				                              & cons. & CRPS $\downarrow$             & SSR $\to 1$      & cov.\ MAE $\downarrow$ & $t_1 \to 1$      \\
				\midrule
				\multicolumn{6}{l}{\emph{Disaggregation}}                                                                                            \\
				\method{} ($R=7$)$^{\dagger}$ & \yes  & \underline{0.3300}\sd{0.0032} & 0.81             & 0.064                  & \textbf{0.95}    \\
				\method{} ($R=1$)             & \yesp & 0.3596\sd{0.0025}             & \underline{0.87} & \underline{0.038}      & \underline{0.92} \\
				Flow matching                 & \yesp & \textbf{0.3038}\sd{0.0013}    & 0.62             & 0.130                  & 1.32             \\
				FlowDPS                       & \yesp & 0.3353\sd{0.0053}             & \textbf{0.99}    & \textbf{0.009}         & 1.11             \\
				Denton--Cholette              & \yes  & 0.7487                        & --               & --                     & 0.24             \\
				CDF-t                         & \no   & 0.8999                        & --               & --                     & 0.38             \\
				\midrule
				\multicolumn{6}{l}{\emph{Interpolation}}                                                                                             \\
				\method{} ($R = 7$)           & \yes  & 0.2376\sd{0.0027}             & \textbf{0.96}    & \textbf{0.015}         & \underline{0.97} \\
				\method{} ($R = 1$)           & \yesp & \textbf{0.1508}\sd{0.0008}    & \underline{0.78} & \underline{0.025}      & \textbf{1.00}    \\
				Flow matching                 & \yesp & 0.2205\sd{0.0013}             & 2.37             & 0.290                  & 3.66             \\
				FlowDPS                       & \yesp & \underline{0.1562}\sd{0.0025} & \underline{1.22} & 0.170                  & 2.43             \\
				Linear                        & \yes  & 0.8976                        & --               & --                     & 0.39             \\
				Cubic spline                  & \yes  & 0.9434                        & --               & --                     & 0.40             \\
				\bottomrule
			\end{tabular}}
		\label{tab:l96-F24-N128}
	\end{subtable}

	\par\medskip

	\begin{subtable}[t]{0.49\textwidth}
		\centering
		\caption{$F=32$ ($N\delta/\tau=9.8$, $\lambda T=6.2$).}
		\resizebox{\linewidth}{!}{%
			\begin{tabular}{lccccc}
				\toprule
				                                & cons. & CRPS $\downarrow$             & SSR $\to 1$      & cov.\ MAE $\downarrow$ & $t_1 \to 1$      \\
				\midrule
				\multicolumn{6}{l}{\emph{Disaggregation}}                                                                                              \\
				\method{} ($R = 7$)$^{\dagger}$ & \yes  & 0.4928\sd{0.0015}             & 0.86             & 0.061                  & \underline{0.93} \\
				\method{} ($R=1$)               & \yesp & 0.4865\sd{0.0017}             & \textbf{1.00}    & \textbf{0.025}         & 0.83             \\
				Flow matching                   & \yesp & \textbf{0.4608}\sd{0.0008}    & 0.86             & 0.049                  & 1.16             \\
				FlowDPS                         & \yesp & \underline{0.4862}\sd{0.0002} & \underline{0.92} & \underline{0.028}      & \textbf{1.01}    \\
				Denton--Cholette                & \yes  & 0.7488                        & --               & --                     & 0.22             \\
				CDF-t                           & \no   & 0.9005                        & --               & --                     & 0.35             \\
				\midrule
				\multicolumn{6}{l}{\emph{Interpolation}}                                                                                               \\
				\method{} ($R = 7$)             & \yes  & 0.3823\sd{0.0044}             & \underline{0.96} & \textbf{0.006}         & \underline{0.94} \\
				\method{} ($R = 1$)             & \yesp & \underline{0.2991}\sd{0.0039} & 0.93             & \underline{0.058}      & \textbf{0.96}    \\
				Flow matching                   & \yesp & 0.3129\sd{0.0014}             & 1.55             & 0.210                  & 3.27             \\
				FlowDPS                         & \yesp & \textbf{0.2696}\sd{0.0016}    & \textbf{1.02}    & 0.104                  & 2.18             \\
				Linear                          & \yes  & 0.9140                        & --               & --                     & 0.34             \\
				Cubic spline                    & \yes  & 0.9578                        & --               & --                     & 0.35             \\
				\bottomrule
			\end{tabular}}
		\label{tab:l96-F32-N128}
	\end{subtable}
	\hfill
	\begin{subtable}[t]{0.49\textwidth}
		\centering
		\caption{$F=64$ ($N\delta/\tau=16.6$, $\lambda T=13.1$).}
		\resizebox{\linewidth}{!}{%
			\begin{tabular}{lccccc}
				\toprule
				                                & cons. & CRPS $\downarrow$             & SSR $\to 1$      & cov.\ MAE $\downarrow$ & $t_1 \to 1$      \\
				\midrule
				\multicolumn{6}{l}{\emph{Disaggregation}}                                                                                              \\
				\method{} ($R = 7$)$^{\dagger}$ & \yes  & 0.5374\sd{0.0002}             & \textbf{1.02}    & \textbf{0.026}         & \underline{1.06} \\
				\method{} ($R=1$)               & \yesp & 0.5525\sd{0.0041}             & \underline{1.04} & 0.093                  & 0.90             \\
				Flow matching                   & \yesp & \textbf{0.5363}\sd{0.0008}    & 0.87             & \underline{0.047}      & \textbf{0.98}    \\
				FlowDPS                         & \yesp & \underline{0.5371}\sd{0.0008} & 0.86             & 0.050                  & 0.93             \\
				Denton--Cholette                & \yes  & 0.7666                        & --               & --                     & 0.15             \\
				CDF-t                           & \no   & 0.9491                        & --               & --                     & 0.27             \\
				\midrule
				\multicolumn{6}{l}{\emph{Interpolation}}                                                                                               \\
				\method{} ($R = 7$)             & \yes  & 0.5041\sd{0.0043}             & \underline{0.90} & \underline{0.036}      & \textbf{0.97}    \\
				\method{} ($R = 1$)             & \yesp & 0.4782\sd{0.0052}             & \textbf{0.98}    & 0.059                  & \underline{0.79} \\
				Flow matching                   & \yesp & \underline{0.4539}\sd{0.0026} & 1.19             & 0.098                  & 2.40             \\
				FlowDPS                         & \yesp & \textbf{0.4528}\sd{0.0064}    & 0.86             & \textbf{0.021}         & 1.62             \\
				Linear                          & \yes  & 0.9538                        & --               & --                     & 0.27             \\
				Cubic spline                    & \yes  & 1.0006                        & --               & --                     & 0.28             \\
				\bottomrule
			\end{tabular}}
		\label{tab:l96-F64-N128}
	\end{subtable}
	\label{tab:l96-N128-sweep}
\end{table}

\FloatBarrier
\subsection{AutoSim}
\label{app:autosim-results}
\begin{figure}[htbp]
	\centering
	\includegraphics[width=\linewidth]{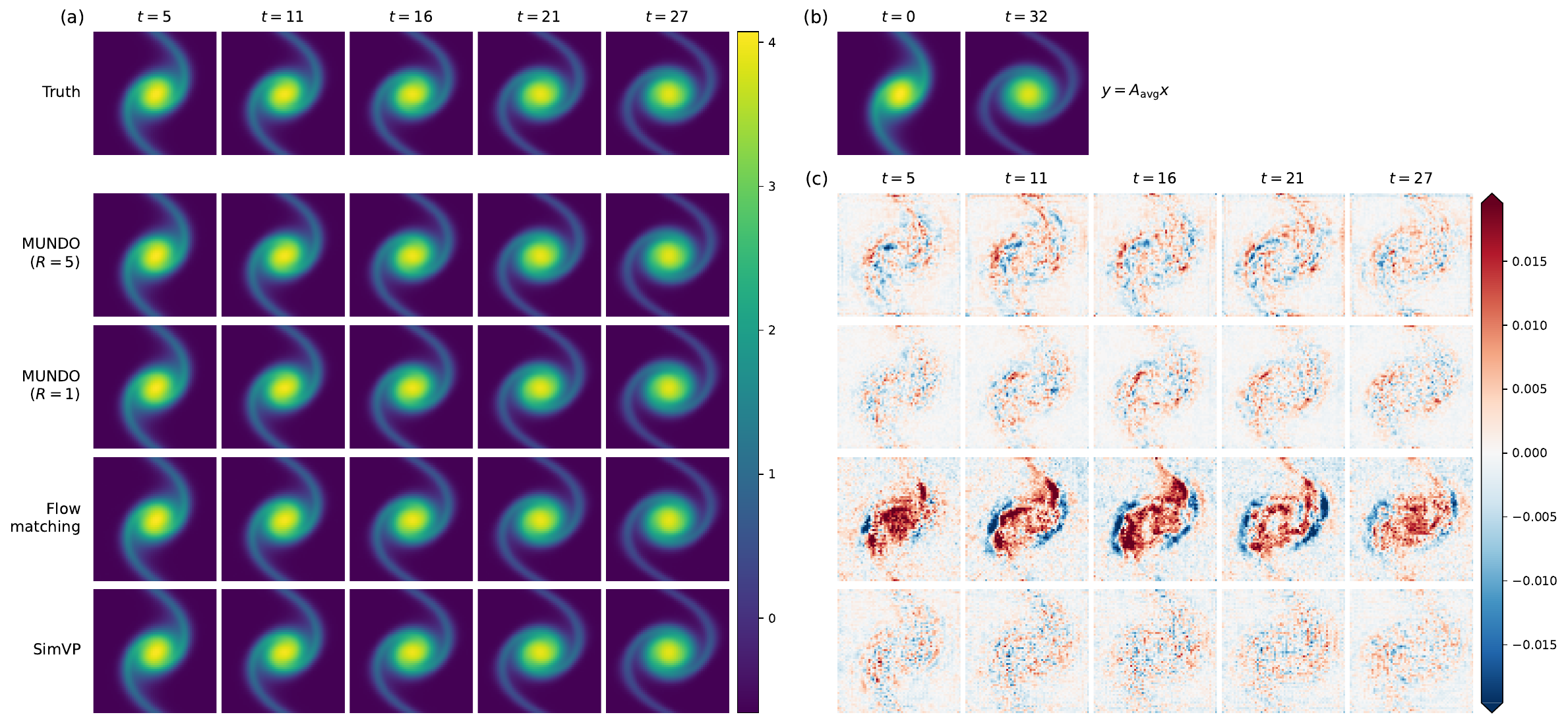}
	\caption{\textbf{Advection-diffusion interpolation at $N=32$}, benchmark tier.
		(a) Ground truth and reconstructions at selected times.
		(b) Observed endpoint states.
		(c) Reconstruction errors $\hat{x}-x$ at selected times; rows follow the method order in (a).}
	\label{fig:advection_interp_app}
\end{figure}

\begin{figure}[htbp]
	\centering
	\includegraphics[width=\linewidth]{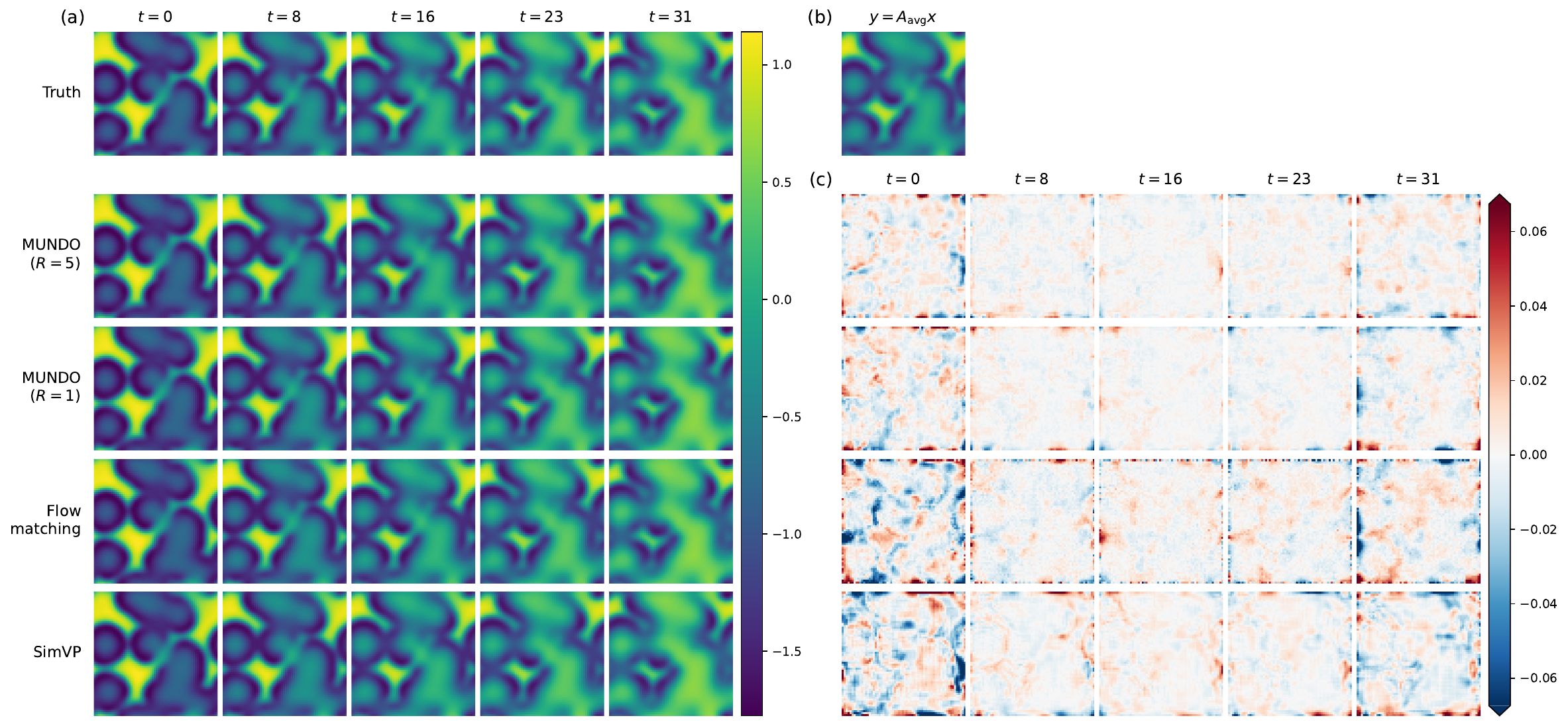}
	\caption{\textbf{Gray--Scott disaggregation at $N=32$}, benchmark tier.
		(a) Ground truth and reconstructions at selected times.
		(b) Coarse observation, given by the temporal mean over the window.
		(c) Reconstruction errors $\hat{x}-x$ at selected times; rows follow the method order in (a).}
	\label{fig:gray_disagg_app}
\end{figure}

Tables~\ref{tab:autosim-L32-chaotic}--\ref{tab:autosim-L8-chaotic} report the remaining AutoSim configurations: the benchmark tier at $N\in\{8,16,24\}$, and the mild and chaotic tiers at $N\in\{8,16,24,32\}$. The mild and chaotic tiers shorten the decorrelation time of the dynamics, through stronger advection, buoyancy, or obstacle forcing, or through a coarser snapshot stride for Gray--Scott (Table~\ref{tab:autosim-gen}), so that a window of fixed length spans more decorrelation times. Panel captions report the corresponding $N\delta/\tau$, defined as for Lorenz-96. In every table, the recursive \method{} variant, plain or fine-tuned, is selected on the validation trajectories (Appendix~\ref{app:nbr-ablation}). Gross--Pitaevskii has no chaotic tier. Figures~\ref{fig:advection_interp_app} and~\ref{fig:gray_disagg_app} show reconstructions on the two systems not illustrated in the main text, with FlowDPS and SimVP added to the rows.

Under interpolation, single-step \method{} attains the lowest CRPS in most settings, whereas recursive \method{} usually attains the lower coverage error. Under disaggregation, single-step \method{} keeps the lower CRPS in most settings, and recursion lowers it where predictability is lowest: on chaotic advection-diffusion at $N\geq16$ by $19$--$39\%$, and on Gross--Pitaevskii at $N\geq24$ in the benchmark tier and $N\geq16$ in the mild tier by $4$--$14\%$. Recursion is the better calibrated sampler: its spread-skill ratio is closer to one in $37$ of the $44$ settings and its coverage error lower in $23$, and it corrects the dispersion of single-step \method{} where it fails, namely overdispersion on Gray--Scott in the benchmark and mild tiers and underdispersion on chaotic advection-diffusion at $N\geq16$. On advection-diffusion, the near-deterministic control, flow matching has the lowest disaggregation CRPS in the chaotic tier at $N\geq16$ and in the mild tier at $N\geq24$, although its ensembles are underdispersed there. Recursive \method{} also slightly overestimates short-lag variability on Navier--Stokes interpolation, where single-step \method{} keeps $t_1$ closer to one. Among the baselines, FlowDPS is the weakest learned method under interpolation. SimVP satisfies the observation and matches the short-lag structure on Gray--Scott and Navier--Stokes, but, being deterministic, its CRPS is about $1.2$ to $10$ times that of \method{}. The classical baselines are the most oversmoothed, particularly on Gross--Pitaevskii.

\FloatBarrier
\subsection{ERA5}
\subsubsection{Non-negative projection of precipitation}
\label{app:posproj}

Table~\ref{tab:posproj} scores the precipitation channel of every method before and after the projection of Equation~\eqref{eq:posproj}. Under disaggregation, the projection lowers the precipitation CRPS of recursive and single-step \method{} by $5\%$ and $4\%$. For the recursive model, it also brings the finest spectral band ratio from 1.56 to 1.05 and $t_1$ from 1.39 to 1.03, while SSR drops from 0.95 to 0.84, since part of the spread came from negative values. Under interpolation, clipping at zero changes the precipitation CRPS of every method by less than $3\%$. The projection does not affect the ranking of the methods in Table~\ref{tab:era5}.

\begin{table}[htbp]
	\centering
	\caption{\textbf{ERA5 total precipitation before and after the non-negative projection.} Neg.\ is the share of generated pixel-hours below zero before projection. CRPS in mm\,h$^{-1}$; Finest is the finest-band spectral power ratio. CDF-t and linear interpolation produce no negative values and are unchanged.}%
	\resizebox{\linewidth}{!}{%
			\begin{tabular}{lccccccccc}
				\toprule
				                    &            & \multicolumn{2}{c}{CRPS $\downarrow$} & \multicolumn{2}{c}{SSR $\to 1$} & \multicolumn{2}{c}{Finest $\to 1$} & \multicolumn{2}{c}{$t_1 \to 1$}                                   \\
				\cmidrule(lr){3-4}\cmidrule(lr){5-6}\cmidrule(lr){7-8}\cmidrule(lr){9-10}
				                    & Neg.\ (\%) & before                                & after                           & before                             & after                           & before & after & before & after \\
				\midrule
				\multicolumn{10}{l}{\emph{Disaggregation}}                                                                                                                                                                          \\
				\method{} ($R = 4$) & 27.9       & 0.1725                                & 0.1634                          & 0.95                               & 0.84                            & 1.56   & 1.05  & 1.39   & 1.03  \\
				\method{} ($R = 1$) & 23.9       & 0.1854                                & 0.1774                          & 0.88                               & 0.84                            & 0.77   & 0.75  & 1.44   & 1.15  \\
				Flow matching       & 20.0       & 0.2324                                & 0.2281                          & 0.49                               & 0.49                            & 0.35   & 0.35  & 1.15   & 0.95  \\
				FlowDPS             & 20.4       & 0.2242                                & 0.2204                          & 0.58                               & 0.56                            & 0.42   & 0.41  & 1.09   & 0.90  \\
				SimVP               & 21.0       & 0.2456                                & 0.2337                          & --                                 & --                              & 0.10   & 0.11  & 0.72   & 0.58  \\
				Denton--Cholette    & 16.6       & 0.3167                                & 0.3007                          & --                                 & --                              & 0.11   & 0.11  & 0.44   & 0.35  \\
				\midrule
				\multicolumn{10}{l}{\emph{Interpolation}}                                                                                                                                                                           \\
				\method{} ($R = 4$) & 20.1       & 0.1728                                & 0.1715                          & 0.88                               & 0.87                            & 0.92   & 0.91  & 1.26   & 1.16  \\
				\method{} ($R = 1$) & 21.1       & 0.1803                                & 0.1770                          & 0.70                               & 0.70                            & 0.47   & 0.46  & 1.50   & 1.36  \\
				Flow matching       & 19.5       & 0.2354                                & 0.2338                          & 0.53                               & 0.52                            & 0.49   & 0.49  & 1.02   & 0.94  \\
				FlowDPS             & 23.1       & 0.2743                                & 0.2726                          & 0.35                               & 0.35                            & 0.07   & 0.07  & 0.95   & 0.85  \\
				SimVP               & 17.3       & 0.2517                                & 0.2456                          & --                                 & --                              & 0.01   & 0.01  & 0.82   & 0.75  \\
				Cubic spline        & 20.5       & 0.3763                                & 0.3666                          & --                                 & --                              & 0.50   & 0.50  & 0.43   & 0.38  \\
				\bottomrule
			\end{tabular}%
		}%
	\label{tab:posproj}
\end{table}

\subsubsection{Hourly scores per variable}
\label{app:pervar}

\begin{figure}[htbp]
	\centering
	\includegraphics[width=\textwidth]{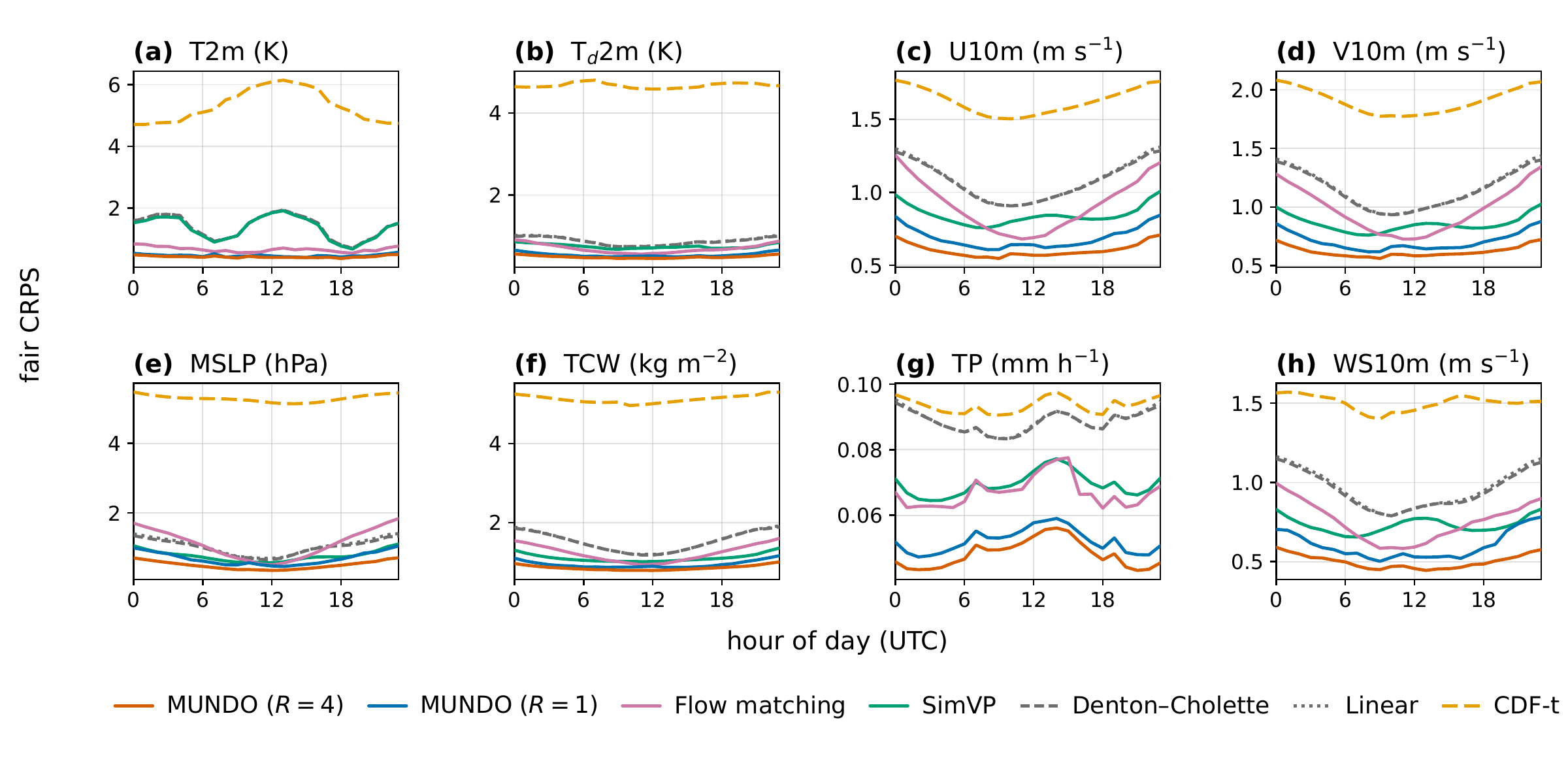}
	\caption{\textbf{Fair CRPS by hour of day under disaggregation.} Per variable in physical units, averaged over the test windows. Wind speed (h) is computed from the two wind components.}%
	\label{fig:hourly_crps_disagg}
\end{figure}

\begin{figure}[htbp]
	\centering
	\includegraphics[width=\textwidth]{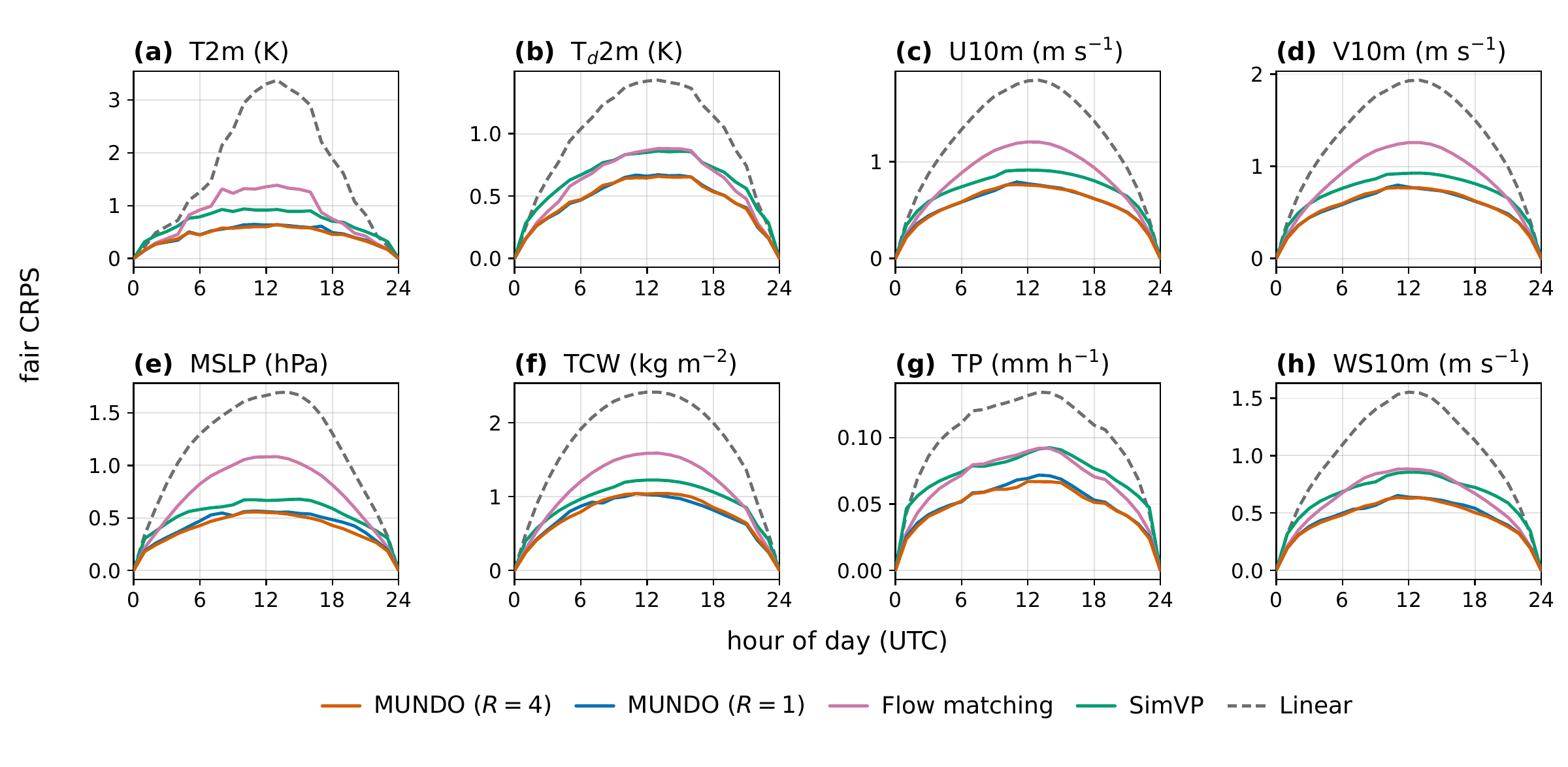}
	\caption{\textbf{Fair CRPS by hour of day under interpolation.} Same construction as Figure~\ref{fig:hourly_crps_disagg}.}%
	\label{fig:hourly_crps_interp}
\end{figure}

Figures~\ref{fig:hourly_crps_disagg} and~\ref{fig:hourly_crps_interp} resolve the fair CRPS by hour of day and by variable, in physical units. Under disaggregation, the ordering of Table~\ref{tab:era5} holds at almost every hour and variable, and the error varies little across the day except for precipitation, which peaks with afternoon convection. Under interpolation, error vanishes at the observed states and peaks near the middle of the window, where the classical interpolants are worst; flow matching falls furthest behind \method{} there, especially for wind and pressure.

\subsubsection{Power spectra}
\label{app:spectra}

\begin{figure}[htbp]
	\centering
	\includegraphics[width=\textwidth]{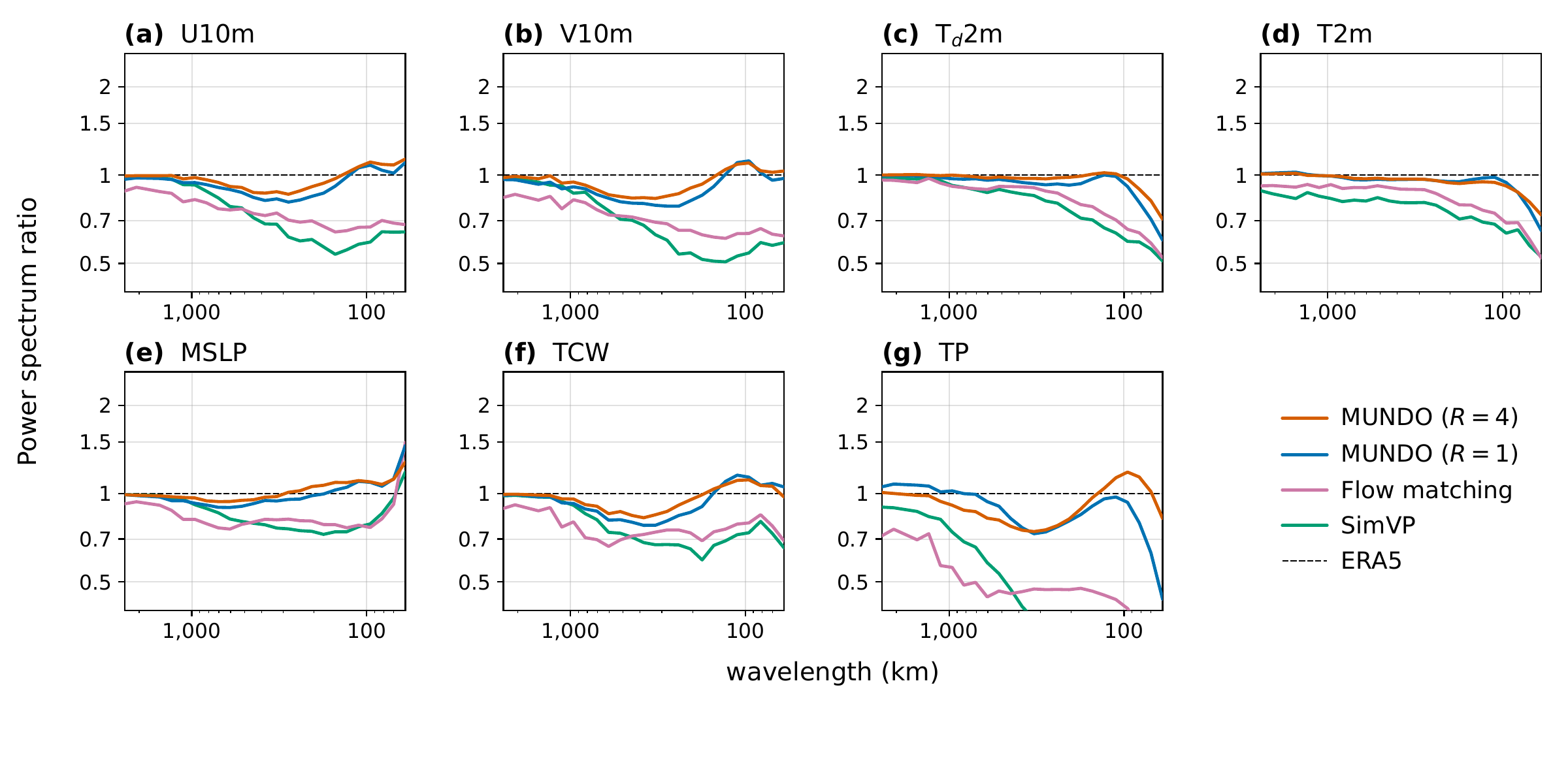}
	\caption{\textbf{Spatial power spectrum ratios under disaggregation.} Ratio of the power spectral density of each method to that of ERA5, per variable, with spectra averaged over test windows and ensemble members. The dashed line marks a ratio of one.}%
	\label{fig:spectra_disagg}
\end{figure}

\begin{figure}[htbp]
	\centering
	\includegraphics[width=\textwidth]{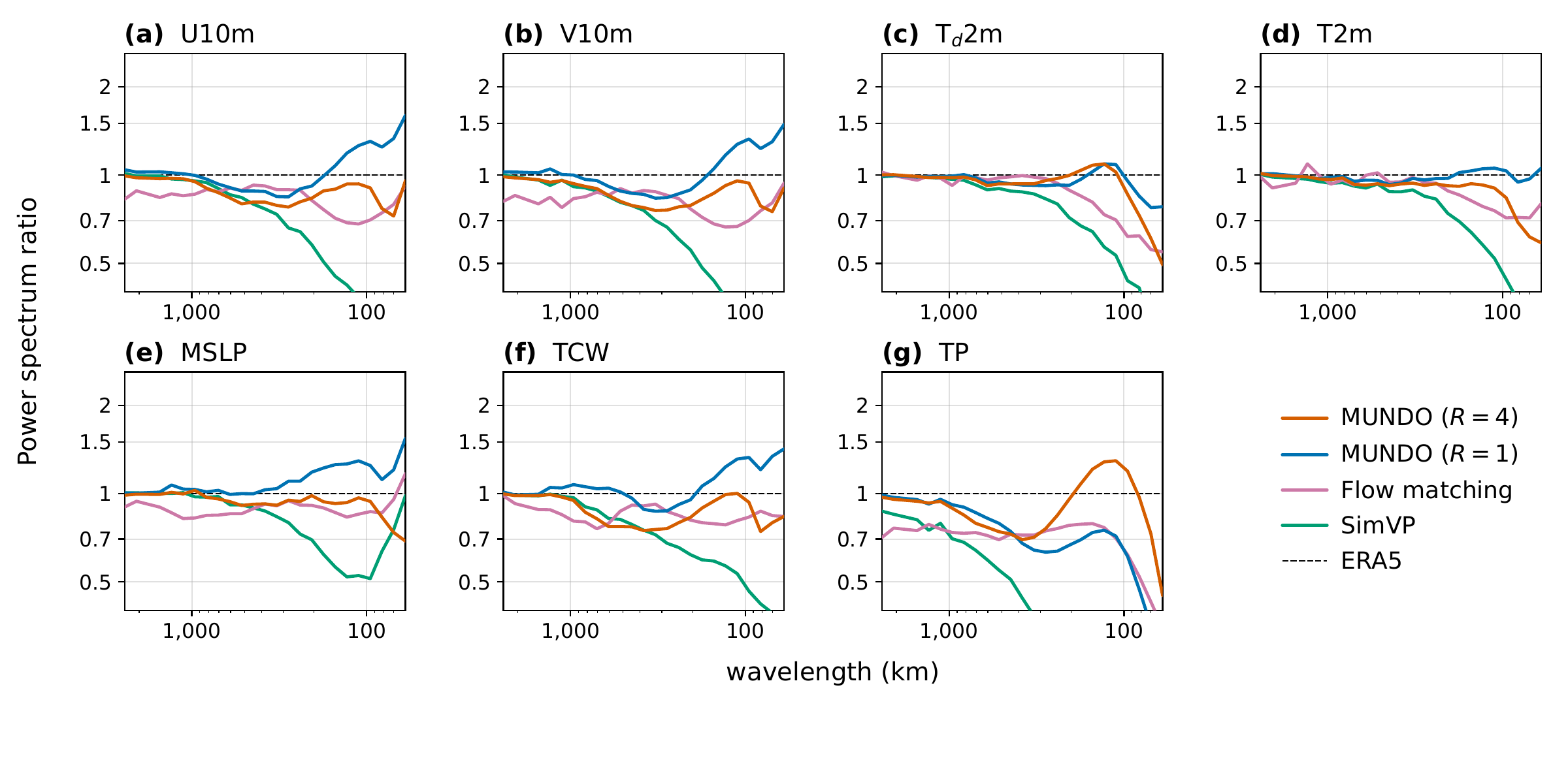}
	\caption{\textbf{Spatial power spectrum ratios under interpolation.} Same construction as Figure~\ref{fig:spectra_disagg}.}%
	\label{fig:spectra_interp}
\end{figure}

The spatial power spectrum measures how variance is distributed across spatial scales, so over-smoothing or excess small-scale noise appears directly as missing or excess fine-scale power.
Figures~\ref{fig:spectra_disagg} and~\ref{fig:spectra_interp} show ratios of power spectra relative to ERA5. Under disaggregation, both \method{} variants remain close to ERA5 across variables, while flow matching and SimVP lose fine-scale power, mildly for winds and dewpoint and most strongly for precipitation, where they remain below the reference across scales. Under interpolation, SimVP undershoots across variables. For precipitation, flow matching and SimVP stay below ERA5 at all scales, while \method{} falls below it at intermediate scales and returns close to it at fine scales.

\subsubsection{Diurnal cycle}
\label{app:diurnal}

\begin{figure}[htbp]
	\centering
	\includegraphics[width=\textwidth]{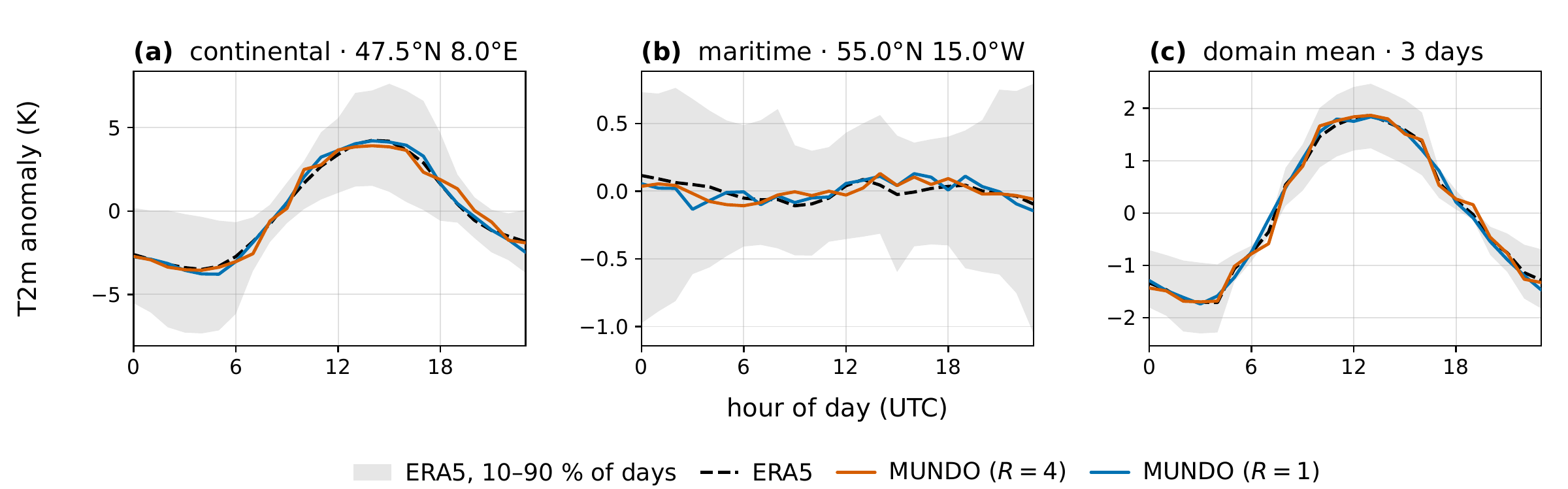}
	\caption{\textbf{Diurnal cycle of 2m temperature under disaggregation.} Shown as anomalies from each day's mean, at a continental pixel, a maritime pixel, and averaged over the domain. Lines are the average over the test days, one member per day; shading is the 10--90\% range of ERA5's day-to-day cycles.}%
	\label{fig:diurnal_disagg}
\end{figure}

\begin{figure}[htbp]
	\centering
	\includegraphics[width=\textwidth]{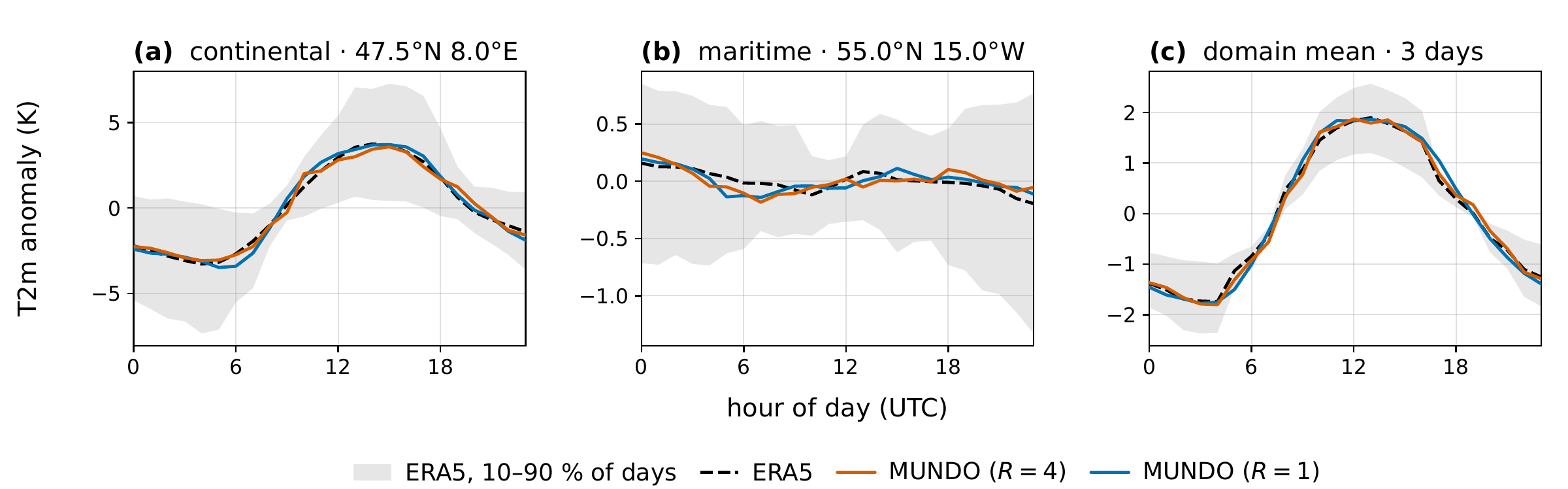}
	\caption{\textbf{Diurnal cycle of 2m temperature under interpolation.} Same construction as Figure~\ref{fig:diurnal_disagg}.}%
	\label{fig:diurnal_interp}
\end{figure}

The diurnal cycle is the dominant sub-daily signal of surface temperature and is absent from a daily mean, so recovering its amplitude and phase tests whether the generated hours carry physically timed structure rather than noise. Figures~\ref{fig:diurnal_disagg} and~\ref{fig:diurnal_interp} show the mean diurnal cycle of 2m temperature, as anomalies from each day's mean, at a continental pixel, a maritime pixel, and averaged over the domain. Both \method{} variants reproduce the amplitude and phase of the mean cycle on both tasks, including the nearly flat maritime cycle. %

\FloatBarrier
\subsection{Full AutoSim tables}
\label{app:autosim-tables}
In all tables below, $^{\ddagger}$ marks end-to-end fine-tuned models; bold entries are best and underlined entries second best per metric.

\begin{table}[htp]
	\centering
	\caption{\textbf{AutoSim at $N = 32$, most chaotic tier of each system.}}%
	\setlength{\tabcolsep}{4pt}
	\begin{subtable}[t]{0.49\textwidth}
		\centering
		\caption{\textbf{Advection-diffusion ($N\delta/\tau = 4.0$)}}%
		\resizebox{\linewidth}{!}{%
\begin{tabular}{lccccc}
				\toprule
				                                 & cons. & CRPS $\downarrow$             & SSR $\to 1$      & cov.\ MAE $\downarrow$ & $t_1 \to 1$      \\
				\midrule
				\multicolumn{6}{l}{\emph{Disaggregation}}                                                                                               \\
				\method{} ($R = 5$)$^{\ddagger}$ & \yes  & \underline{0.0263}\sd{0.0008} & \textbf{1.02}    & \textbf{0.011}         & \underline{0.94} \\
				\method{} ($R = 1$)              & \yesp & 0.0432\sd{0.0063}             & 0.69             & \underline{0.038}      & 0.71             \\
				Flow matching                    & \yesp & \textbf{0.0222}\sd{0.0002}    & \underline{0.74} & 0.049                  & \textbf{0.95}    \\
				FlowDPS                          & \yesp & 0.0441\sd{0.0018}             & 0.54             & 0.159                  & 0.77             \\
				SimVP                            & \yesp & 0.0331                        & --               & --                     & 0.78             \\
				Denton--Cholette                 & \yes  & 0.0986                        & --               & --                     & 0.33             \\
\bottomrule
			\end{tabular}}%
		\label{tab:autosim-advection-diffusion-L32-chaotic-sub}
	\end{subtable}\hfill
	\begin{subtable}[t]{0.49\textwidth}
		\centering
		\caption{\textbf{Gray--Scott ($N\delta/\tau = 3.2$)}}%
		\resizebox{\linewidth}{!}{%
			\begin{tabular}{lccccc}
				\toprule
				                                 & cons. & CRPS $\downarrow$             & SSR $\to 1$      & cov.\ MAE $\downarrow$ & $t_1 \to 1$      \\
				\midrule
				\multicolumn{6}{l}{\emph{Disaggregation}}                                                                                               \\
				\method{} ($R = 5$)$^{\ddagger}$ & \yes  & 0.0199\sd{0.0006}             & \textbf{1.03}    & 0.047                  & \underline{1.01} \\
				\method{} ($R = 1$)              & \yesp & \textbf{0.0146}\sd{0.0001}    & 0.82             & \underline{0.025}      & \underline{1.01} \\
				Flow matching                    & \yesp & \underline{0.0173}\sd{0.0003} & 0.89             & 0.082                  & 1.02             \\
				FlowDPS                          & \yesp & 0.0467\sd{0.0009}             & \underline{0.95} & \textbf{0.015}         & 1.02             \\
				SimVP                            & \yesp & 0.0309                        & --               & --                     & \textbf{1.00}    \\
				Denton--Cholette                 & \yes  & 0.6247                        & --               & --                     & 0.56             \\
\bottomrule
			\end{tabular}}%
		\label{tab:autosim-gray-scott-L32-chaotic-sub}
	\end{subtable}\par\medskip
	\begin{subtable}[t]{0.49\textwidth}
		\centering
		\caption{\textbf{Navier--Stokes ($N\delta/\tau = 1.5$)}}%
		\resizebox{\linewidth}{!}{%
			\begin{tabular}{lccccc}
				\toprule
				                                 & cons. & CRPS $\downarrow$             & SSR $\to 1$      & cov.\ MAE $\downarrow$ & $t_1 \to 1$      \\
				\midrule
				\multicolumn{6}{l}{\emph{Disaggregation}}                                                                                               \\
				\method{} ($R = 5$)$^{\ddagger}$ & \yes  & 0.1076\sd{0.0008}             & \textbf{0.82}    & \textbf{0.027}         & 1.24             \\
				\method{} ($R = 1$)              & \yesp & \underline{0.0977}\sd{0.0038} & \underline{0.76} & \textbf{0.027}         & \textbf{0.96}    \\
				Flow matching                    & \yesp & \textbf{0.0948}\sd{0.0008}    & 0.74             & \underline{0.045}      & 1.29             \\
				FlowDPS                          & \yesp & 0.1313\sd{0.0018}             & 0.49             & 0.081                  & \underline{1.06} \\
				SimVP                            & \yesp & 0.1352                        & --               & --                     & \underline{0.94} \\
				Denton--Cholette                 & \yes  & 0.2178                        & --               & --                     & 0.62             \\
\bottomrule
			\end{tabular}}%
		\label{tab:autosim-conditioned-ns-L32-chaotic-sub}
	\end{subtable}
	\label{tab:autosim-L32-chaotic}
\end{table}

\begin{table}[htp]
	\centering
	\caption{\textbf{AutoSim at $N = 24$.}}%
	\setlength{\tabcolsep}{4pt}
	\begin{subtable}[t]{0.49\textwidth}
		\centering
		\caption{\textbf{Advection-diffusion}}%
		\resizebox{\linewidth}{!}{%
			\begin{tabular}{lccccc}
				\toprule
				                    & cons. & CRPS $\downarrow$             & SSR $\to 1$      & cov.\ MAE $\downarrow$ & $t_1 \to 1$      \\
				\midrule
				\multicolumn{6}{l}{\emph{Disaggregation}}                                                                                  \\
				\method{} ($R = 4$) & \yes  & \underline{0.0020}\sd{0.0000} & \textbf{0.93}    & \underline{0.114}      & \textbf{1.02}    \\
				\method{} ($R = 1$) & \yesp & \textbf{0.0016}\sd{0.0000}    & \underline{0.84} & 0.158                  & 1.05             \\
				Flow matching       & \yesp & 0.0022\sd{0.0000}             & 0.81             & 0.184                  & 1.19             \\
				FlowDPS             & \yesp & 0.0049\sd{0.0001}             & 0.48             & \textbf{0.093}         & 1.20             \\
				SimVP               & \yesp & 0.0037                        & --               & --                     & \underline{1.04} \\
				Denton--Cholette    & \yes  & 0.0274                        & --               & --                     & 0.93             \\
\bottomrule
			\end{tabular}}%
		\label{tab:autosim-advection-diffusion-L24-base-sub}
	\end{subtable}\hfill
	\begin{subtable}[t]{0.49\textwidth}
		\centering
		\caption{\textbf{Gray--Scott}}%
		\resizebox{\linewidth}{!}{%
			\begin{tabular}{lccccc}
				\toprule
				                    & cons. & CRPS $\downarrow$             & SSR $\to 1$      & cov.\ MAE $\downarrow$ & $t_1 \to 1$      \\
				\midrule
				\multicolumn{6}{l}{\emph{Disaggregation}}                                                                                  \\
				\method{} ($R = 4$) & \yes  & \textbf{0.0028}\sd{0.0000}    & \underline{1.10} & \textbf{0.043}         & \textbf{1.00}    \\
				\method{} ($R = 1$) & \yesp & \textbf{0.0028}\sd{0.0000}    & 1.35             & \underline{0.048}      & \underline{1.01} \\
				Flow matching       & \yesp & \underline{0.0040}\sd{0.0000} & \textbf{0.94}    & 0.128                  & 1.06             \\
				FlowDPS             & \yesp & 0.0067\sd{0.0000}             & 0.79             & 0.089                  & 1.07             \\
				SimVP               & \yesp & 0.0081                        & --               & --                     & \underline{1.01} \\
				Denton--Cholette    & \yes  & 0.0805                        & --               & --                     & 0.92             \\
\bottomrule
			\end{tabular}}%
		\label{tab:autosim-gray-scott-L24-base-sub}
	\end{subtable}\par\medskip
	\begin{subtable}[t]{0.49\textwidth}
		\centering
		\caption{\textbf{Navier--Stokes}}%
		\resizebox{\linewidth}{!}{%
			\begin{tabular}{lccccc}
				\toprule
				                                 & cons. & CRPS $\downarrow$             & SSR $\to 1$      & cov.\ MAE $\downarrow$ & $t_1 \to 1$      \\
				\midrule
				\multicolumn{6}{l}{\emph{Disaggregation}}                                                                                               \\
				\method{} ($R = 4$)$^{\ddagger}$ & \yes  & \underline{0.0088}\sd{0.0000} & \underline{0.75} & \textbf{0.020}         & \underline{1.05} \\
				\method{} ($R = 1$)              & \yesp & \textbf{0.0077}\sd{0.0000}    & 0.62             & 0.052                  & \textbf{1.02}    \\
				Flow matching                    & \yesp & 0.0116\sd{0.0001}             & \textbf{0.85}    & \underline{0.028}      & 1.14             \\
				FlowDPS                          & \yesp & 0.0160\sd{0.0001}             & 0.60             & 0.045                  & 1.11             \\
				SimVP                            & \yesp & 0.0200                        & --               & --                     & \textbf{1.02}    \\
				Denton--Cholette                 & \yes  & 0.0570                        & --               & --                     & 0.94             \\
\bottomrule
			\end{tabular}}%
		\label{tab:autosim-conditioned-ns-L24-base-sub}
	\end{subtable}\hfill
	\begin{subtable}[t]{0.49\textwidth}
		\centering
		\caption{\textbf{Gross--Pitaevskii}}%
		\resizebox{\linewidth}{!}{%
			\begin{tabular}{lccccc}
				\toprule
				                                 & cons. & CRPS $\downarrow$             & SSR $\to 1$      & cov.\ MAE $\downarrow$ & $t_1 \to 1$      \\
				\midrule
				\multicolumn{6}{l}{\emph{Disaggregation}}                                                                                               \\
				\method{} ($R = 4$)$^{\ddagger}$ & \yes  & \textbf{0.0799}\sd{0.0010}    & \textbf{0.92}    & 0.065                  & \underline{1.06} \\
				\method{} ($R = 1$)              & \yesp & \underline{0.0828}\sd{0.0008} & 0.83             & \textbf{0.057}         & \textbf{0.98}    \\
				Flow matching                    & \yesp & 0.1001\sd{0.0011}             & 0.88             & 0.089                  & \underline{1.06} \\
				FlowDPS                          & \yesp & 0.1402\sd{0.0006}             & \underline{0.90} & \underline{0.058}      & \textbf{0.98}    \\
				SimVP                            & \yesp & 0.1536                        & --               & --                     & 0.91             \\
				Denton--Cholette                 & \yes  & 0.4900                        & --               & --                     & 0.29             \\
\bottomrule
			\end{tabular}}%
		\label{tab:autosim-gpe-wake-L24-base-sub}
	\end{subtable}
	\label{tab:autosim-L24-base}
\end{table}

\begin{table}[htp]
	\centering
	\caption{\textbf{AutoSim at $N = 16$.}}%
	\setlength{\tabcolsep}{4pt}
	\begin{subtable}[t]{0.49\textwidth}
		\centering
		\caption{\textbf{Advection-diffusion}}%
		\resizebox{\linewidth}{!}{%
			\begin{tabular}{lccccc}
				\toprule
				                    & cons. & CRPS $\downarrow$             & SSR $\to 1$      & cov.\ MAE $\downarrow$ & $t_1 \to 1$      \\
				\midrule
				\multicolumn{6}{l}{\emph{Disaggregation}}                                                                                  \\
				\method{} ($R = 4$) & \yes  & \underline{0.0016}\sd{0.0000} & \textbf{0.94}    & \textbf{0.132}         & \textbf{1.03}    \\
				\method{} ($R = 1$) & \yesp & \textbf{0.0012}\sd{0.0000}    & \underline{0.92} & \underline{0.175}      & \underline{1.05} \\
				Flow matching       & \yesp & \underline{0.0016}\sd{0.0000} & 0.79             & 0.199                  & 1.16             \\
				FlowDPS             & \yesp & 0.0036\sd{0.0001}             & 0.47             & \textbf{0.132}         & 1.17             \\
				SimVP               & \yesp & 0.0040                        & --               & --                     & 1.08             \\
				Denton--Cholette    & \yes  & 0.0127                        & --               & --                     & 0.94             \\
\bottomrule
			\end{tabular}}%
		\label{tab:autosim-advection-diffusion-L16-base-sub}
	\end{subtable}\hfill
	\begin{subtable}[t]{0.49\textwidth}
		\centering
		\caption{\textbf{Gray--Scott}}%
		\resizebox{\linewidth}{!}{%
			\begin{tabular}{lccccc}
				\toprule
				                    & cons. & CRPS $\downarrow$             & SSR $\to 1$      & cov.\ MAE $\downarrow$ & $t_1 \to 1$      \\
				\midrule
				\multicolumn{6}{l}{\emph{Disaggregation}}                                                                                  \\
				\method{} ($R = 4$) & \yes  & \textbf{0.0023}\sd{0.0000}    & \underline{1.19} & \textbf{0.053}         & \textbf{1.00}    \\
				\method{} ($R = 1$) & \yesp & \underline{0.0025}\sd{0.0000} & 1.26             & \underline{0.057}      & \underline{1.01} \\
				Flow matching       & \yesp & 0.0031\sd{0.0001}             & \textbf{1.00}    & 0.138                  & 1.05             \\
				FlowDPS             & \yesp & 0.0051\sd{0.0001}             & 0.75             & 0.106                  & 1.06             \\
				SimVP               & \yesp & 0.0081                        & --               & --                     & \underline{1.01} \\
				Denton--Cholette    & \yes  & 0.0536                        & --               & --                     & 0.89             \\
\bottomrule
			\end{tabular}}%
		\label{tab:autosim-gray-scott-L16-base-sub}
	\end{subtable}\par\medskip
	\begin{subtable}[t]{0.49\textwidth}
		\centering
		\caption{\textbf{Navier--Stokes}}%
		\resizebox{\linewidth}{!}{%
			\begin{tabular}{lccccc}
				\toprule
				                                 & cons. & CRPS $\downarrow$             & SSR $\to 1$      & cov.\ MAE $\downarrow$ & $t_1 \to 1$      \\
				\midrule
				\multicolumn{6}{l}{\emph{Disaggregation}}                                                                                               \\
				\method{} ($R = 4$)$^{\ddagger}$ & \yes  & \underline{0.0043}\sd{0.0000} & \textbf{0.85}    & 0.039                  & 1.05             \\
				\method{} ($R = 1$)              & \yesp & \textbf{0.0037}\sd{0.0000}    & \underline{0.74} & \textbf{0.022}         & \textbf{1.03}    \\
				Flow matching                    & \yesp & 0.0056\sd{0.0001}             & 0.73             & 0.031                  & 1.07             \\
				FlowDPS                          & \yesp & 0.0076\sd{0.0001}             & 0.59             & \underline{0.030}      & 1.08             \\
				SimVP                            & \yesp & 0.0127                        & --               & --                     & \textbf{1.03}    \\
				Denton--Cholette                 & \yes  & 0.0222                        & --               & --                     & \underline{0.96} \\
\bottomrule
			\end{tabular}}%
		\label{tab:autosim-conditioned-ns-L16-base-sub}
	\end{subtable}\hfill
	\begin{subtable}[t]{0.49\textwidth}
		\centering
		\caption{\textbf{Gross--Pitaevskii}}%
		\resizebox{\linewidth}{!}{%
			\begin{tabular}{lccccc}
				\toprule
				                                 & cons. & CRPS $\downarrow$             & SSR $\to 1$      & cov.\ MAE $\downarrow$ & $t_1 \to 1$      \\
				\midrule
				\multicolumn{6}{l}{\emph{Disaggregation}}                                                                                               \\
				\method{} ($R = 4$)$^{\ddagger}$ & \yes  & \underline{0.0509}\sd{0.0004} & \textbf{0.96}    & \underline{0.073}      & 1.06             \\
				\method{} ($R = 1$)              & \yesp & \textbf{0.0490}\sd{0.0005}    & \underline{0.91} & 0.079                  & \textbf{1.03}    \\
				Flow matching                    & \yesp & 0.0577\sd{0.0012}             & 0.88             & 0.080                  & \textbf{1.03}    \\
				FlowDPS                          & \yesp & 0.0898\sd{0.0008}             & 0.75             & \textbf{0.049}         & \underline{0.96} \\
				SimVP                            & \yesp & 0.1023                        & --               & --                     & 0.95             \\
				Denton--Cholette                 & \yes  & 0.4263                        & --               & --                     & 0.42             \\
\bottomrule
			\end{tabular}}%
		\label{tab:autosim-gpe-wake-L16-base-sub}
	\end{subtable}
	\label{tab:autosim-L16-base}
\end{table}

\begin{table}[htp]
	\centering
	\caption{\textbf{AutoSim at $N = 8$.}}%
	\setlength{\tabcolsep}{4pt}
	\begin{subtable}[t]{0.49\textwidth}
		\centering
		\caption{\textbf{Advection-diffusion}}%
		\resizebox{\linewidth}{!}{%
			\begin{tabular}{lccccc}
				\toprule
				                                 & cons. & CRPS $\downarrow$             & SSR $\to 1$      & cov.\ MAE $\downarrow$ & $t_1 \to 1$      \\
				\midrule
				\multicolumn{6}{l}{\emph{Disaggregation}}                                                                                               \\
				\method{} ($R = 3$)$^{\ddagger}$ & \yes  & \textbf{0.0004}\sd{0.0000}    & \underline{0.89} & \textbf{0.178}         & \textbf{1.02}    \\
				\method{} ($R = 1$)              & \yesp & \underline{0.0005}\sd{0.0000} & \textbf{0.96}    & 0.219                  & \underline{1.05} \\
				Flow matching                    & \yesp & 0.0009\sd{0.0000}             & 0.68             & 0.238                  & 1.13             \\
				FlowDPS                          & \yesp & 0.0015\sd{0.0000}             & 0.53             & \underline{0.192}      & 1.14             \\
				SimVP                            & \yesp & 0.0027                        & --               & --                     & \underline{1.05} \\
				Denton--Cholette                 & \yes  & 0.0026                        & --               & --                     & \textbf{0.98}    \\
\bottomrule
			\end{tabular}}%
		\label{tab:autosim-advection-diffusion-L8-base-sub}
	\end{subtable}\hfill
	\begin{subtable}[t]{0.49\textwidth}
		\centering
		\caption{\textbf{Gray--Scott}}%
		\resizebox{\linewidth}{!}{%
			\begin{tabular}{lccccc}
				\toprule
				                                 & cons. & CRPS $\downarrow$             & SSR $\to 1$      & cov.\ MAE $\downarrow$ & $t_1 \to 1$      \\
				\midrule
				\multicolumn{6}{l}{\emph{Disaggregation}}                                                                                               \\
				\method{} ($R = 3$)$^{\ddagger}$ & \yes  & \textbf{0.0011}\sd{0.0000}    & 1.84             & \underline{0.092}      & \textbf{1.00}    \\
				\method{} ($R = 1$)              & \yesp & \underline{0.0012}\sd{0.0000} & \underline{1.22} & \textbf{0.091}         & \underline{1.01} \\
				Flow matching                    & \yesp & 0.0014\sd{0.0000}             & \textbf{0.95}    & 0.183                  & 1.03             \\
				FlowDPS                          & \yesp & 0.0019\sd{0.0000}             & \underline{0.78} & 0.162                  & 1.04             \\
				SimVP                            & \yesp & 0.0049                        & --               & --                     & \underline{1.01} \\
				Denton--Cholette                 & \yes  & 0.0204                        & --               & --                     & 0.95             \\
\bottomrule
			\end{tabular}}%
		\label{tab:autosim-gray-scott-L8-base-sub}
	\end{subtable}\par\medskip
	\begin{subtable}[t]{0.49\textwidth}
		\centering
		\caption{\textbf{Navier--Stokes}}%
		\resizebox{\linewidth}{!}{%
			\begin{tabular}{lccccc}
				\toprule
				                    & cons. & CRPS $\downarrow$             & SSR $\to 1$      & cov.\ MAE $\downarrow$ & $t_1 \to 1$      \\
				\midrule
				\multicolumn{6}{l}{\emph{Disaggregation}}                                                                                  \\
				\method{} ($R = 3$) & \yes  & \textbf{0.0018}\sd{0.0000}    & \textbf{0.93}    & 0.067                  & 1.04             \\
				\method{} ($R = 1$) & \yesp & \textbf{0.0018}\sd{0.0000}    & \underline{0.84} & \underline{0.049}      & 1.03             \\
				Flow matching       & \yesp & \underline{0.0024}\sd{0.0000} & 0.69             & 0.050                  & 1.05             \\
				FlowDPS             & \yesp & 0.0026\sd{0.0000}             & 0.63             & \textbf{0.043}         & 1.06             \\
				SimVP               & \yesp & 0.0066                        & --               & --                     & \underline{1.02} \\
				Denton--Cholette    & \yes  & 0.0041                        & --               & --                     & \textbf{0.99}    \\
\bottomrule
			\end{tabular}}%
		\label{tab:autosim-conditioned-ns-L8-base-sub}
	\end{subtable}\hfill
	\begin{subtable}[t]{0.49\textwidth}
		\centering
		\caption{\textbf{Gross--Pitaevskii}}%
		\resizebox{\linewidth}{!}{%
			\begin{tabular}{lccccc}
				\toprule
				                                 & cons. & CRPS $\downarrow$             & SSR $\to 1$      & cov.\ MAE $\downarrow$ & $t_1 \to 1$      \\
				\midrule
				\multicolumn{6}{l}{\emph{Disaggregation}}                                                                                               \\
				\method{} ($R = 3$)$^{\ddagger}$ & \yes  & \underline{0.0169}\sd{0.0003} & \textbf{0.96}    & 0.094                  & 1.04             \\
				\method{} ($R = 1$)              & \yesp & \textbf{0.0152}\sd{0.0000}    & \underline{0.92} & \underline{0.087}      & 1.04             \\
				Flow matching                    & \yesp & 0.0188\sd{0.0002}             & 0.88             & 0.089                  & 1.04             \\
				FlowDPS                          & \yesp & 0.0407\sd{0.0002}             & 0.71             & \textbf{0.048}         & \textbf{1.01}    \\
				SimVP                            & \yesp & 0.0408                        & --               & --                     & \underline{1.02} \\
				Denton--Cholette                 & \yes  & 0.2263                        & --               & --                     & 0.69             \\
\bottomrule
			\end{tabular}}%
		\label{tab:autosim-gpe-wake-L8-base-sub}
	\end{subtable}
	\label{tab:autosim-L8-base}
\end{table}

\begin{table}[htp]
	\centering
	\caption{\textbf{AutoSim at $N = 32$, mild tier of each system.}}%
	\setlength{\tabcolsep}{4pt}
	\begin{subtable}[t]{0.49\textwidth}
		\centering
		\caption{\textbf{Advection-diffusion ($N\delta/\tau = 2.0$)}}%
		\resizebox{\linewidth}{!}{%
			\begin{tabular}{lccccc}
				\toprule
				                                 & cons. & CRPS $\downarrow$             & SSR $\to 1$      & cov.\ MAE $\downarrow$ & $t_1 \to 1$      \\
				\midrule
				\multicolumn{6}{l}{\emph{Disaggregation}}                                                                                               \\
				\method{} ($R = 5$)              & \yes  & 0.0091\sd{0.0001}             & \textbf{1.02}    & \textbf{0.022}         & \underline{1.02} \\
				\method{} ($R = 1$)              & \yesp & \underline{0.0082}\sd{0.0001} & 0.54             & \underline{0.049}      & \underline{0.98} \\
				Flow matching                    & \yesp & \textbf{0.0074}\sd{0.0000}    & \underline{0.78} & 0.070                  & 1.08             \\
				FlowDPS                          & \yesp & 0.0183\sd{0.0006}             & 0.42             & 0.054                  & \textbf{1.01}    \\
				SimVP                            & \yesp & 0.0127                        & --               & --                     & 0.96             \\
				Denton--Cholette                 & \yes  & 0.0748                        & --               & --                     & 0.65             \\
\bottomrule
			\end{tabular}}%
		\label{tab:autosim-advection-diffusion-L32-mild-sub}
	\end{subtable}\hfill
	\begin{subtable}[t]{0.49\textwidth}
		\centering
		\caption{\textbf{Gray--Scott ($N\delta/\tau = 1.6$)}}%
		\resizebox{\linewidth}{!}{%
			\begin{tabular}{lccccc}
				\toprule
				                                 & cons. & CRPS $\downarrow$             & SSR $\to 1$      & cov.\ MAE $\downarrow$ & $t_1 \to 1$      \\
				\midrule
				\multicolumn{6}{l}{\emph{Disaggregation}}                                                                                               \\
				\method{} ($R = 5$)              & \yes  & \underline{0.0088}\sd{0.0002} & 0.79             & \underline{0.043}      & \underline{1.01} \\
				\method{} ($R = 1$)              & \yesp & \textbf{0.0066}\sd{0.0001}    & 1.17             & \textbf{0.042}         & \underline{1.01} \\
				Flow matching                    & \yesp & \underline{0.0088}\sd{0.0001} & \textbf{0.91}    & 0.119                  & 1.04             \\
				FlowDPS                          & \yesp & 0.0187\sd{0.0003}             & \underline{1.10} & 0.049                  & 1.04             \\
				SimVP                            & \yesp & 0.0147                        & --               & --                     & \textbf{1.00}    \\
				Denton--Cholette                 & \yes  & 0.2933                        & --               & --                     & 0.90             \\
\bottomrule
			\end{tabular}}%
		\label{tab:autosim-gray-scott-L32-mild-sub}
	\end{subtable}\par\medskip
	\begin{subtable}[t]{0.49\textwidth}
		\centering
		\caption{\textbf{Navier--Stokes ($N\delta/\tau = 1.3$)}}%
		\resizebox{\linewidth}{!}{%
			\begin{tabular}{lccccc}
				\toprule
				                                 & cons. & CRPS $\downarrow$             & SSR $\to 1$      & cov.\ MAE $\downarrow$ & $t_1 \to 1$      \\
				\midrule
				\multicolumn{6}{l}{\emph{Disaggregation}}                                                                                               \\
				\method{} ($R = 5$)$^{\ddagger}$ & \yes  & 0.0757\sd{0.0022}             & \underline{0.75} & \textbf{0.021}         & 1.18             \\
				\method{} ($R = 1$)              & \yesp & \underline{0.0718}\sd{0.0030} & 0.72             & \underline{0.032}      & \textbf{0.99}    \\
				Flow matching                    & \yesp & \textbf{0.0703}\sd{0.0005}    & \textbf{0.79}    & 0.051                  & 1.29             \\
				FlowDPS                          & \yesp & 0.1049\sd{0.0009}             & 0.49             & 0.073                  & 1.11             \\
				SimVP                            & \yesp & 0.1014                        & --               & --                     & \underline{0.97} \\
				Denton--Cholette                 & \yes  & 0.1989                        & --               & --                     & 0.69             \\
\bottomrule
			\end{tabular}}%
		\label{tab:autosim-conditioned-ns-L32-mild-sub}
	\end{subtable}\hfill
	\begin{subtable}[t]{0.49\textwidth}
		\centering
		\caption{\textbf{Gross--Pitaevskii ($N\delta/\tau = 4.0$)}}%
		\resizebox{\linewidth}{!}{%
			\begin{tabular}{lccccc}
				\toprule
				                                 & cons. & CRPS $\downarrow$             & SSR $\to 1$      & cov.\ MAE $\downarrow$ & $t_1 \to 1$      \\
				\midrule
				\multicolumn{6}{l}{\emph{Disaggregation}}                                                                                               \\
				\method{} ($R = 5$)$^{\ddagger}$ & \yes  & \textbf{0.1324}\sd{0.0045}    & \underline{0.95} & \underline{0.058}      & \textbf{1.02}    \\
				\method{} ($R = 1$)              & \yesp & \underline{0.1532}\sd{0.0098} & 0.85             & \textbf{0.052}         & 0.86             \\
				Flow matching                    & \yesp & 0.1636\sd{0.0003}             & 0.93             & 0.127                  & 1.33             \\
				FlowDPS                          & \yesp & 0.2612\sd{0.0013}             & \textbf{0.97}    & 0.087                  & \underline{1.12} \\
				SimVP                            & \yesp & 0.2059                        & --               & --                     & 0.84             \\
				Denton--Cholette                 & \yes  & 0.5340                        & --               & --                     & 0.16             \\
\bottomrule
			\end{tabular}}%
		\label{tab:autosim-gpe-wake-L32-mild-sub}
	\end{subtable}
	\label{tab:autosim-L32-mild}
\end{table}

\begin{table}[htp]
	\centering
	\caption{\textbf{AutoSim at $N = 24$, mild tier of each system.}}%
	\setlength{\tabcolsep}{4pt}
	\begin{subtable}[t]{0.49\textwidth}
		\centering
		\caption{\textbf{Advection-diffusion ($N\delta/\tau = 1.5$)}}%
		\resizebox{\linewidth}{!}{%
			\begin{tabular}{lccccc}
				\toprule
				                    & cons. & CRPS $\downarrow$             & SSR $\to 1$      & cov.\ MAE $\downarrow$ & $t_1 \to 1$      \\
				\midrule
				\multicolumn{6}{l}{\emph{Disaggregation}}                                                                                  \\
				\method{} ($R = 4$) & \yes  & 0.0073\sd{0.0001}             & \textbf{1.01}    & \textbf{0.039}         & \underline{1.02} \\
				\method{} ($R = 1$) & \yesp & \underline{0.0063}\sd{0.0001} & 0.67             & 0.067                  & \textbf{1.00}    \\
				Flow matching       & \yesp & \textbf{0.0062}\sd{0.0000}    & \underline{0.77} & 0.073                  & 1.07             \\
				FlowDPS             & \yesp & 0.0162\sd{0.0003}             & 0.40             & \underline{0.049}      & \textbf{1.00}    \\
				SimVP               & \yesp & 0.0114                        & --               & --                     & \underline{0.98} \\
				Denton--Cholette    & \yes  & 0.0673                        & --               & --                     & 0.72             \\
\bottomrule
			\end{tabular}}%
		\label{tab:autosim-advection-diffusion-L24-mild-sub}
	\end{subtable}\hfill
	\begin{subtable}[t]{0.49\textwidth}
		\centering
		\caption{\textbf{Gray--Scott ($N\delta/\tau = 1.2$)}}%
		\resizebox{\linewidth}{!}{%
			\begin{tabular}{lccccc}
				\toprule
				                    & cons. & CRPS $\downarrow$             & SSR $\to 1$      & cov.\ MAE $\downarrow$ & $t_1 \to 1$      \\
				\midrule
				\multicolumn{6}{l}{\emph{Disaggregation}}                                                                                  \\
				\method{} ($R = 4$) & \yes  & \underline{0.0059}\sd{0.0001} & 1.12             & \textbf{0.045}         & \textbf{1.00}    \\
				\method{} ($R = 1$) & \yesp & \textbf{0.0056}\sd{0.0001}    & 1.56             & \underline{0.046}      & \underline{1.01} \\
				Flow matching       & \yesp & 0.0068\sd{0.0001}             & \underline{0.91} & 0.102                  & 1.03             \\
				FlowDPS             & \yesp & 0.0128\sd{0.0003}             & \textbf{1.02}    & 0.053                  & 1.03             \\
				SimVP               & \yesp & 0.0124                        & --               & --                     & \textbf{1.00}    \\
				Denton--Cholette    & \yes  & 0.2044                        & --               & --                     & 0.95             \\
\bottomrule
			\end{tabular}}%
		\label{tab:autosim-gray-scott-L24-mild-sub}
	\end{subtable}\par\medskip
	\begin{subtable}[t]{0.49\textwidth}
		\centering
		\caption{\textbf{Navier--Stokes ($N\delta/\tau = 1.0$)}}%
		\resizebox{\linewidth}{!}{%
			\begin{tabular}{lccccc}
				\toprule
				                                 & cons. & CRPS $\downarrow$             & SSR $\to 1$      & cov.\ MAE $\downarrow$ & $t_1 \to 1$      \\
				\midrule
				\multicolumn{6}{l}{\emph{Disaggregation}}                                                                                               \\
				\method{} ($R = 4$)$^{\ddagger}$ & \yes  & 0.0508\sd{0.0004}             & \textbf{0.74}    & \underline{0.023}      & 1.09             \\
				\method{} ($R = 1$)              & \yesp & \underline{0.0449}\sd{0.0004} & 0.70             & 0.032                  & \textbf{1.01}    \\
				Flow matching                    & \yesp & \textbf{0.0436}\sd{0.0005}    & \underline{0.71} & \textbf{0.022}         & 1.08             \\
				FlowDPS                          & \yesp & 0.0684\sd{0.0002}             & 0.57             & 0.081                  & \underline{1.02} \\
				SimVP                            & \yesp & 0.0742                        & --               & --                     & \textbf{0.99}    \\
				Denton--Cholette                 & \yes  & 0.1731                        & --               & --                     & 0.76             \\
\bottomrule
			\end{tabular}}%
		\label{tab:autosim-conditioned-ns-L24-mild-sub}
	\end{subtable}\hfill
	\begin{subtable}[t]{0.49\textwidth}
		\centering
		\caption{\textbf{Gross--Pitaevskii ($N\delta/\tau = 3.0$)}}%
		\resizebox{\linewidth}{!}{%
			\begin{tabular}{lccccc}
				\toprule
				                                 & cons. & CRPS $\downarrow$             & SSR $\to 1$      & cov.\ MAE $\downarrow$ & $t_1 \to 1$      \\
				\midrule
				\multicolumn{6}{l}{\emph{Disaggregation}}                                                                                               \\
				\method{} ($R = 4$)$^{\ddagger}$ & \yes  & \textbf{0.0939}\sd{0.0014}    & \underline{0.92} & 0.063                  & \textbf{1.03}    \\
				\method{} ($R = 1$)              & \yesp & \underline{0.1000}\sd{0.0021} & 0.82             & \textbf{0.050}         & 0.92             \\
				Flow matching                    & \yesp & 0.1156\sd{0.0001}             & 0.86             & 0.083                  & \underline{1.04} \\
				FlowDPS                          & \yesp & 0.1839\sd{0.0030}             & \textbf{1.00}    & \underline{0.062}      & 0.93             \\
				SimVP                            & \yesp & 0.1631                        & --               & --                     & 0.87             \\
				Denton--Cholette                 & \yes  & 0.5234                        & --               & --                     & 0.20             \\
\bottomrule
			\end{tabular}}%
		\label{tab:autosim-gpe-wake-L24-mild-sub}
	\end{subtable}
	\label{tab:autosim-L24-mild}
\end{table}

\begin{table}[htp]
	\centering
	\caption{\textbf{AutoSim at $N = 16$, mild tier of each system.}}%
	\setlength{\tabcolsep}{4pt}
	\begin{subtable}[t]{0.49\textwidth}
		\centering
		\caption{\textbf{Advection-diffusion ($N\delta/\tau = 1.0$)}}%
		\resizebox{\linewidth}{!}{%
			\begin{tabular}{lccccc}
				\toprule
				                                 & cons. & CRPS $\downarrow$             & SSR $\to 1$      & cov.\ MAE $\downarrow$ & $t_1 \to 1$      \\
				\midrule
				\multicolumn{6}{l}{\emph{Disaggregation}}                                                                                               \\
				\method{} ($R = 4$)$^{\ddagger}$ & \yes  & 0.0056\sd{0.0001}             & \textbf{0.92}    & \underline{0.056}      & 1.02             \\
				\method{} ($R = 1$)              & \yesp & \textbf{0.0044}\sd{0.0001}    & \underline{0.90} & 0.097                  & \underline{1.01} \\
				Flow matching                    & \yesp & \underline{0.0047}\sd{0.0000} & 0.78             & 0.064                  & 1.06             \\
				FlowDPS                          & \yesp & 0.0126\sd{0.0001}             & 0.43             & \textbf{0.042}         & \textbf{1.00}    \\
				SimVP                            & \yesp & 0.0110                        & --               & --                     & 1.02             \\
				Denton--Cholette                 & \yes  & 0.0426                        & --               & --                     & 0.80             \\
\bottomrule
			\end{tabular}}%
		\label{tab:autosim-advection-diffusion-L16-mild-sub}
	\end{subtable}\hfill
	\begin{subtable}[t]{0.49\textwidth}
		\centering
		\caption{\textbf{Gray--Scott ($N\delta/\tau = 0.8$)}}%
		\resizebox{\linewidth}{!}{%
			\begin{tabular}{lccccc}
				\toprule
				                    & cons. & CRPS $\downarrow$             & SSR $\to 1$      & cov.\ MAE $\downarrow$ & $t_1 \to 1$      \\
				\midrule
				\multicolumn{6}{l}{\emph{Disaggregation}}                                                                                  \\
				\method{} ($R = 4$) & \yes  & \textbf{0.0039}\sd{0.0000}    & 1.20             & \textbf{0.042}         & \textbf{1.00}    \\
				\method{} ($R = 1$) & \yesp & \underline{0.0040}\sd{0.0001} & 1.36             & \underline{0.047}      & \textbf{1.00}    \\
				Flow matching       & \yesp & 0.0047\sd{0.0001}             & \textbf{0.94}    & 0.094                  & \underline{1.02} \\
				FlowDPS             & \yesp & 0.0087\sd{0.0001}             & \underline{0.89} & 0.061                  & \underline{1.02} \\
				SimVP               & \yesp & 0.0112                        & --               & --                     & \textbf{1.00}    \\
				Denton--Cholette    & \yes  & 0.1227                        & --               & --                     & 0.93             \\
\bottomrule
			\end{tabular}}%
		\label{tab:autosim-gray-scott-L16-mild-sub}
	\end{subtable}\par\medskip
	\begin{subtable}[t]{0.49\textwidth}
		\centering
		\caption{\textbf{Navier--Stokes ($N\delta/\tau = 0.6$)}}%
		\resizebox{\linewidth}{!}{%
			\begin{tabular}{lccccc}
				\toprule
				                                 & cons. & CRPS $\downarrow$             & SSR $\to 1$      & cov.\ MAE $\downarrow$ & $t_1 \to 1$      \\
				\midrule
				\multicolumn{6}{l}{\emph{Disaggregation}}                                                                                               \\
				\method{} ($R = 4$)              & \yes  & 0.0246\sd{0.0004}             & \textbf{0.83}    & 0.040                  & 1.04             \\
				\method{} ($R = 1$)              & \yesp & \textbf{0.0197}\sd{0.0002}    & 0.71             & \textbf{0.017}         & \textbf{1.02}    \\
				Flow matching                    & \yesp & \underline{0.0210}\sd{0.0003} & \underline{0.73} & \underline{0.018}      & 1.04             \\
				FlowDPS                          & \yesp & 0.0340\sd{0.0003}             & 0.57             & 0.062                  & \underline{1.03} \\
				SimVP                            & \yesp & 0.0399                        & --               & --                     & \underline{1.03} \\
				Denton--Cholette                 & \yes  & 0.1092                        & --               & --                     & 0.84             \\
\bottomrule
			\end{tabular}}%
		\label{tab:autosim-conditioned-ns-L16-mild-sub}
	\end{subtable}\hfill
	\begin{subtable}[t]{0.49\textwidth}
		\centering
		\caption{\textbf{Gross--Pitaevskii ($N\delta/\tau = 2.0$)}}%
		\resizebox{\linewidth}{!}{%
			\begin{tabular}{lccccc}
				\toprule
				                                 & cons. & CRPS $\downarrow$             & SSR $\to 1$      & cov.\ MAE $\downarrow$ & $t_1 \to 1$      \\
				\midrule
				\multicolumn{6}{l}{\emph{Disaggregation}}                                                                                               \\
				\method{} ($R = 4$)$^{\ddagger}$ & \yes  & \textbf{0.0636}\sd{0.0004}    & \textbf{0.95}    & 0.066                  & 1.04             \\
				\method{} ($R = 1$)              & \yesp & \underline{0.0667}\sd{0.0004} & 0.78             & \textbf{0.051}         & \underline{0.97} \\
				Flow matching                    & \yesp & 0.0784\sd{0.0006}             & \underline{0.85} & 0.063                  & \textbf{1.00}    \\
				FlowDPS                          & \yesp & 0.1123\sd{0.0009}             & 0.71             & \underline{0.054}      & 0.91             \\
				SimVP                            & \yesp & 0.1246                        & --               & --                     & 0.90             \\
				Denton--Cholette                 & \yes  & 0.5085                        & --               & --                     & 0.28             \\
\bottomrule
			\end{tabular}}%
		\label{tab:autosim-gpe-wake-L16-mild-sub}
	\end{subtable}
	\label{tab:autosim-L16-mild}
\end{table}

\begin{table}[htp]
	\centering
	\caption{\textbf{AutoSim at $N = 8$, mild tier of each system.}}%
	\setlength{\tabcolsep}{4pt}
	\begin{subtable}[t]{0.49\textwidth}
		\centering
		\caption{\textbf{Advection-diffusion ($N\delta/\tau = 0.5$)}}%
		\resizebox{\linewidth}{!}{%
			\begin{tabular}{lccccc}
				\toprule
				                                 & cons. & CRPS $\downarrow$             & SSR $\to 1$      & cov.\ MAE $\downarrow$ & $t_1 \to 1$      \\
				\midrule
				\multicolumn{6}{l}{\emph{Disaggregation}}                                                                                               \\
				\method{} ($R = 3$)$^{\ddagger}$ & \yes  & 0.0031\sd{0.0001}             & \underline{0.89} & 0.112                  & \underline{1.02} \\
				\method{} ($R = 1$)              & \yesp & \textbf{0.0021}\sd{0.0001}    & \textbf{0.99}    & 0.145                  & 1.03             \\
				Flow matching                    & \yesp & \underline{0.0026}\sd{0.0000} & 0.78             & \underline{0.108}      & 1.05             \\
				FlowDPS                          & \yesp & 0.0062\sd{0.0001}             & 0.53             & \textbf{0.068}         & 1.03             \\
				SimVP                            & \yesp & 0.0063                        & --               & --                     & \textbf{1.01}    \\
				Denton--Cholette                 & \yes  & 0.0169                        & --               & --                     & 0.90             \\
\bottomrule
			\end{tabular}}%
		\label{tab:autosim-advection-diffusion-L8-mild-sub}
	\end{subtable}\hfill
	\begin{subtable}[t]{0.49\textwidth}
		\centering
		\caption{\textbf{Gray--Scott ($N\delta/\tau = 0.4$)}}%
		\resizebox{\linewidth}{!}{%
			\begin{tabular}{lccccc}
				\toprule
				                                 & cons. & CRPS $\downarrow$             & SSR $\to 1$      & cov.\ MAE $\downarrow$ & $t_1 \to 1$      \\
				\midrule
				\multicolumn{6}{l}{\emph{Disaggregation}}                                                                                               \\
				\method{} ($R = 3$)$^{\ddagger}$ & \yes  & \textbf{0.0018}\sd{0.0000}    & 1.89             & \underline{0.063}      & \textbf{1.00}    \\
				\method{} ($R = 1$)              & \yesp & \underline{0.0019}\sd{0.0001} & 1.43             & \textbf{0.059}         & \textbf{1.00}    \\
				Flow matching                    & \yesp & 0.0020\sd{0.0000}             & \textbf{1.01}    & 0.096                  & \underline{1.01} \\
				FlowDPS                          & \yesp & 0.0031\sd{0.0000}             & \underline{0.79} & 0.065                  & 1.02             \\
				SimVP                            & \yesp & 0.0068                        & --               & --                     & \textbf{1.00}    \\
				Denton--Cholette                 & \yes  & 0.0374                        & --               & --                     & 0.94             \\
\bottomrule
			\end{tabular}}%
		\label{tab:autosim-gray-scott-L8-mild-sub}
	\end{subtable}\par\medskip
	\begin{subtable}[t]{0.49\textwidth}
		\centering
		\caption{\textbf{Navier--Stokes ($N\delta/\tau = 0.3$)}}%
		\resizebox{\linewidth}{!}{%
			\begin{tabular}{lccccc}
				\toprule
				                                 & cons. & CRPS $\downarrow$             & SSR $\to 1$      & cov.\ MAE $\downarrow$ & $t_1 \to 1$      \\
				\midrule
				\multicolumn{6}{l}{\emph{Disaggregation}}                                                                                               \\
				\method{} ($R = 3$)              & \yes  & \underline{0.0064}\sd{0.0002} & \textbf{0.89}    & 0.071                  & \underline{1.03} \\
				\method{} ($R = 1$)              & \yesp & \textbf{0.0053}\sd{0.0000}    & \underline{0.77} & 0.041                  & \textbf{1.02}    \\
				Flow matching                    & \yesp & 0.0065\sd{0.0000}             & 0.73             & \textbf{0.025}         & \underline{1.03} \\
				FlowDPS                          & \yesp & 0.0098\sd{0.0001}             & 0.55             & \underline{0.032}      & 1.04             \\
				SimVP                            & \yesp & 0.0159                        & --               & --                     & \textbf{1.02}    \\
				Denton--Cholette                 & \yes  & 0.0330                        & --               & --                     & 0.95             \\
\bottomrule
			\end{tabular}}%
		\label{tab:autosim-conditioned-ns-L8-mild-sub}
	\end{subtable}\hfill
	\begin{subtable}[t]{0.49\textwidth}
		\centering
		\caption{\textbf{Gross--Pitaevskii ($N\delta/\tau = 1.0$)}}%
		\resizebox{\linewidth}{!}{%
			\begin{tabular}{lccccc}
				\toprule
				                                 & cons. & CRPS $\downarrow$             & SSR $\to 1$      & cov.\ MAE $\downarrow$ & $t_1 \to 1$      \\
				\midrule
				\multicolumn{6}{l}{\emph{Disaggregation}}                                                                                               \\
				\method{} ($R = 3$)$^{\ddagger}$ & \yes  & \underline{0.0216}\sd{0.0000} & \textbf{0.98}    & 0.083                  & 1.04             \\
				\method{} ($R = 1$)              & \yesp & \textbf{0.0193}\sd{0.0002}    & \underline{0.97} & 0.077                  & \underline{1.03} \\
				Flow matching                    & \yesp & 0.0261\sd{0.0003}             & 0.88             & \underline{0.073}      & \textbf{1.02}    \\
				FlowDPS                          & \yesp & 0.0535\sd{0.0009}             & 0.59             & \textbf{0.059}         & 0.96             \\
				SimVP                            & \yesp & 0.0546                        & --               & --                     & \textbf{0.98}    \\
				Denton--Cholette                 & \yes  & 0.3224                        & --               & --                     & 0.61             \\
\bottomrule
			\end{tabular}}%
		\label{tab:autosim-gpe-wake-L8-mild-sub}
	\end{subtable}
	\label{tab:autosim-L8-mild}
\end{table}

\begin{table}[htp]
	\centering
	\caption{\textbf{AutoSim at $N = 24$, most chaotic tier of each system.}}%
	\setlength{\tabcolsep}{4pt}
	\begin{subtable}[t]{0.49\textwidth}
		\centering
		\caption{\textbf{Advection-diffusion ($N\delta/\tau = 3.0$)}}%
		\resizebox{\linewidth}{!}{%
			\begin{tabular}{lccccc}
				\toprule
				                                 & cons. & CRPS $\downarrow$             & SSR $\to 1$      & cov.\ MAE $\downarrow$ & $t_1 \to 1$      \\
				\midrule
				\multicolumn{6}{l}{\emph{Disaggregation}}                                                                                               \\
				\method{} ($R = 4$)$^{\ddagger}$ & \yes  & \underline{0.0228}\sd{0.0003} & \textbf{0.99}    & \textbf{0.011}         & \textbf{0.96}    \\
				\method{} ($R = 1$)              & \yesp & 0.0349\sd{0.0015}             & 0.58             & \underline{0.040}      & 0.77             \\
				Flow matching                    & \yesp & \textbf{0.0205}\sd{0.0001}    & \underline{0.78} & 0.046                  & \textbf{0.96}    \\
				FlowDPS                          & \yesp & 0.0431\sd{0.0004}             & 0.45             & 0.173                  & 0.76             \\
				SimVP                            & \yesp & 0.0314                        & --               & --                     & \underline{0.82} \\
				Denton--Cholette                 & \yes  & 0.0984                        & --               & --                     & 0.41             \\
\bottomrule
			\end{tabular}}%
		\label{tab:autosim-advection-diffusion-L24-chaotic-sub}
	\end{subtable}\hfill
	\begin{subtable}[t]{0.49\textwidth}
		\centering
		\caption{\textbf{Gray--Scott ($N\delta/\tau = 2.4$)}}%
		\resizebox{\linewidth}{!}{%
			\begin{tabular}{lccccc}
				\toprule
				                                 & cons. & CRPS $\downarrow$             & SSR $\to 1$      & cov.\ MAE $\downarrow$ & $t_1 \to 1$      \\
				\midrule
				\multicolumn{6}{l}{\emph{Disaggregation}}                                                                                               \\
				\method{} ($R = 4$)$^{\ddagger}$ & \yes  & 0.0126\sd{0.0001}             & 0.79             & 0.054                  & \underline{1.01} \\
				\method{} ($R = 1$)              & \yesp & \textbf{0.0100}\sd{0.0000}    & 0.76             & \underline{0.031}      & \textbf{1.00}    \\
				Flow matching                    & \yesp & \underline{0.0118}\sd{0.0002} & \underline{0.81} & 0.077                  & \underline{1.01} \\
				FlowDPS                          & \yesp & 0.0303\sd{0.0003}             & \textbf{1.04}    & \textbf{0.018}         & \underline{1.01} \\
				SimVP                            & \yesp & 0.0221                        & --               & --                     & \textbf{1.00}    \\
				Denton--Cholette                 & \yes  & 0.5010                        & --               & --                     & 0.73             \\
\bottomrule
			\end{tabular}}%
		\label{tab:autosim-gray-scott-L24-chaotic-sub}
	\end{subtable}\par\medskip
	\begin{subtable}[t]{0.49\textwidth}
		\centering
		\caption{\textbf{Navier--Stokes ($N\delta/\tau = 1.1$)}}%
		\resizebox{\linewidth}{!}{%
			\begin{tabular}{lccccc}
				\toprule
				                                 & cons. & CRPS $\downarrow$             & SSR $\to 1$      & cov.\ MAE $\downarrow$ & $t_1 \to 1$      \\
				\midrule
				\multicolumn{6}{l}{\emph{Disaggregation}}                                                                                               \\
				\method{} ($R = 4$)$^{\ddagger}$ & \yes  & 0.0740\sd{0.0008}             & \underline{0.77} & 0.027                  & 1.13             \\
				\method{} ($R = 1$)              & \yesp & \underline{0.0715}\sd{0.0028} & \textbf{0.79}    & \textbf{0.016}         & \textbf{0.99}    \\
				Flow matching                    & \yesp & \textbf{0.0618}\sd{0.0001}    & 0.71             & \underline{0.021}      & 1.06             \\
				FlowDPS                          & \yesp & 0.0917\sd{0.0005}             & 0.58             & 0.087                  & \underline{0.98} \\
				SimVP                            & \yesp & 0.1002                        & --               & --                     & 0.96             \\
				Denton--Cholette                 & \yes  & 0.1877                        & --               & --                     & 0.70             \\
\bottomrule
			\end{tabular}}%
		\label{tab:autosim-conditioned-ns-L24-chaotic-sub}
	\end{subtable}
	\label{tab:autosim-L24-chaotic}
\end{table}

\begin{table}[htp]
	\centering
	\caption{\textbf{AutoSim at $N = 16$, most chaotic tier of each system.}}%
	\setlength{\tabcolsep}{4pt}
	\begin{subtable}[t]{0.49\textwidth}
		\centering
		\caption{\textbf{Advection-diffusion ($N\delta/\tau = 2.0$)}}%
		\resizebox{\linewidth}{!}{%
			\begin{tabular}{lccccc}
				\toprule
				                                 & cons. & CRPS $\downarrow$             & SSR $\to 1$      & cov.\ MAE $\downarrow$ & $t_1 \to 1$      \\
				\midrule
				\multicolumn{6}{l}{\emph{Disaggregation}}                                                                                               \\
				\method{} ($R = 4$)$^{\ddagger}$ & \yes  & \underline{0.0186}\sd{0.0001} & \textbf{0.95}    & \textbf{0.012}         & \underline{0.95} \\
				\method{} ($R = 1$)              & \yesp & 0.0229\sd{0.0035}             & 0.66             & \underline{0.020}      & 0.86             \\
				Flow matching                    & \yesp & \textbf{0.0159}\sd{0.0001}    & \underline{0.79} & 0.052                  & \textbf{0.97}    \\
				FlowDPS                          & \yesp & 0.0338\sd{0.0004}             & 0.42             & 0.160                  & 0.79             \\
				SimVP                            & \yesp & 0.0292                        & --               & --                     & 0.85             \\
				Denton--Cholette                 & \yes  & 0.0794                        & --               & --                     & 0.51             \\
\bottomrule
			\end{tabular}}%
		\label{tab:autosim-advection-diffusion-L16-chaotic-sub}
	\end{subtable}\hfill
	\begin{subtable}[t]{0.49\textwidth}
		\centering
		\caption{\textbf{Gray--Scott ($N\delta/\tau = 1.6$)}}%
		\resizebox{\linewidth}{!}{%
			\begin{tabular}{lccccc}
				\toprule
				                                 & cons. & CRPS $\downarrow$             & SSR $\to 1$      & cov.\ MAE $\downarrow$ & $t_1 \to 1$      \\
				\midrule
				\multicolumn{6}{l}{\emph{Disaggregation}}                                                                                               \\
				\method{} ($R = 4$)$^{\ddagger}$ & \yes  & 0.0087\sd{0.0000}             & \textbf{0.96}    & 0.053                  & \textbf{1.00}    \\
				\method{} ($R = 1$)              & \yesp & \textbf{0.0070}\sd{0.0002}    & \underline{0.83} & \underline{0.038}      & \textbf{1.00}    \\
				Flow matching                    & \yesp & \underline{0.0080}\sd{0.0001} & 0.78             & 0.060                  & \underline{1.01} \\
				FlowDPS                          & \yesp & 0.0173\sd{0.0002}             & \textbf{1.04}    & \textbf{0.027}         & \underline{1.01} \\
				SimVP                            & \yesp & 0.0168                        & --               & --                     & \textbf{1.00}    \\
				Denton--Cholette                 & \yes  & 0.3029                        & --               & --                     & 0.88             \\
\bottomrule
			\end{tabular}}%
		\label{tab:autosim-gray-scott-L16-chaotic-sub}
	\end{subtable}\par\medskip
	\begin{subtable}[t]{0.49\textwidth}
		\centering
		\caption{\textbf{Navier--Stokes ($N\delta/\tau = 0.8$)}}%
		\resizebox{\linewidth}{!}{%
			\begin{tabular}{lccccc}
				\toprule
				                                 & cons. & CRPS $\downarrow$             & SSR $\to 1$      & cov.\ MAE $\downarrow$ & $t_1 \to 1$      \\
				\midrule
				\multicolumn{6}{l}{\emph{Disaggregation}}                                                                                               \\
				\method{} ($R = 4$)$^{\ddagger}$ & \yes  & 0.0402\sd{0.0007}             & \textbf{0.78}    & 0.048                  & 1.09             \\
				\method{} ($R = 1$)              & \yesp & \underline{0.0338}\sd{0.0001} & 0.68             & \underline{0.018}      & \underline{1.02} \\
				Flow matching                    & \yesp & \textbf{0.0330}\sd{0.0002}    & \underline{0.71} & \textbf{0.017}         & 1.03             \\
				FlowDPS                          & \yesp & 0.0507\sd{0.0005}             & 0.59             & 0.066                  & \textbf{1.00}    \\
				SimVP                            & \yesp & 0.0594                        & --               & --                     & \underline{1.02} \\
				Denton--Cholette                 & \yes  & 0.1341                        & --               & --                     & 0.78             \\
\bottomrule
			\end{tabular}}%
		\label{tab:autosim-conditioned-ns-L16-chaotic-sub}
	\end{subtable}
	\label{tab:autosim-L16-chaotic}
\end{table}

\begin{table}[htp]
	\centering
	\caption{\textbf{AutoSim at $N = 8$, most chaotic tier of each system.}}%
	\setlength{\tabcolsep}{4pt}
	\begin{subtable}[t]{0.49\textwidth}
		\centering
		\caption{\textbf{Advection-diffusion ($N\delta/\tau = 1.0$)}}%
		\resizebox{\linewidth}{!}{%
			\begin{tabular}{lccccc}
				\toprule
				                                 & cons. & CRPS $\downarrow$             & SSR $\to 1$      & cov.\ MAE $\downarrow$ & $t_1 \to 1$      \\
				\midrule
				\multicolumn{6}{l}{\emph{Disaggregation}}                                                                                               \\
				\method{} ($R = 3$)$^{\ddagger}$ & \yes  & 0.0121\sd{0.0002}             & \textbf{0.94}    & \textbf{0.041}         & \textbf{1.00}    \\
				\method{} ($R = 1$)              & \yesp & \textbf{0.0098}\sd{0.0002}    & \underline{0.87} & \underline{0.053}      & \underline{0.98} \\
				Flow matching                    & \yesp & \underline{0.0099}\sd{0.0000} & 0.80             & 0.054                  & \underline{0.98} \\
				FlowDPS                          & \yesp & 0.0205\sd{0.0003}             & 0.51             & 0.109                  & 0.87             \\
				SimVP                            & \yesp & 0.0213                        & --               & --                     & 0.88             \\
				Denton--Cholette                 & \yes  & 0.0504                        & --               & --                     & 0.66             \\
\bottomrule
			\end{tabular}}%
		\label{tab:autosim-advection-diffusion-L8-chaotic-sub}
	\end{subtable}\hfill
	\begin{subtable}[t]{0.49\textwidth}
		\centering
		\caption{\textbf{Gray--Scott ($N\delta/\tau = 0.8$)}}%
		\resizebox{\linewidth}{!}{%
			\begin{tabular}{lccccc}
				\toprule
				                                 & cons. & CRPS $\downarrow$             & SSR $\to 1$      & cov.\ MAE $\downarrow$ & $t_1 \to 1$      \\
				\midrule
				\multicolumn{6}{l}{\emph{Disaggregation}}                                                                                               \\
				\method{} ($R = 3$)$^{\ddagger}$ & \yes  & \underline{0.0028}\sd{0.0000} & 1.48             & 0.046                  & \textbf{1.00}    \\
				\method{} ($R = 1$)              & \yesp & \textbf{0.0027}\sd{0.0000}    & 1.35             & 0.043                  & \textbf{1.00}    \\
				Flow matching                    & \yesp & \underline{0.0028}\sd{0.0000} & \textbf{0.92}    & \underline{0.041}      & \textbf{1.00}    \\
				FlowDPS                          & \yesp & 0.0051\sd{0.0001}             & \underline{0.85} & \textbf{0.021}         & \textbf{1.00}    \\
				SimVP                            & \yesp & 0.0081                        & --               & --                     & \textbf{1.00}    \\
				Denton--Cholette                 & \yes  & 0.0938                        & --               & --                     & \underline{0.98} \\
\bottomrule
			\end{tabular}}%
		\label{tab:autosim-gray-scott-L8-chaotic-sub}
	\end{subtable}\par\medskip
	\begin{subtable}[t]{0.49\textwidth}
		\centering
		\caption{\textbf{Navier--Stokes ($N\delta/\tau = 0.4$)}}%
		\resizebox{\linewidth}{!}{%
			\begin{tabular}{lccccc}
				\toprule
				                                 & cons. & CRPS $\downarrow$             & SSR $\to 1$      & cov.\ MAE $\downarrow$ & $t_1 \to 1$      \\
				\midrule
				\multicolumn{6}{l}{\emph{Disaggregation}}                                                                                               \\
				\method{} ($R = 3$)$^{\ddagger}$ & \yes  & 0.0107\sd{0.0002}             & \textbf{0.80}    & 0.079                  & 1.04             \\
				\method{} ($R = 1$)              & \yesp & \textbf{0.0088}\sd{0.0001}    & \underline{0.72} & \underline{0.042}      & \underline{1.03} \\
				Flow matching                    & \yesp & \underline{0.0099}\sd{0.0001} & 0.69             & \textbf{0.019}         & \underline{1.03} \\
				FlowDPS                          & \yesp & 0.0154\sd{0.0002}             & 0.52             & 0.046                  & \underline{1.03} \\
				SimVP                            & \yesp & 0.0224                        & --               & --                     & \textbf{1.02}    \\
				Denton--Cholette                 & \yes  & 0.0494                        & --               & --                     & 0.92             \\
\bottomrule
			\end{tabular}}%
		\label{tab:autosim-conditioned-ns-L8-chaotic-sub}
	\end{subtable}
	\label{tab:autosim-L8-chaotic}
\end{table}

\end{document}